%% file: main.tex
\documentclass[10pt,twocolumn,letterpaper]{article}

\usepackage{iccv}              


\definecolor{iccvblue}{rgb}{0.21,0.49,0.74}
\usepackage[pagebackref,breaklinks,colorlinks,allcolors=iccvblue]{hyperref}
\usepackage{subcaption}
\usepackage{adjustbox}   
\usepackage{multirow}

\def\paperID{9042} 
\def\confName{ICCV}
\def\confYear{2025}

\title{Detecting Human Artifacts from Text-to-Image Models}

\author{Kaihong Wang\thanks{Work partially done during an internship at Adobe.}\\
Boston University\\
{\tt\small kaiwkh@bu.edu}
\and
Lingzhi Zhang\\
Adobe Research\\
{\tt\small lingzzha@adobe.com}
\and
Jianming Zhang\\
Adobe Research\\
{\tt\small jianmzha@adobe.com}
}

\newcommand{\ourdataset}{Human Artifact Dataset}
\newcommand{\abbrourdataset}{HAD}

\newcommand{\ourmodel}{Human Artifact Detection Models}
\newcommand{\abbrourmodel}{HADM}

\newcommand{\ourlocalmodel}{Local Human Artifact Detection Model}
\newcommand{\abbrourlocalmodel}{HADM-L}

\newcommand{\ourglobalmodel}{Global Human Artifact Detection Model}
\newcommand{\abbrourglobalmodel}{HADM-G}

\begin{document}

\maketitle

\begin{abstract}
Despite recent advancements, text-to-image generation models often produce images containing artifacts, especially in human figures. These artifacts appear as poorly generated human bodies, including distorted, missing, or extra body parts, leading to visual inconsistencies with typical human anatomy and greatly impairing overall fidelity. In this study, we address this challenge by curating Human Artifact Dataset (HAD), a diverse dataset specifically designed to localize human artifacts. HAD comprises over 37,000 images generated by several popular text-to-image models, annotated for human artifact localization. Using this dataset, we train the Human Artifact Detection Models (HADM), which can identify different artifacts across multiple generative domains and demonstrate strong generalization, even on images from unseen generators. Additionally, to further improve generators’ perception of human structural coherence, we use the predictions from our HADM as feedback for diffusion model finetuning. Our experiments confirm a reduction in human artifacts in the resulting model. Furthermore, we showcase a novel application of our HADM in an iterative inpainting framework to correct human artifacts in arbitrary images directly, demonstrating its utility in improving image quality. Our dataset and detection models are available at: \url{https://github.com/wangkaihong/HADM}.
\end{abstract}

\vspace{-0.5cm}

\section{Introduction}
\label{sec:intro}

\begin{figure}[t]
  \centering
   \includegraphics[width=\linewidth]{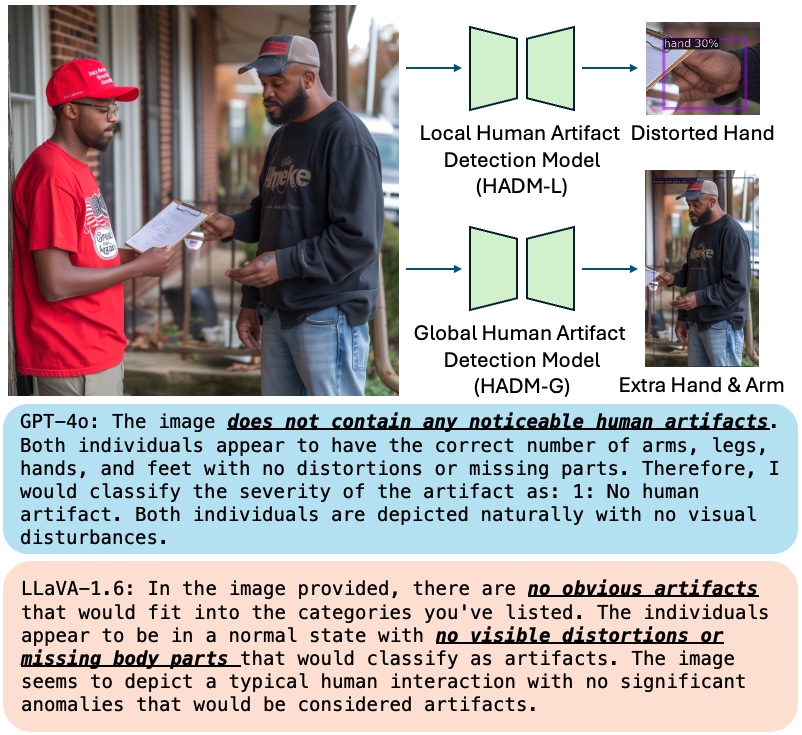}
   \vspace{-0.6cm}
   \caption{Comparison between our ~\ourmodel~(\abbrourmodel) and state-of-the-art vision-language models in detecting human artifacts in an influential deepfake image circulating on social media during the 2024 U.S. presidential election. While advanced VL models fail to detect visible artifacts of the right human figure in the image (shown in the responses at the bottom), our models successfully identify and localize the distorted hands and the extra limb (top right).
   Source: \url{https://farid.berkeley.edu/deepfakes2024election/}.
}
\vspace{-0.6cm}
   \label{fig:teaser}
\end{figure}

Recent advancements in text-to-image generation~\cite{MansimovPaBaSa15,ZhuPaChYa19,YeYaTaSuJi21,TaoTaWuJiBaXu22,GafniPpAsShPaTa22,RoBlaLoEsOm22} have led to applications in various areas, including image editing~\cite{BrooksHoEf23,KawarZaLaToChDeMoIr23,SheyninPoSiKiZoAsPaTa24}, representation learning from synthetic data~\cite{TianFaIsChKr23,TianFaChKaKrIs24,HammoudItPiToBiGh24,SinghNaHoScRo24}, and more. Current state-of-the-art diffusion models~\cite{RoBlaLoEsOm22,openai2022dalle2,openai2023dalle3,midjourney2023,PoEnLaBlDoMuPeRo24,flux2024,XieChChCaTaLiZhLiZhLuHa24} are primarily trained on large proprietary datasets, with prohibitively high training costs, resulting in impressive overall quality improvements in generated images.
Despite these advancements, \textit{\textbf{human artifacts}} remain persistent challenges across models, frequently leading to anomalies such as distorted, missing, or extra body parts.
Such issues significantly compromise \textit{\textbf{human structural coherence}}, which we define as the consistency with typical human anatomy, and greatly undermine the overall fidelity of images.
For instance, in Fig.~\ref{fig:teaser}, artifacts on the human figure on the right are clearly visible, as highlighted by detection results from our models. However, when analyzed by advanced vision-language (VL) models, such as GPT-4o~\cite{openai2024gpt4o} and LLaVA~\cite{liu2024llavanext}, these artifacts are not identified. This highlights the surprisingly limited grasp of human structural coherence even in state-of-the-art VL models, despite their strong performance in various downstream tasks. In this work, we propose an alternative approach that effectively detects and addresses human artifacts in images from various text-to-image models.

Existing works on evaluating generative image quality primarily focus on global attributes like text-image alignment~\cite{HuangSuXiLiLi23,LinPaLiLiXiNeZhRa24,LiLiPaLiFeWuLiXiZhNeRa24}, aesthetics~\cite{WuSuZhZhLi23,KiPoSiMaPeLe23,XuLiWuToLiDiTaDo23,WuHaSuChZhZhLi23,LiHeLiLiKlCaSuPoYoYaKeDvCoLuLoKoRaNa24}, or general local artifacts~\cite{ZhangXuBaZhLiZhAmLiShSh23,CaoYuLiLiSuLiZh24,LiHeLiLiKlCaSuPoYoYaKeDvCoLuLoKoRaNa24}. While theoretically related to human artifact detection, these methods are often distracted by irrelevant elements such as image layout or background objects. More recent efforts~\cite{FangYaGuHaJiXuLiLi24,CaoYuLiLiSuLiZh24,WangMaWaLiWaTi24} attempt to tackle human-related artifacts but struggle to localize diverse human artifacts under challenging scenarios robustly.

To effectively identify different types of human artifacts in generated images, we first define two primary categories: local artifacts, which focus on localized regions of human body parts with poorly rendered features (e.g., flawed patterns, textures, or shapes), and global artifacts, which reflect structural or anatomical inconsistency to typical human figures, such as missing or extra body parts. We curate a dataset focusing on human artifacts in synthetically generated images, namely~\ourdataset~(\abbrourdataset), from four widely used diffusion models, i.e., SDXL~\cite{PoEnLaBlDoMuPeRo24}, DALLE-2~\cite{openai2022dalle2}, DALLE-3~\cite{openai2023dalle3}, and Midjourney~\cite{midjourney2023}, annotated with bounding boxes and artifact types, resulting in 37,554 samples with 84,852 labeled instances covering various types of local and global artifacts.

We train the \ourmodel~(\abbrourmodel) on our dataset to identify local and global human artifacts, respectively. Utilizing a state-of-the-art architecture and regularized with diverse real human images, \abbrourmodel~demonstrate significant performance advantages over existing methods, including advanced VL models, excelling at detecting various human artifacts in our dataset and generalizing effectively to images from more advanced generators not included in the training data.

Furthermore, beyond identifying human artifacts, we demonstrate that ~\abbrourmodel~ provide valuable guidance for mitigating these artifacts. Specifically, we finetune a diffusion model on images paired with special identifiers representing different types of detected human artifacts, which are later applied as negative prompts during inference. This approach enables the diffusion model to recognize and avoid poorly generated human features, resulting in reduced human artifacts, as validated by our experimental results. Additionally, we showcase another utility of ~\abbrourmodel~ as a guide to effectively automate the inpainting pipeline to identify and rectify human artifacts in arbitrary images.

To summarize, the contributions of this work include:
\begin{itemize}
\item We curate the \ourdataset~(\abbrourdataset), a diverse dataset dedicated to detecting human artifacts in images generated by different text-to-image models, featuring bounding box annotations for precise localization.
\item By training on ~\abbrourdataset, the \ourmodel~(\abbrourmodel) effectively detect various human artifacts from different generative models, demonstrating robust performance and generalization on images from different generators, including unseen ones.
\item We highlight the impactful application of ~\abbrourmodel’s predictions as guidance for finetuning diffusion models and assisting inpainting workflows, further reducing human artifacts in generated images.
\end{itemize}

\section{Related Works}
\label{sec:relatedworks}

\subsection{Text-to-Image Generation}
\label{sec:relatedworks_dm}

Text-to-image (T2I) models have evolved through several different main architectures. GANs~\cite{ZhuPaChYa19,YeYaTaSuJi21,TaoTaWuJiBaXu22} were early approaches, training a generator-discriminator pair to produce realistic images. Autoregressive models~\cite{MansimovPaBaSa15}, such as DALLE~\cite{RameshPaGoGrVoRaChSu21}, introduced sequential generation, later refined by VQ-VAE~\cite{GafniPpAsShPaTa22} for better control and detail.
The current state-of-the-art, Diffusion Models (DMs)~\cite{SoWeMaGa15,HoJaAb20,NiDRaShMiMcSuCh22}, generate images by reversing noise, capturing high diversity and quality. Latent Diffusion Models (LDMs)~\cite{RoBlaLoEsOm22} improve efficiency by operating in latent space, with SDXL~\cite{PoEnLaBlDoMuPeRo24} enabling high-resolution synthesis.

\subsection{Evaluation of Text-to-Image Generators}
\label{sec:relatedworks_hfdm}

Typical evaluation metrics for text-to-image models include automated scores such as FID~\cite{HeuselRaUnNeHo17,ParmarZhZh22}, Inception Score~\cite{SalimansGoZaChRaCh16}, and CLIPScore~\cite{HesselHoFoBrCh21}. However, with rapid advancements in T2I models, human perception evaluation~\cite{ZhouGoKrNaLiBe19,OtaniToSaIsNaRaHeSa23} has become increasingly important, and more recent works propose comprehensive benchmarks that assess various aspects of T2I generation~\cite{HuangSuXiLiLi23,LeYaMeMaPaGuZhNaTeBeKaPaLeZhLiWuErLi23,WuHaSuChZhZhLi23,LinPaLiLiXiNeZhRa24,LiLiPaLiFeWuLiXiZhNeRa24}.
Inspired by Reinforcement Learning from Human Feedback (RLHF), several recent works 
aim to model human preference, mainly on aesthetics or identifying general artifacts, from curated datasets and further align T2I models or directly address local artifacts.
HPD-v1~\cite{WuSuZhZhLi23} assigns scores to images to guide models toward human-preferred quality, while Pick-A-Pic~\cite{KiPoSiMaPeLe23} and ImageReward~\cite{XuLiWuToLiDiTaDo23} use paired comparisons for more effective preference modeling. 
RichHF-18K~\cite{LiHeLiLiKlCaSuPoYoYaKeDvCoLuLoKoRaNa24} further includes annotations for general local artifacts, implausibilities, and misalignment between text and image. Zhang et al.~\cite{ZhangXuBaZhLiZhAmLiShSh23} train binary segmentation model on their dataset with pixel-level annotations.
Despite these efforts, none of these methods explicitly identify and localize human artifacts. 

\begin{figure*}[t] 
  \centering
  \begin{subfigure}[b]{0.23\linewidth}
    \centering
    \includegraphics[width=\linewidth]{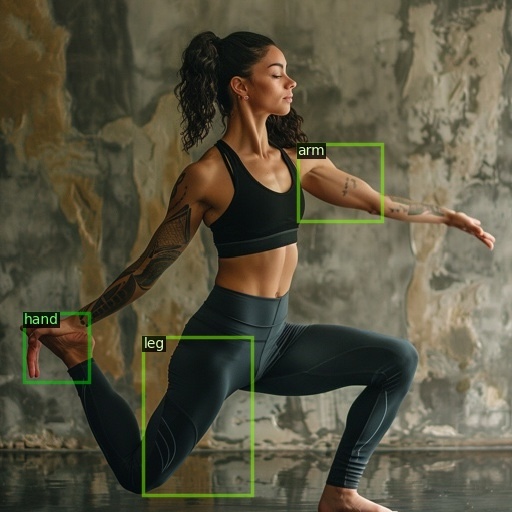}
    \caption{Local artifacts in Midjourney.}
    \label{fig:subfig_a}
  \end{subfigure}
  \hfill
  \begin{subfigure}[b]{0.23\linewidth}
    \centering
    \includegraphics[width=\linewidth]{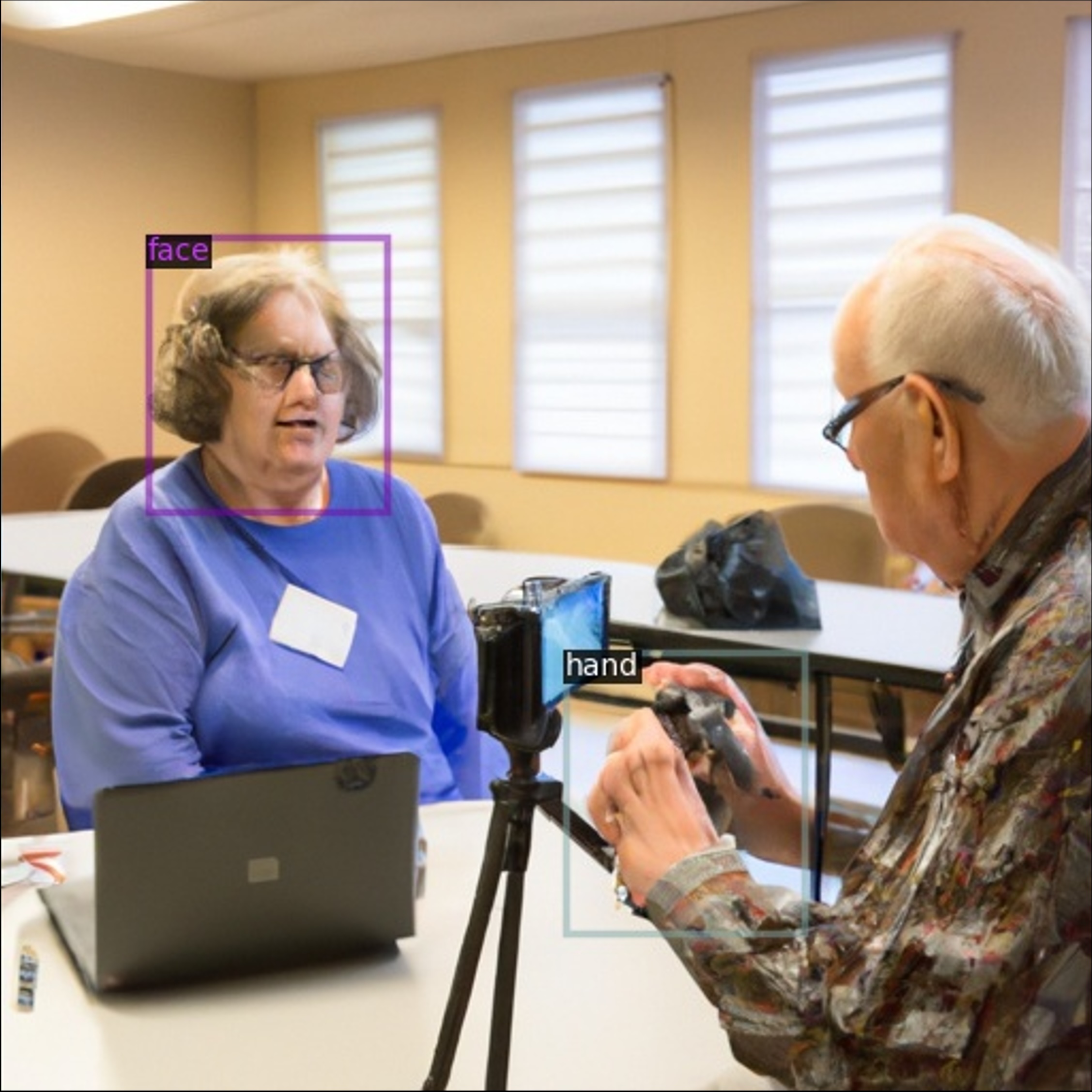}
    \caption{Local artifacts in DALLE-2.}
    \label{fig:subfig_b}
  \end{subfigure}
  \hfill
  \begin{subfigure}[b]{0.23\linewidth}
    \centering
    \includegraphics[width=\linewidth]{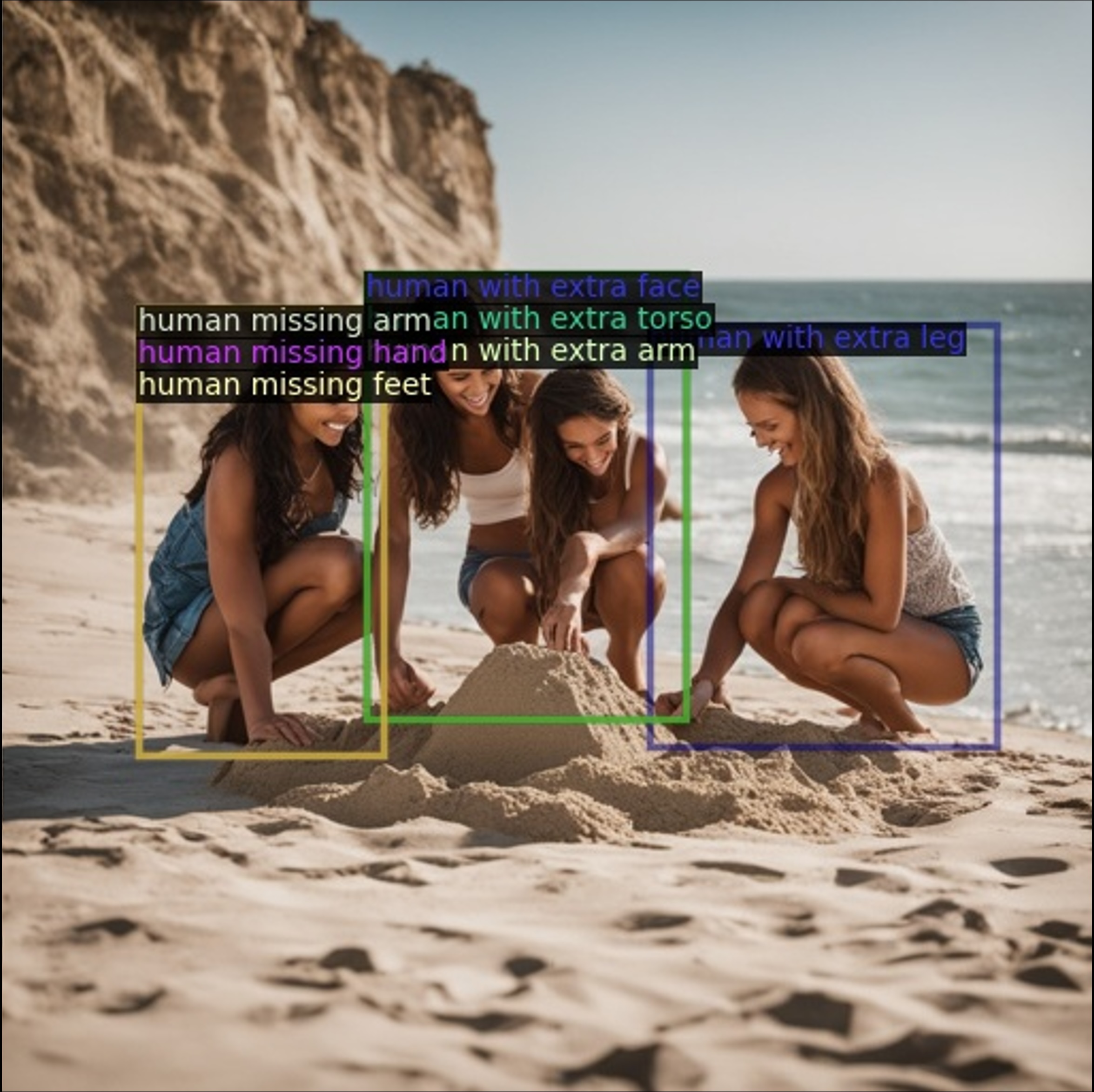}
    \caption{Global artifacts in SDXL.}
    \label{fig:subfig_c}
  \end{subfigure}
  \hfill
  \begin{subfigure}[b]{0.23\linewidth}
    \centering
    \includegraphics[width=\linewidth]{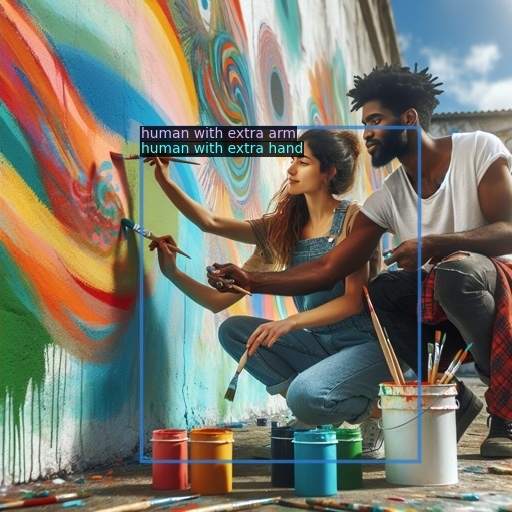}
    \caption{Global artifacts in DALLE-3.}
    \label{fig:subfig_d}
  \end{subfigure}
\vspace{-0.3cm}
  \caption{Example annotations from different generators in ~\ourdataset.}
\vspace{-0.6cm}
  \label{fig:example_anno}
\end{figure*}

\subsection{Human Artifacts in Text-to-Image Generation}
\label{sec:artifacts_in_gen}
Hand generation issues are particularly prevalent in T2I models, with most of the research on human artifacts focusing on this specific challenge. HandRefiner~\cite{LuXuZhWaTa23} and HandCraft~\cite{QinZhLiCa24} both use post-processing techniques with ControlNet~\cite{ZhangRaAg23}, conditioning on hand mesh models, shape, and depth masks to correct malformed hands. 
Hand-to-Diffusion~\cite{PelykhSiBo24} and HanDiffuser~\cite{NarasimhaswamyBhChDaMiHo24} adopt end-to-end approaches that integrate hand generation directly into the synthesis pipeline, conditioning on detailed hand-related parameters, focus solely on human image generation.
However, with arbitrary images, all these methods are unable to assess hand artifacts by severity. 
Recent works like SynArtifact~\cite{CaoYuLiLiSuLiZh24} and HumanCalibrator~\cite{WangMaWaLiWaTi24} introduce datasets with user-annotated captions and bounding boxes to train vision-language models for artifact detection. However, their effectiveness is constrained by the size and diversity of their data, and the limitations of vision-language models in detection tasks.
HumanRefiner~\cite{FangYaGuHaJiXuLiLi24} may be considered the closest prior work, as it also curates a human artifact dataset and trains a detection model. However, its reliance on SDXL limits the detector’s generalization, which is essential for robustly evaluating and benchmarking human generation quality across different T2I models, especially amid their rapid evolution. In contrast, ~\abbrourmodel~ achieves strong localization performance across diverse T2I models while also supporting diffusion model finetuning and artifact correction through inpainting.




\vspace{-0.3cm}
\section{Human Feedback for Human Artifacts}
\label{sec:method}

\subsection{Dataset Overview}
\label{sec:dataset_overview}


\subsubsection{Prompt and Image Collection Pipeline}
\label{sec:collect_pipeline}


We collect prompts from GPT-4~\cite{openai2023gpt4} and generate samples using four advanced text-to-image models: SDXL~\cite{PoEnLaBlDoMuPeRo24}, DALLE-2~\cite{openai2022dalle2}, DALLE-3~\cite{openai2023dalle3}, and Midjourney~\cite{midjourney2023}. 
For prompt generation, GPT-4 is queried to produce diverse prompts focused on humans in various backgrounds, activities, and framing styles. A total of 4,426 prompts are collected and divided into 90\% for the training set and 10\% for the validation set.
Using these prompts, images are generated via SDXL with different random seeds. A subset of the prompts is also used to generate samples from DALLE-2, DALLE-3, and Midjourney.
In total, 25,565 samples are collected from SDXL, 1,605 from DALLE-2, 2,419 from DALLE-3, and 7,965 from Midjourney.
Additionally, for diffusion models finetuning, we generate 3,260 prompts for the training set and 255 for the validation set.
More details are available in the appendix.

\subsubsection{Annotation Pipeline}
\label{sec:annotation_pipeline}

During the annotation process, annotators are tasked with identifying and marking artifacts using bounding boxes. Example annotations are shown in Fig.~\ref{fig:example_anno}. 
After visually inspecting sample images with human artifacts from all four generators, we empirically observe that many artifacts frequently occur at the extremities of human bodies, necessitating a finer granularity in category definition. Consequently, we divide the human body into six parts: \textit{face}, \textit{torso}, \textit{arm}, \textit{hand}, \textit{leg}, and \textit{feet}, and classify structural inconsistencies in the images as either local or global artifacts. 
For \textbf{local artifacts}, annotators label specific body parts, as defined above, that they consider poorly generated. For instance, in Fig.~\ref{fig:subfig_a}, the right hand has unusually long fingers with strange textures, and the arm and leg exhibit texture issues on the muscle. Similarly, in Fig.~\ref{fig:subfig_b}, the face of the left human figure and the hand of the right human figure appear distorted.
\textbf{Global artifacts} involve broader anatomical inconsistencies that affect the entire human figure, such as extra or missing one or more of the six defined parts (excluding cases of reasonable occlusion or those explicitly specified by textual prompts). These artifacts are grouped into a total of 12 classes. Due to the difficulty of pinpointing specific affected areas in such cases, annotators are instructed to mark the entire human figure. For example, in Fig.~\ref{fig:subfig_c}, the middle human figure is generated with an extra face, torso, and arm, while the right human figure appears to have an extra leg. Similarly, in Fig.~\ref{fig:subfig_d}, the highlighted human figure has three arms and hands.
Note that multiple class labels may share the same bounding box.

\begin{figure*}[t]
  \centering
   \includegraphics[width=\linewidth]{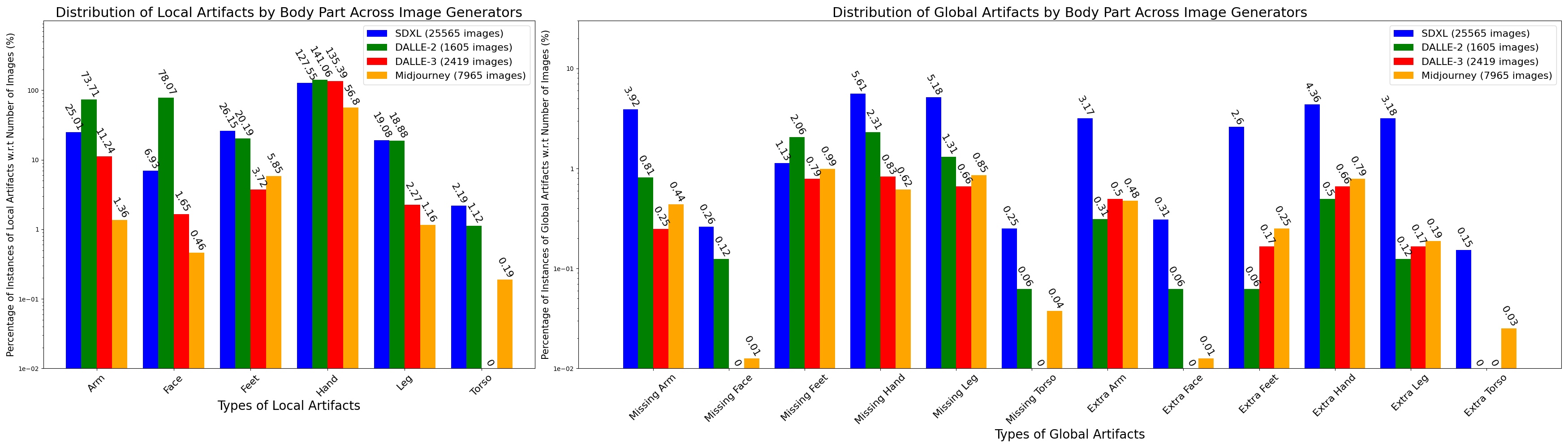}
    \vspace{-0.7cm}
   \caption{Distribution of local (left) and global (right) artifacts by categories across four different image generators in our ~\ourdataset. From the figure, several key observations emerge: Local artifacts are significantly more common than global artifacts, particularly in the hands, across all generators. Generator-wise, SDXL produces the highest number of global artifacts, while DALLE-2 produces the highest number of local artifacts. Both DALLE-3 and Midjourney exhibit stronger human structural coherence, with fewer overall artifacts. DALLE-3 shows a slight advantage in avoiding global artifacts, whereas Midjourney performs marginally better in avoiding local artifacts.}
   \label{fig:data_stat}
   \vspace{-0.5cm}
\end{figure*}

\subsubsection{Dataset Statistics}
\label{sec:data_statistics}

We summarize the data from our ~\ourdataset~ after collecting the prompts, generated images, and annotations. 
The training set includes 33,374 images with 67,685 local artifacts annotated with bounding boxes and 8,766 global artifacts. The validation set comprises 4,180 images with 7,528 local artifacts annotated with bounding boxes and 873 global artifacts.
Fig.~\ref{fig:data_stat} presents a detailed breakdown of the annotations for various artifact classes across the four image generators included in ~\abbrourdataset. 







\subsection{Model Overview}
\label{sec:model_overview}
To optimize performance while maintaining generalizability, we utilize the off-the-shelf ViTDet model~\cite{LiMaGiHe22} with EVA-02~\cite{FangSuWaHuWaCa24} as the backbone and Cascade-RCNN~\cite{CaiVa18} as the detection head. Due to the differing nature of the local and global artifacts we defined, we train two separate models: \textbf{\ourlocalmodel~(\abbrourlocalmodel)} for local artifacts and \textbf{\ourglobalmodel~(\abbrourglobalmodel)} for global artifacts.
\abbrourlocalmodel~is a standard object detection model trained on images annotated with six types of local artifacts, each with corresponding bounding boxes, while \abbrourglobalmodel~ is designed to handle a multi-label object detection task. To accommodate this, we use binary cross-entropy loss to independently model the 12 categories. 
To improve robustness and reduce false positives on higher-quality generated body parts, we incorporate a broad range of real human datasets for different tasks, including human parsing~\cite{LiZhWeLaLiFe17}, human instance segmentation~\cite{ZhangLiDoRoCaHaYaHuHu19}, human detection~\cite{ShiLiJi13,ShaoZhLiXiYuZhSu18}, and human interaction recognition~\cite{TanisikZaIk2016facial}, as well as the COCO dataset~\cite{LinMaBeHaPeRaDoZi14}. These images are included in the training process with empty annotations, ensuring the models learn to differentiate between realistic human body parts and those with artifacts.
More details regarding model and training design choices are available in the appendix. 

\subsection{Finetuning Diffusion Models with Guidance from ~\abbrourmodel}
\label{sec:finetuning_model}

\begin{table*}[t!]
\centering
\vspace{-0.2cm}
\caption{Local Detection Results. Each cell shows the AP50 score / number of instances in the val set. }
\vspace{-0.2cm}
\begin{tabular}{l|cccccc|r}
\toprule
Domain & Face & Torso & Arm & Hand & Leg & Feet & Average \\
\midrule
SDXL   & 26.0 / 238 & 26.8 / 39 & 25.4 / 672 & 80.1 / 3493 & 28.8 / 477 & 50.0 / 672 & 39.5  \\
DALLE-2 & 86.6 / 145 & 100.0 / 1 & 52.7 / 131 & 88.9 / 228  & 39.4 / 36  & 56.8 / 42  & 70.7  \\
DALLE-3 & 2.7 / 7    & - / 0     & 8.8 / 46   & 48.2 / 563  & 7.4 / 10   & 20.0 / 27  & 17.4  \\
Midjourney     & 5.5 / 4    & - / 0     & 9.8 / 22   & 54.4 / 586  & 14.0 / 13  & 27.8 / 51  & 22.3  \\
\midrule
ALL    & 53.4 / 398 & 27.6 / 43 & 28.9 / 875 & 74.5 / 4875 & 28.1 / 539 & 47.3 / 798 & 43.3  \\
\bottomrule
\end{tabular}
\label{tab:local_quant}
\end{table*}

\vspace{-0.2cm}

To demonstrate the application of \abbrourmodel, we employ LoRA\cite{HuShWaAlLiWaWaCh22} to finetune SDXL~\cite{PoEnLaBlDoMuPeRo24} on newly generated images with predictions from \abbrourmodel. 
Specifically, we generate images from the training set of the prompts for diffusion model finetuning, as described in Sec.~\ref{sec:collect_pipeline}.
Once the images are generated, we apply ~\abbrourmodel~ to detect local and global artifacts. We then select the top $k\%$ of predicted bounding boxes from each category to identify artifacts.
During finetuning, we use the original prompt as the caption for generated images and prepend a special identifier, e.g., “weird [ARTIFACT-TYPE]”, for each identified artifact. 
At inference, these identifiers serve as negative prompts to prevent artifact generation and are applied to both the original and finetuned models for evaluation.
Additionally, we incorporate images from COCO~\cite{LinMaBeHaPeRaDoZi14} along with corresponding prompts as regularization to prevent the model from deviating aggressively from the original weights while improving human structural coherence. 



\section{Experiments}
\label{sec:experiments}



The performance and robustness of our ~\abbrourmodel~ are assessed on both in-domain and out-of-domain data. In-domain data includes the images from the validation set of ~\abbrourdataset, while out-of-domain data includes images generated by Stable Diffusion 1.4~\cite{RoBlaLoEsOm22}, PixArt-\(\Sigma\)~\cite{ChenGeXiWuYaReWaLuLuLi24}, FLUX.1-dev~\cite{flux2024}, Sana~\cite{XieChChCaTaLiZhLiZhLuHa24}, and the real human dataset 300W~\cite{SagonasTzZaPa13}. More details are provided in the appendix.


\begin{table*}[t]
\centering
\vspace{-0.2cm}
\caption{Global Detection Results. Each cell shows the AP50 score / number of instances in the val set.}
\vspace{-0.2cm}
\setlength{\tabcolsep}{8pt} 
\renewcommand{\arraystretch}{1.2} 
\begin{adjustbox}{width=\textwidth}
\begin{tabular}{l|cccccc|cccccc|r} 
\toprule
\multirow{2}{*}{Domain}  & \multicolumn{6}{c|}{Missing Artifacts} & \multicolumn{6}{c|}{Extra Artifacts} & \multirow{2}{*}{Average} \\
\cmidrule(lr){2-7} \cmidrule(lr){8-13} 
& Face & Torso & Arm & Hand & Leg & Feet & 
Face & Torso & Arm & Hand & Leg & Feet & \\
\midrule
SDXL   & 6.7 / 6   & 7.7 / 6   & 0.0 / 4   & 0.0 / 2   & 23.5 / 93  & 39.9 / 77  & 29.9 / 123 & 40.1 / 112 & 35.2 / 113 & 51.9 / 71  & 41.5 / 142 & 52.2 / 55 & 27.4  \\
DALLE-2 & - / 0     & - / 0     & - / 0     & - / 0     & 0.0 / 1    & - / 0      & 0.0 / 1    & - / 0      & 14.3 / 2   & - / 0      & 16.7 / 1   & 0.0 / 1  & 6.2    \\
DALLE-3 & - / 0     & - / 0     & 0.0 / 1   & - / 0     & - / 0      & 33.3 / 7   & 33.3 / 1   & 33.3 / 9   & 11.3 / 2   & - / 0      & 6.8 / 3    & - / 0    & 19.7   \\
Midjourney     & - / 0     & - / 0     & - / 0     & - / 0     & 1.9 / 5    & 6.3 / 2    & 3.4 / 6    & 20.5 / 4   & 24.0 / 7   & 0.0 / 2    & 22.6 / 8   & 8.4 / 3  & 10.9   \\
\midrule
ALL    & 5.2 / 6   & 2.4 / 6   & 0.0 / 5   & 0.0 / 2   & 21.0 / 101 & 38.2 / 86  & 26.9 / 133 & 38.5 / 125 & 32.0 / 123 & 42.9 / 73  & 37.5 / 154 & 41.7 / 59 & 23.9  \\
\bottomrule
\end{tabular}
\end{adjustbox}
\label{tab:global_quant}
\vspace{-0.5cm}
\end{table*}

\subsection{Quantitative Results}
\label{sec:det_exp_performance}

\noindent\textbf{\ourlocalmodel}:
To evaluate the model’s performance, we use the AP50 metric. The results for \abbrourlocalmodel~are shown in Tab.~\ref{tab:local_quant}. We observe that generators with a higher frequency of human artifacts, such as SDXL and DALLE-2, achieve higher AP50 scores compared to more advanced generators like DALLE-3 and Midjourney. Furthermore, categories with more frequent artifacts, such as \textit{hand} and \textit{feet}, achieve higher AP50 scores compared to less common categories like \textit{torso} and \textit{leg}.

\noindent\textbf{\ourglobalmodel}:
The results for ~\abbrourglobalmodel~ are provided in Tab.~\ref{tab:global_quant}. Compared to \abbrourlocalmodel, the limited number of global artifact annotations leads to lower overall performance. Among the generators, \abbrourglobalmodel~ shows comparatively better performance on SDXL, which has the majority of global artifact annotations. In contrast, DALLE-2, despite strong performance in detecting local artifacts and abundant local annotations, performs worse in detecting global artifacts, likely due to fewer annotations available for this task. Further discussion and analysis of these results can be found in Sec.~\ref{sec:det_exp_discussion}.

\begin{figure}[t]
  \centering
   \includegraphics[width=\linewidth]{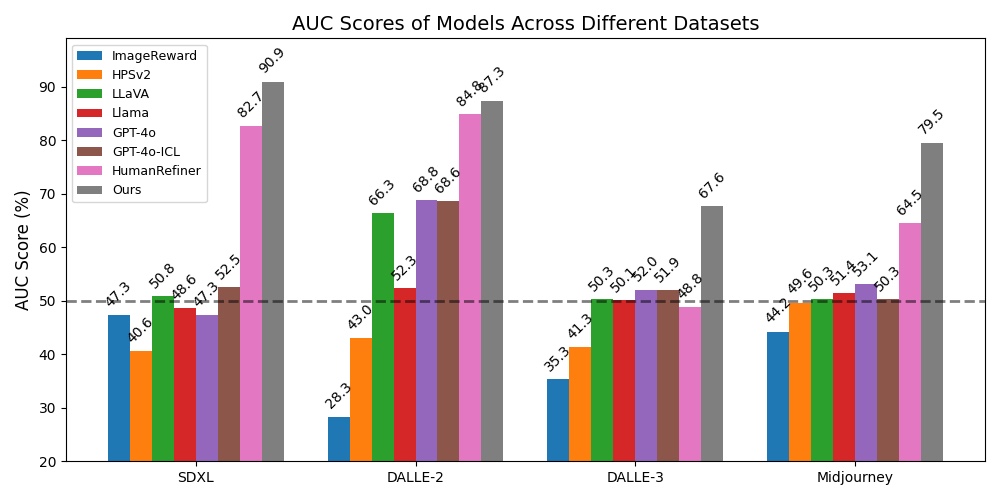}
   \vspace{-0.7cm}
   \caption{Comparison of the AUC scores of ~\abbrourmodel~ against baseline methods. ICL represents in-context learning. }
   \label{fig:baseline_compare}
   \vspace{-0.7cm}
\end{figure}

\begin{figure*}[t] 
  \centering
  \begin{subfigure}[b]{0.24\linewidth}
    \centering
    \includegraphics[width=\linewidth]{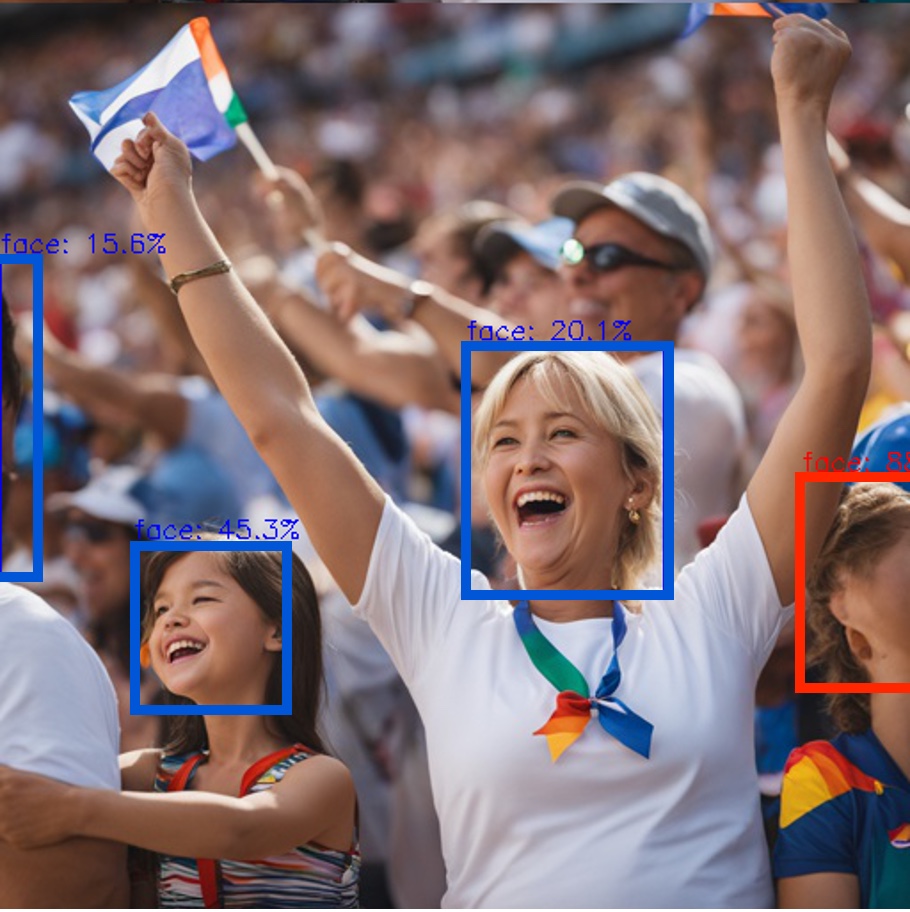}
    \caption{FP face in SDXL}
    \label{fig:id_example_s_2}
  \end{subfigure}
  \hfill
  \begin{subfigure}[b]{0.24\linewidth}
    \centering
    \includegraphics[width=\linewidth]{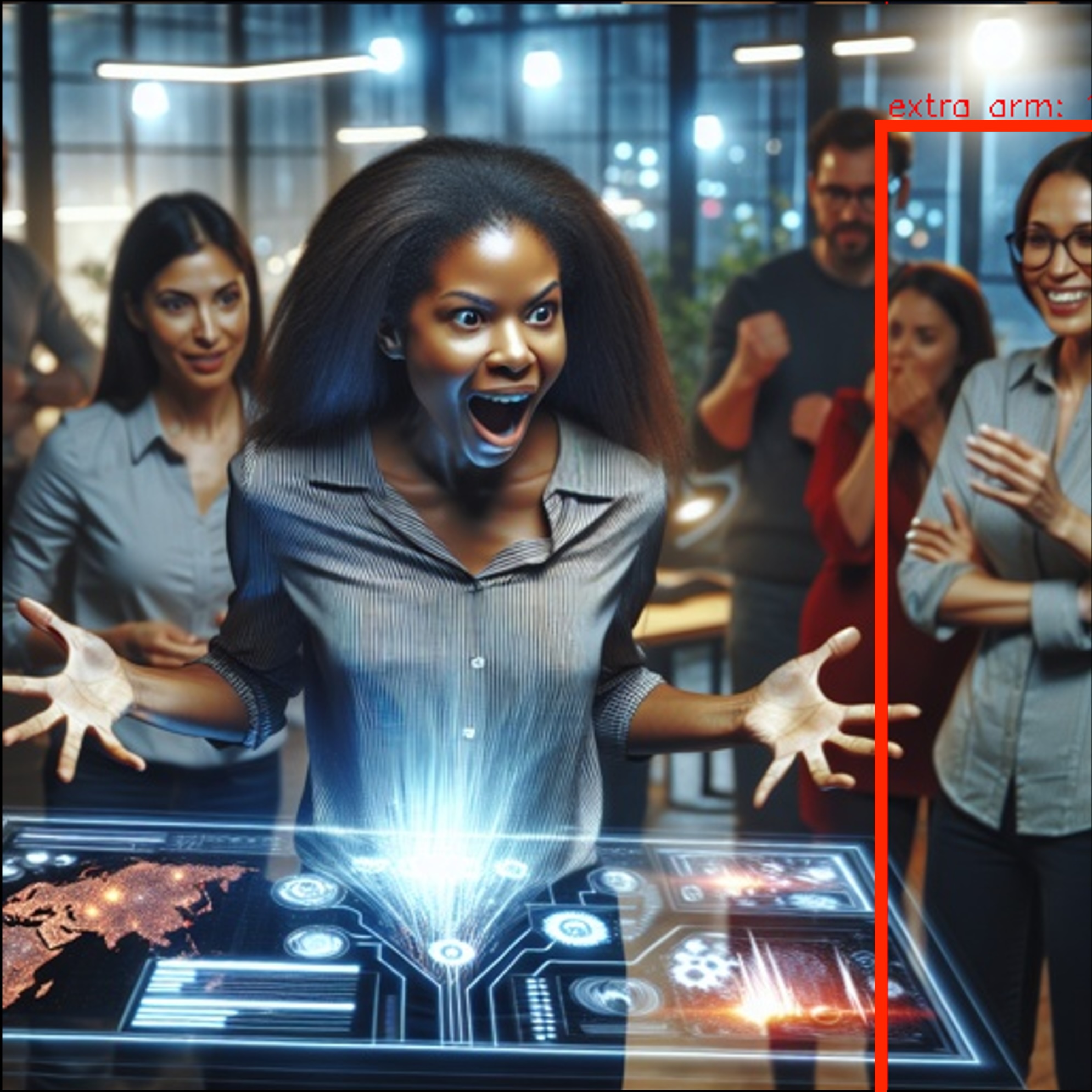}
    \caption{FP extra arm in DALLE-3}
    \label{fig:id_example_d3_2}
  \end{subfigure}
  \hfill
  \begin{subfigure}[b]{0.24\linewidth}
    \centering
    \includegraphics[width=\linewidth]{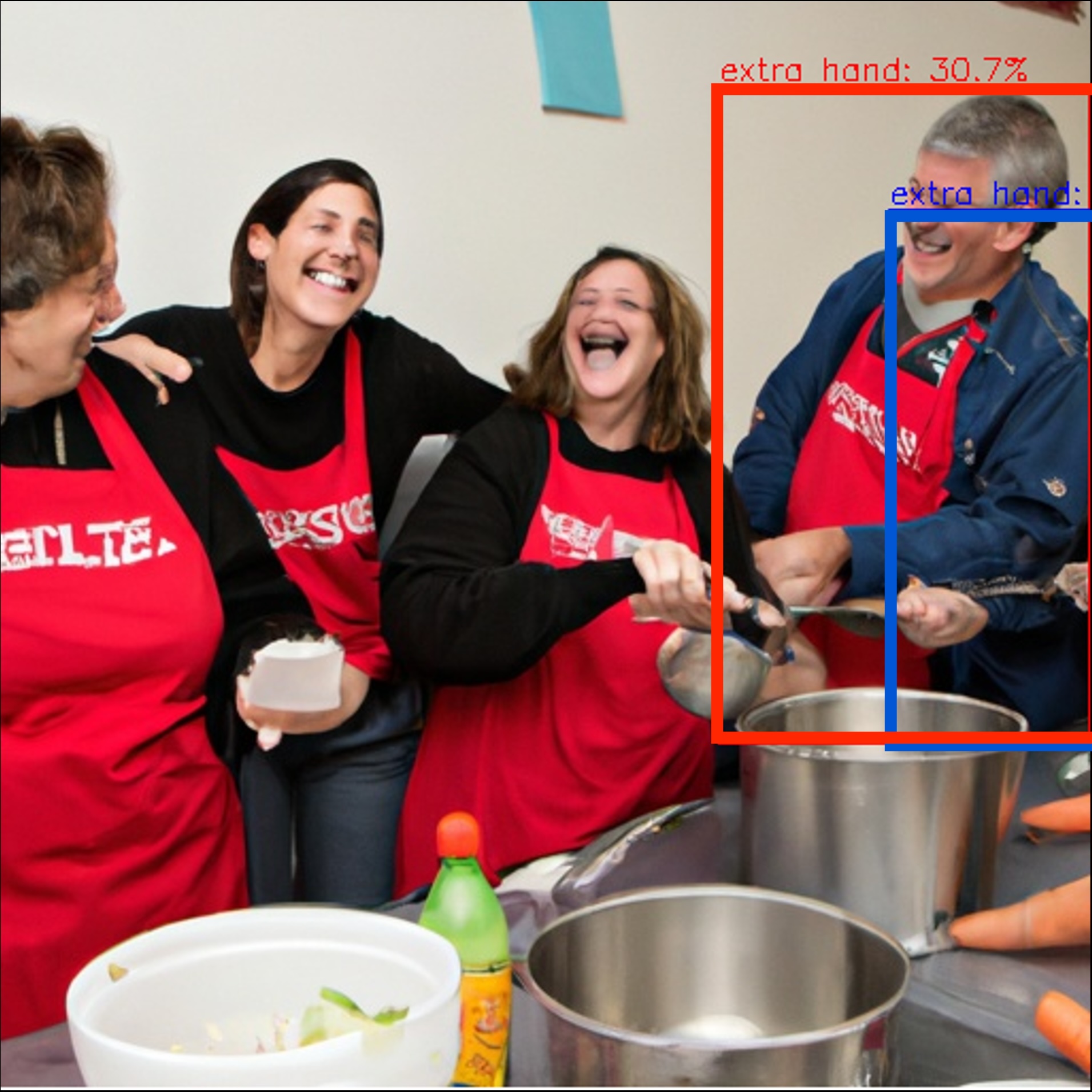}
    \caption{FP extra hand in DALLE-2}
    \label{fig:id_example_d2_3}
  \end{subfigure}
  \hfill
  \begin{subfigure}[b]{0.24\linewidth}
    \centering
    \includegraphics[width=\linewidth]{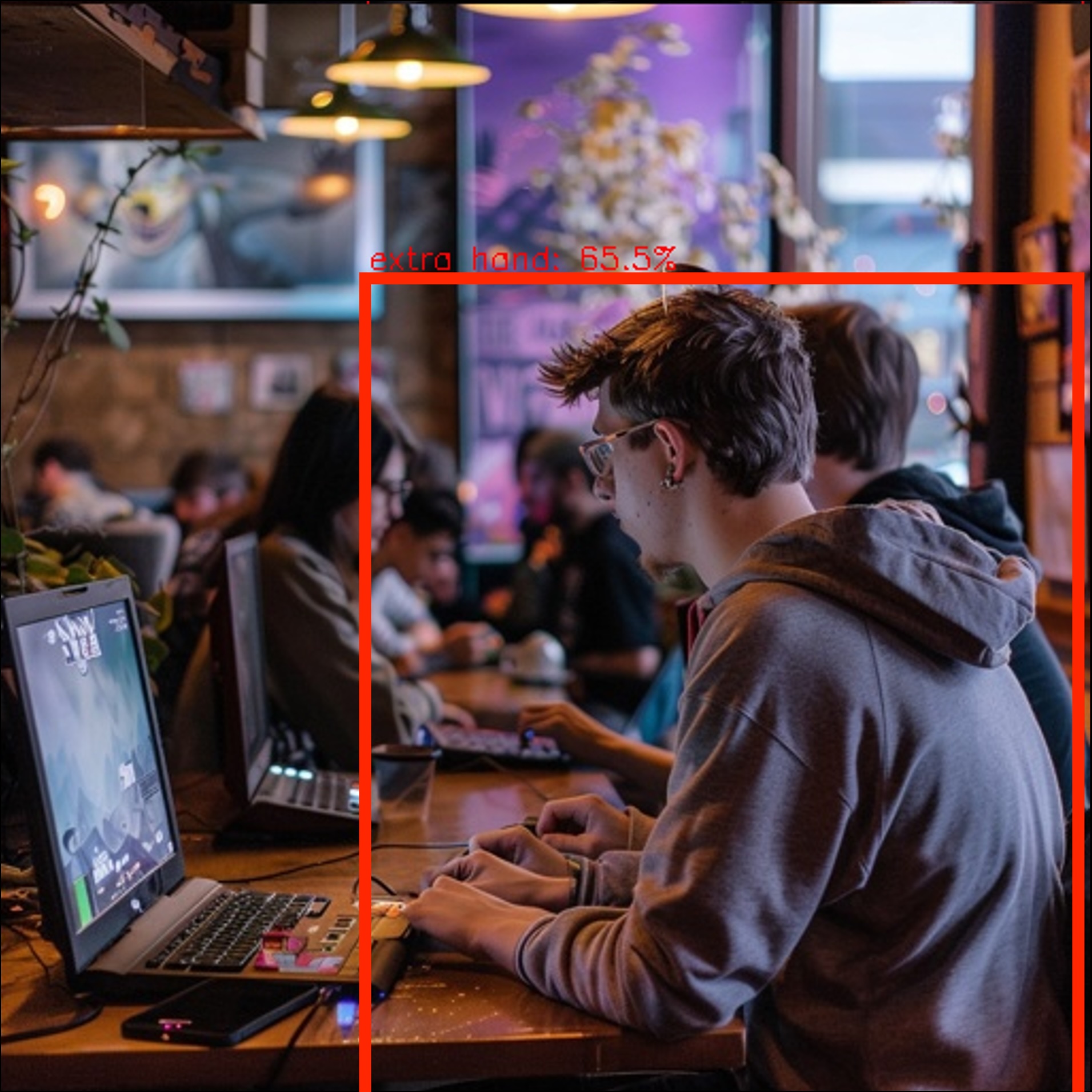}
    \caption{FP extra hand in Midjourney}
    \label{fig:id_example_mj_4}
  \end{subfigure}
  \vspace{-0.3cm}
  \caption{Examples of predictions from our ~\abbrourmodel~ considered mistakes during evaluation on SDXL (a), DALLE-3 (b), DALLE-2 (c), and Midjourney (d). 
  FP: false positive, FN: false negative.
  Red bounding boxes represent the detected artifact with top prediction scores, blue bounding boxes represent other detected bounding boxes with the same class label.
  }
  \vspace{-0.1cm}
  \label{fig:id_example}
\end{figure*}

\begin{figure*}[t] 
  \centering
  \begin{subfigure}[b]{0.24\linewidth}
    \centering
    \includegraphics[width=\linewidth]{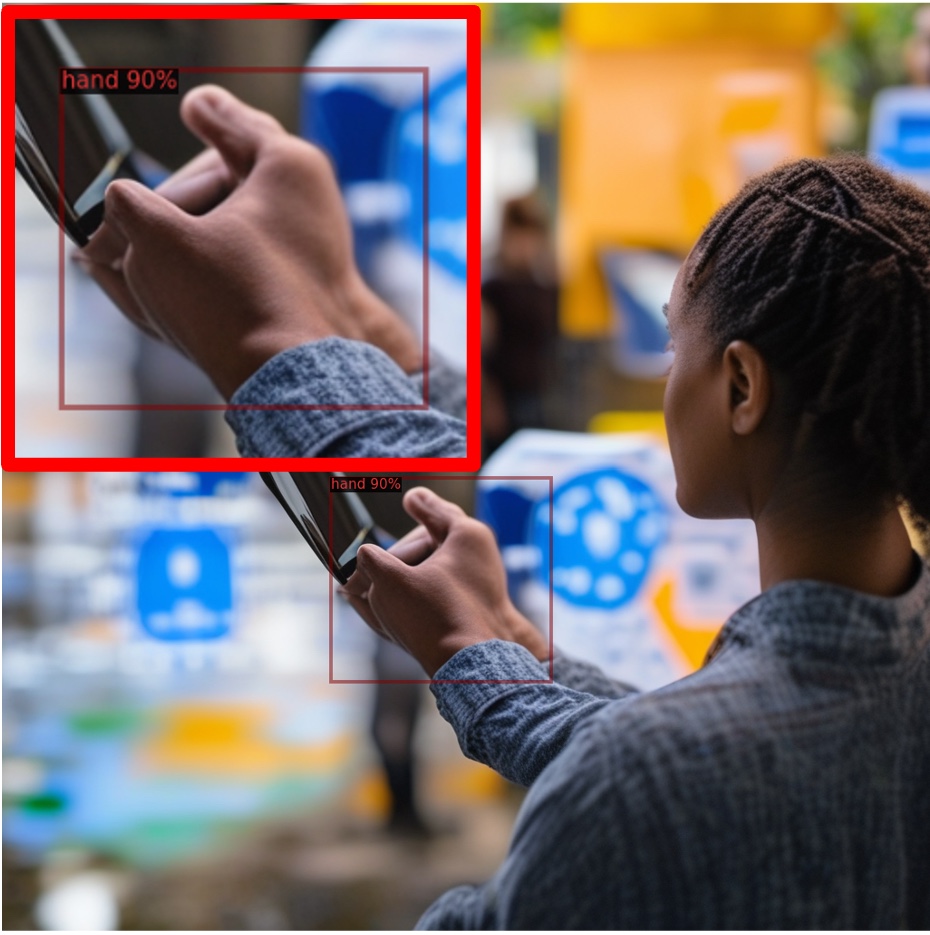}
    \caption{Weird hand (90\%)}
    \label{fig:ood_example_p_a}
  \end{subfigure}
  \hfill
  \begin{subfigure}[b]{0.24\linewidth}
    \centering
    \includegraphics[width=\linewidth]{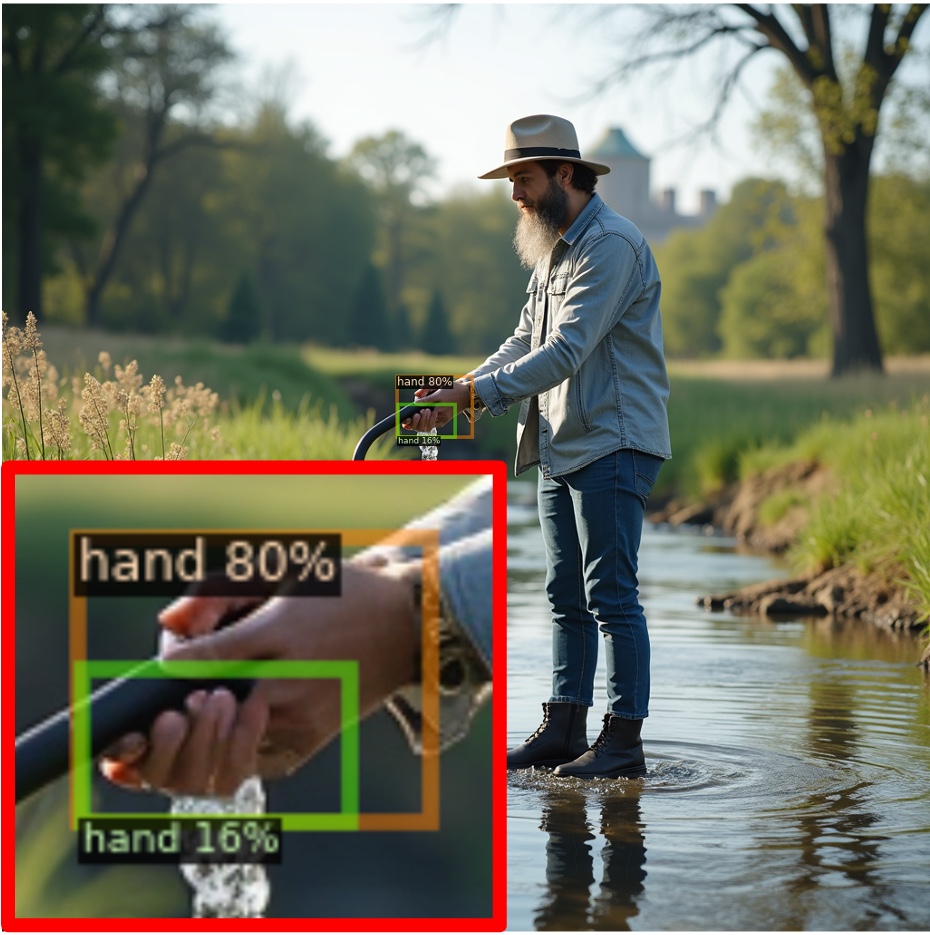}
    \caption{Weird hand (80\%)}
    \label{fig:ood_example_f_a}
  \end{subfigure}
  \hfill
  \begin{subfigure}[b]{0.24\linewidth}
    \centering
    \includegraphics[width=\linewidth]{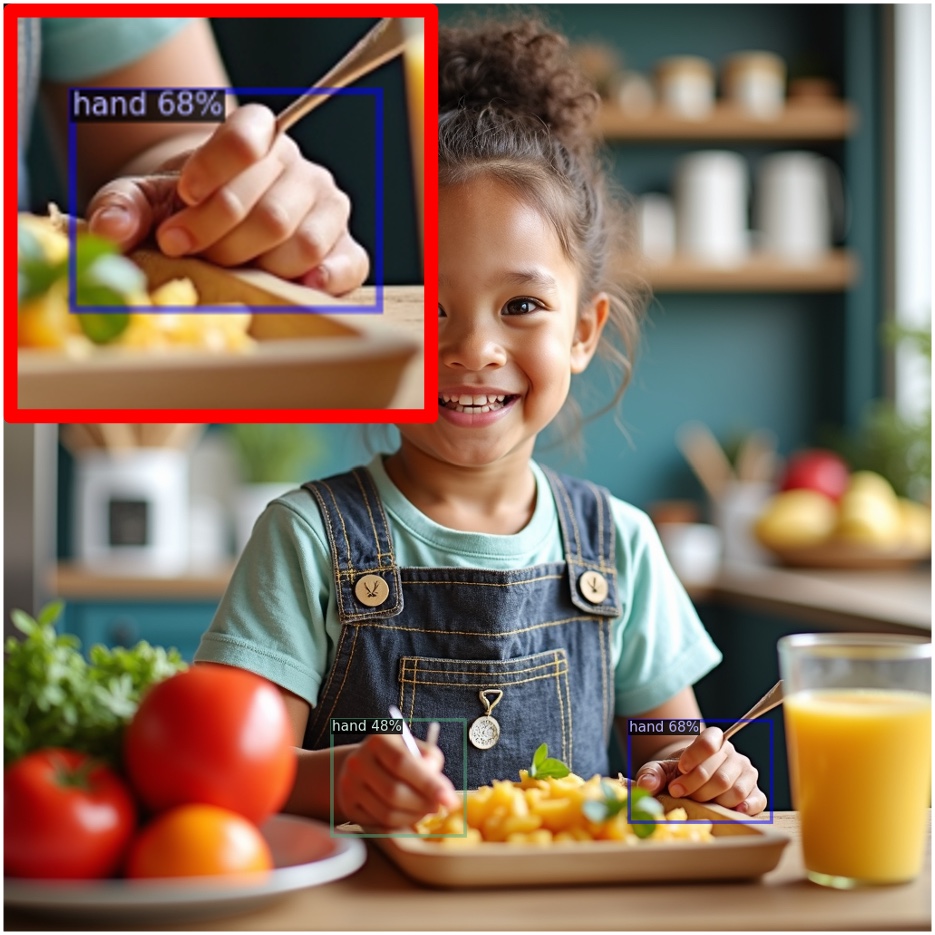}
    \caption{Weird hand (68\%)}
    \label{fig:ood_example_f_b}
  \end{subfigure}
  \hfill
  \begin{subfigure}[b]{0.24\linewidth}
    \centering
    \includegraphics[width=\linewidth]{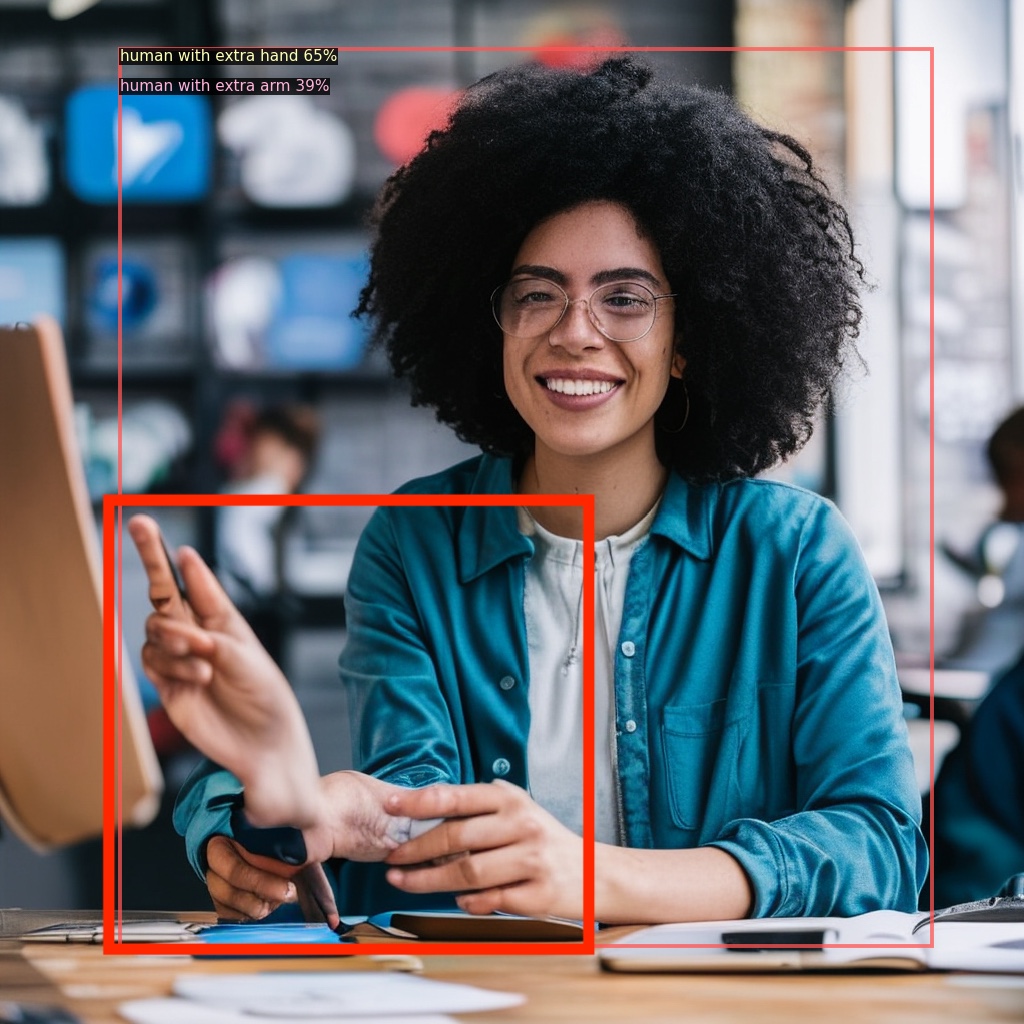}
    \caption{Extra hand (65\%) \& arm (39\%)}
    \label{fig:ood_example_sn_1}
  \end{subfigure}
    \vspace{-0.3cm}

  \caption{(a): Top predictions on PixArt-\(\Sigma\). (b), (c): Top predictions on FLUX.1-dev. (d): Top predictions on Sana.}
  \label{fig:ood_example}
  \vspace{-0.5cm}
\end{figure*}

\begin{figure}[t]  
  \centering
  \begin{subfigure}[b]{0.41\linewidth}  
    \centering
    \includegraphics[width=\linewidth]{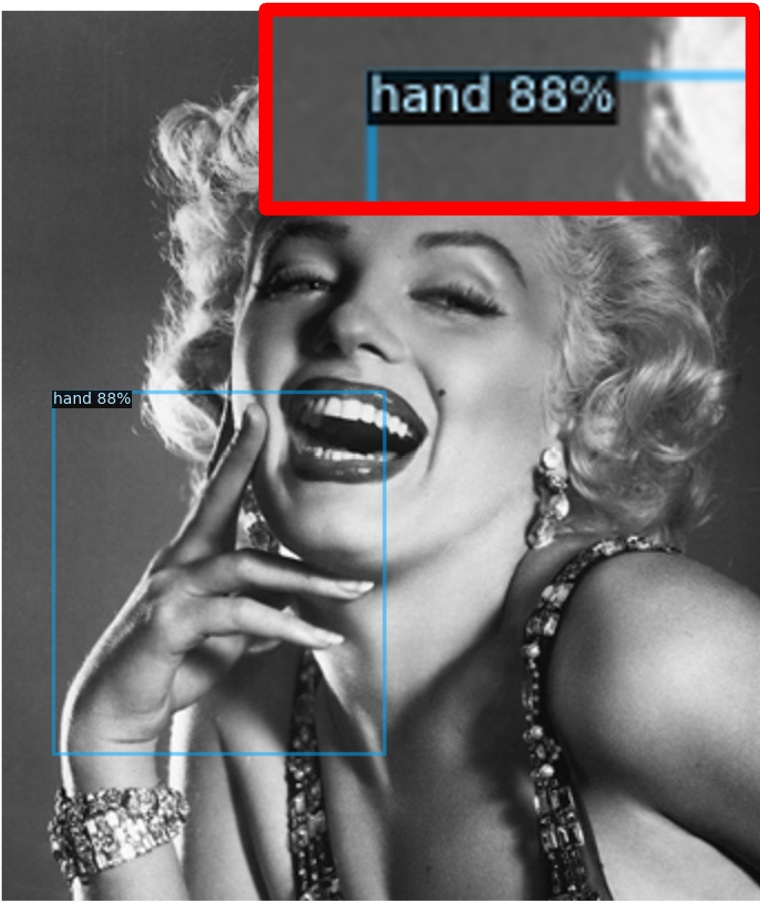}
    \caption{Weird hand (88\%)}
    \label{fig:real1}
  \end{subfigure}
  \hfill
  \begin{subfigure}[b]{0.52\linewidth}  
    \centering
    \includegraphics[width=\linewidth]{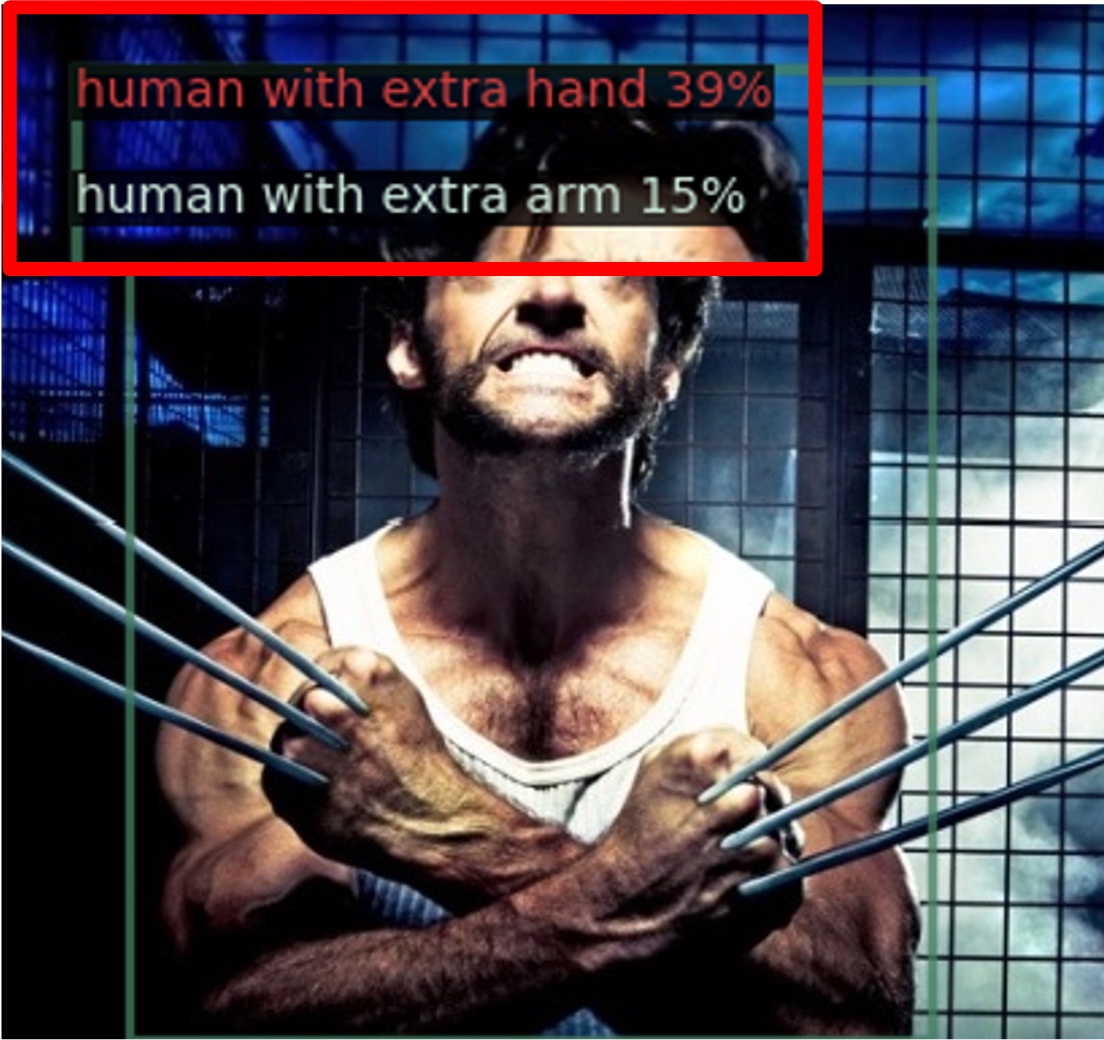}
    \caption{Extra hand (39\%) \& arm (15\%)}
    \label{fig:real3}
  \end{subfigure}
\vspace{-0.2cm}
  \caption{Top predictions of ~\abbrourmodel~ on real images from 300W.}
\vspace{-0.5cm}
  \label{fig:real_example}
\end{figure}

\noindent\textbf{Comparison to Baselines}: We compare \abbrourmodel~ against HumanRefiner~\cite{FangYaGuHaJiXuLiLi24}, as well as the publicly available state-of-the-art methods in aesthetics evaluation, specifically HPS-v2~\cite{WuHaSuChZhZhLi23} and ImageReward~\cite{XuLiWuToLiDiTaDo23} due to the presumed correlation between aesthetic quality and human structural coherence. 
Additionally, we test three competitive large vision-language models: GPT-4o~\cite{openai2024gpt4o}, LLaVA-1.6-vicuna-13B~\cite{liu2024llavanext}, and Llama-3.2-11B~\cite{llama32}.

Given the different artifact definitions in HumanRefiner and the lack of bounding box predictions in other baselines, we reframe our detection task as a classification problem. Images in our validation set annotated with any global or local artifacts are labeled as positive, while the rest are labeled negative, resulting in 2,917 positive and 1,263 negative samples. For testing, HPS-v2 and ImageReward produce scalar scores representing preference, while GPT-4o, LLaVA-1.6, and Llama-3.2 are instructed to rate human-related artifacts on a scale from 1 (no artifact) to 5 (catastrophic artifact). We also test GPT-4o with in-context learning, providing it with four sample images with their human artifacts highlighted before prompting for predictions.
For ~\abbrourmodel~ and HumanRefiner, we use the highest score among all bounding boxes marked as human artifacts as the image’s final score.

We evaluate models using AUC scores to measure their effectiveness in ranking human artifact severity. As shown in Fig.~\ref{fig:baseline_compare}, aesthetics-based methods perform poorly, with AUC scores below 50\%, suggesting that aesthetic evaluation is insufficient for detecting human artifacts. Similarly, vision-language models like GPT-4o and LLaVA perform relatively well on DALLE-2, where local artifacts are more prevalent (Fig.~\ref{fig:data_stat}, Tab.~\ref{tab:local_quant}), but their performance drops to near 50\% on advanced generators, indicating limited capacity in human artifact identification. Notably, in-context learning does not improve artifact identification, highlighting the difficulty of this task.
HumanRefiner performs well on SDXL and approaches our accuracy, aligning with its training data. However, its performance deteriorates significantly on other generators, especially the more advanced Midjourney and DALLE-3, due to poor generalization. In contrast, \abbrourmodel~maintains strong performance across all datasets benefitting from the diverse training data and strong detection model, substantially outperforming all baselines. This result demonstrates its ability to detect human-related artifacts consistently. This ability is essential to truly benchmark human-generation quality among diverse T2I models. 

\subsection{Discussion}
\label{sec:det_exp_discussion}

\noindent\textbf{In-Domain Performance}:
To analyze failure cases, we examine high-confidence false positives, as visualized in Fig.~\ref{fig:id_example}. We find that some errors appear to stem from occasional oversight by annotators. For example, the distorted face in Fig.~\ref{fig:id_example_s_2} and the extra arms in Fig.~\ref{fig:id_example_d3_2} are overlooked, likely because they appear in a less prominent area of the image.
We also notice subjective ambiguity in some regions, where their severity is difficult for annotators to assess. 
For instance, \abbrourmodel~ detect extra hands in Fig.~\ref{fig:id_example_d2_3} and Fig.~\ref{fig:id_example_mj_4}, which are missed by annotators. We consider these as corner cases, as the seemingly extra hands could also be interpreted as belonging to other human figures, albeit with very high uncertainty.
Overall, visual inspection verifies our models’ robustness, demonstrating that they can still make reasonable predictions under challenging conditions.

\begin{figure*}[h!] 
  \centering
  \begin{subfigure}[b]{0.22\linewidth}
    \centering
    \includegraphics[width=\linewidth]{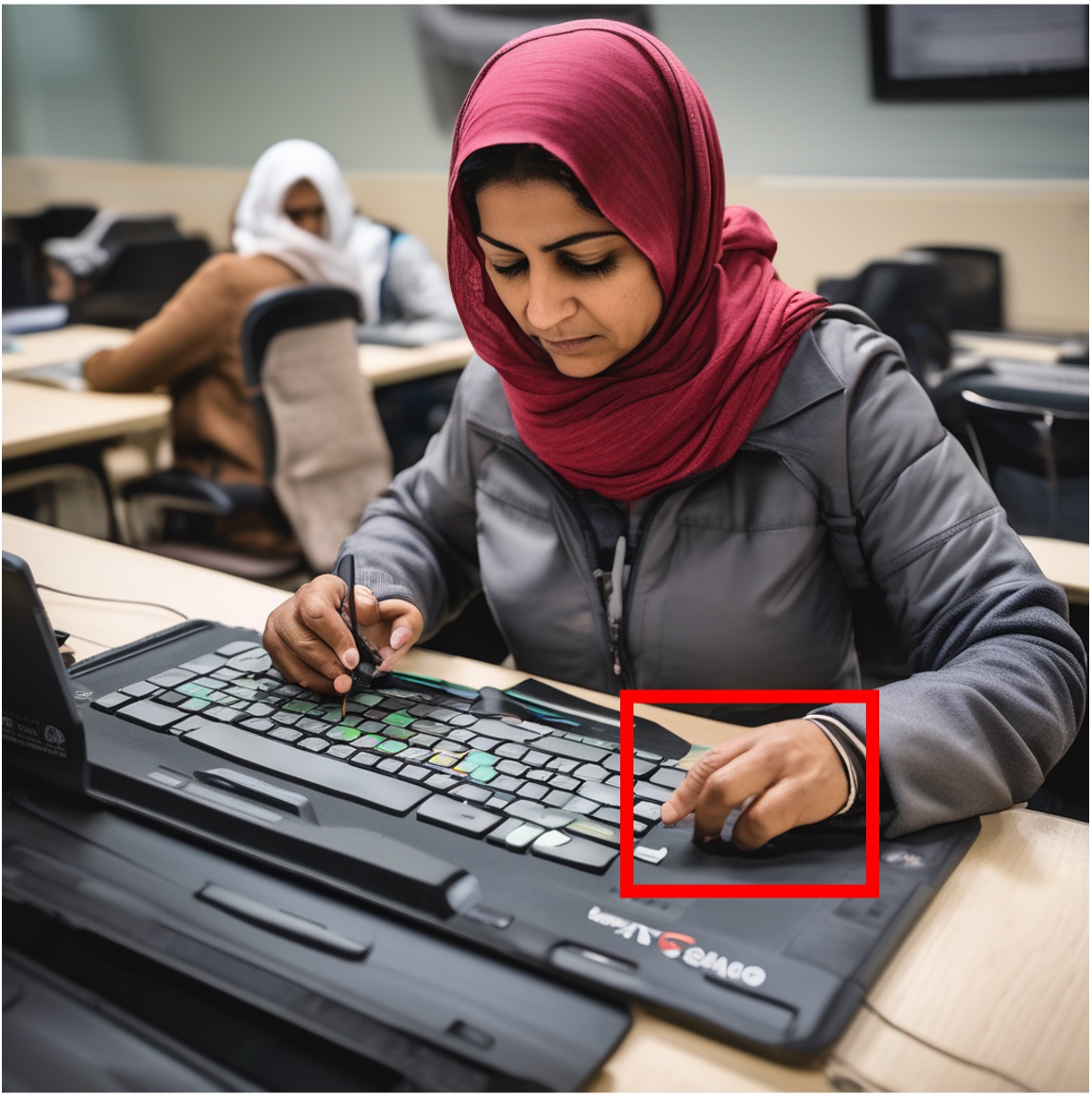}
    \label{fig:ft_example_b1}
  \end{subfigure}
  \hfill
  \begin{subfigure}[b]{0.22\linewidth}
    \centering
    \includegraphics[width=\linewidth]{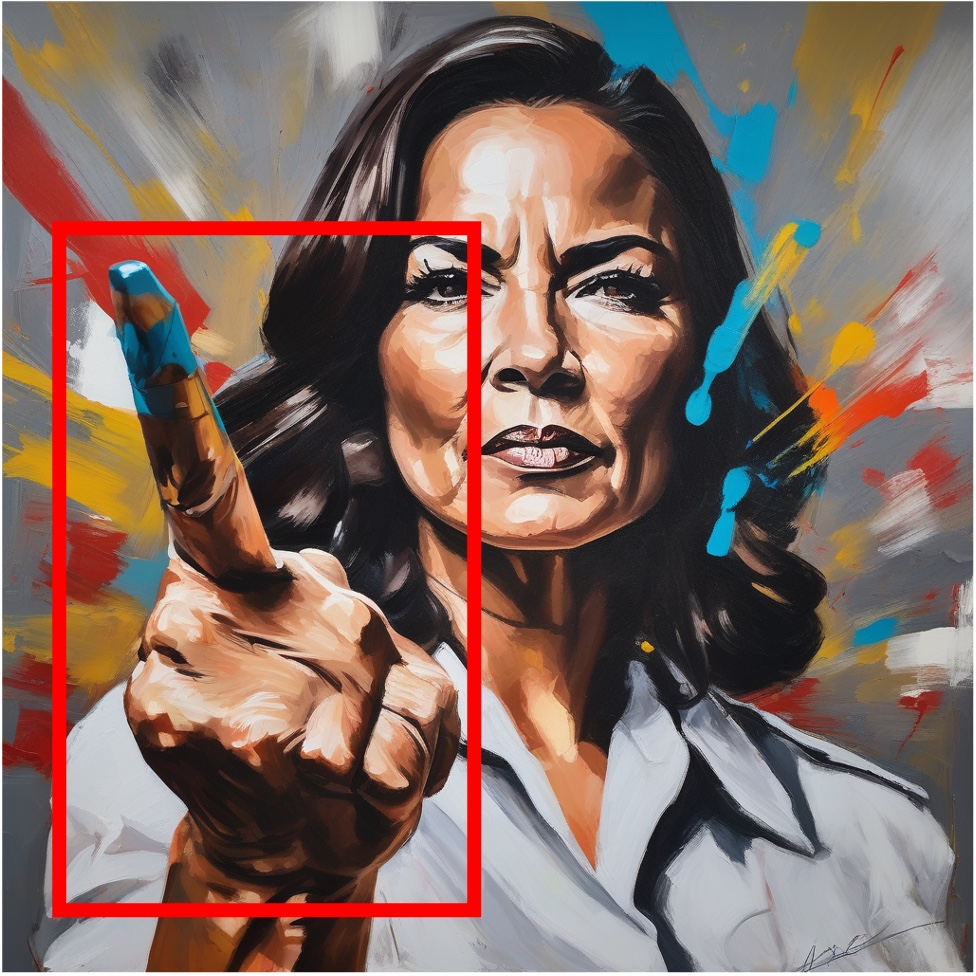}
    \label{fig:ft_example_b2}
  \end{subfigure}
  \hfill
  \begin{subfigure}[b]{0.22\linewidth}
    \centering
    \includegraphics[width=\linewidth]{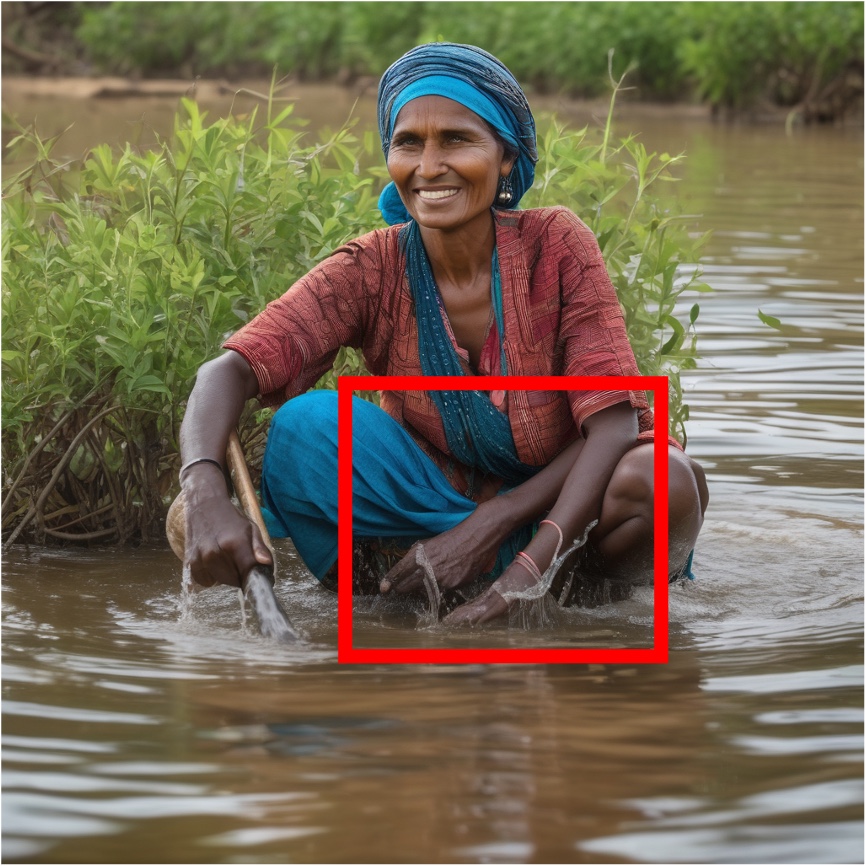}
    \label{fig:ft_example_b3}
  \end{subfigure}
  \hfill
  \begin{subfigure}[b]{0.22\linewidth}
    \centering
    \includegraphics[width=\linewidth]{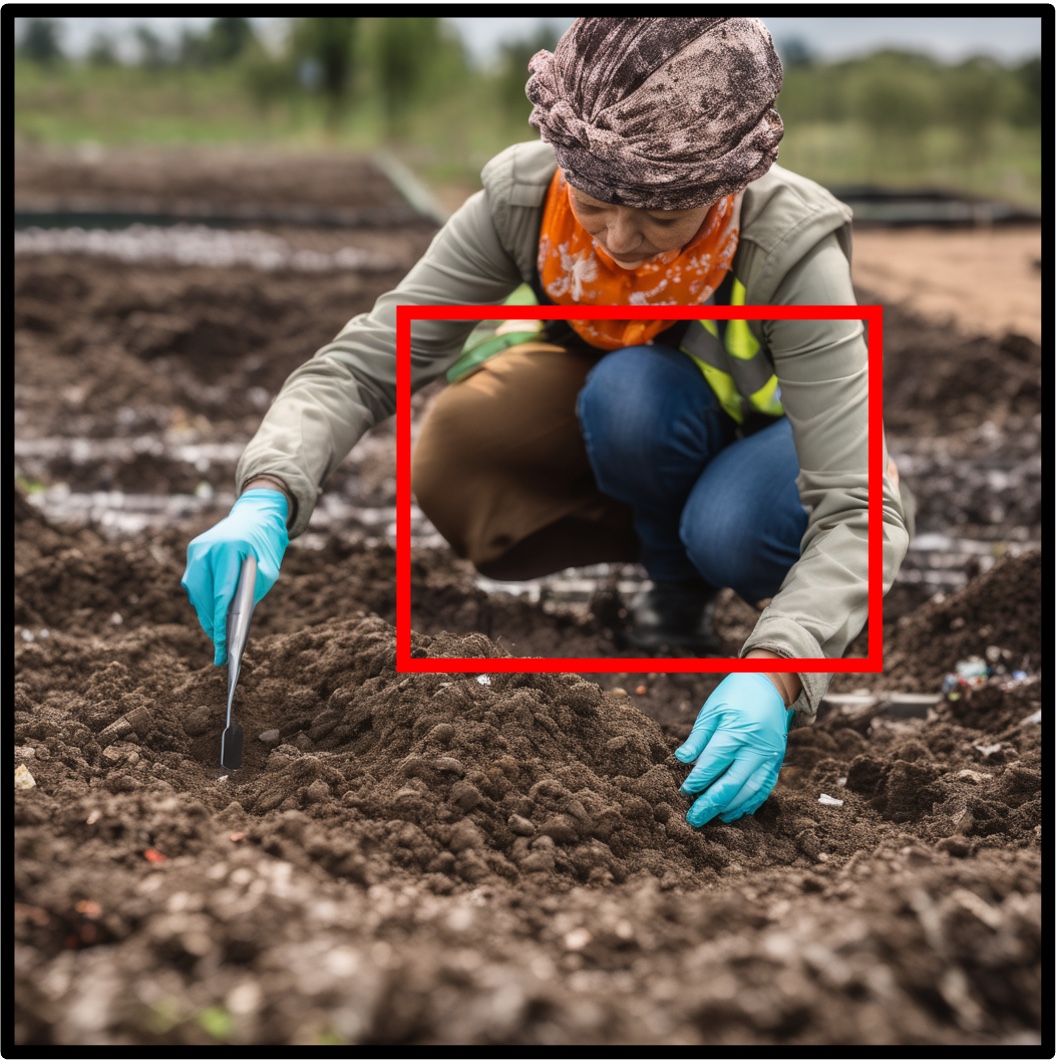}
    \label{fig:ft_example_b4}
  \end{subfigure}

  \begin{subfigure}[b]{0.22\linewidth}
    \centering
    \includegraphics[width=\linewidth]{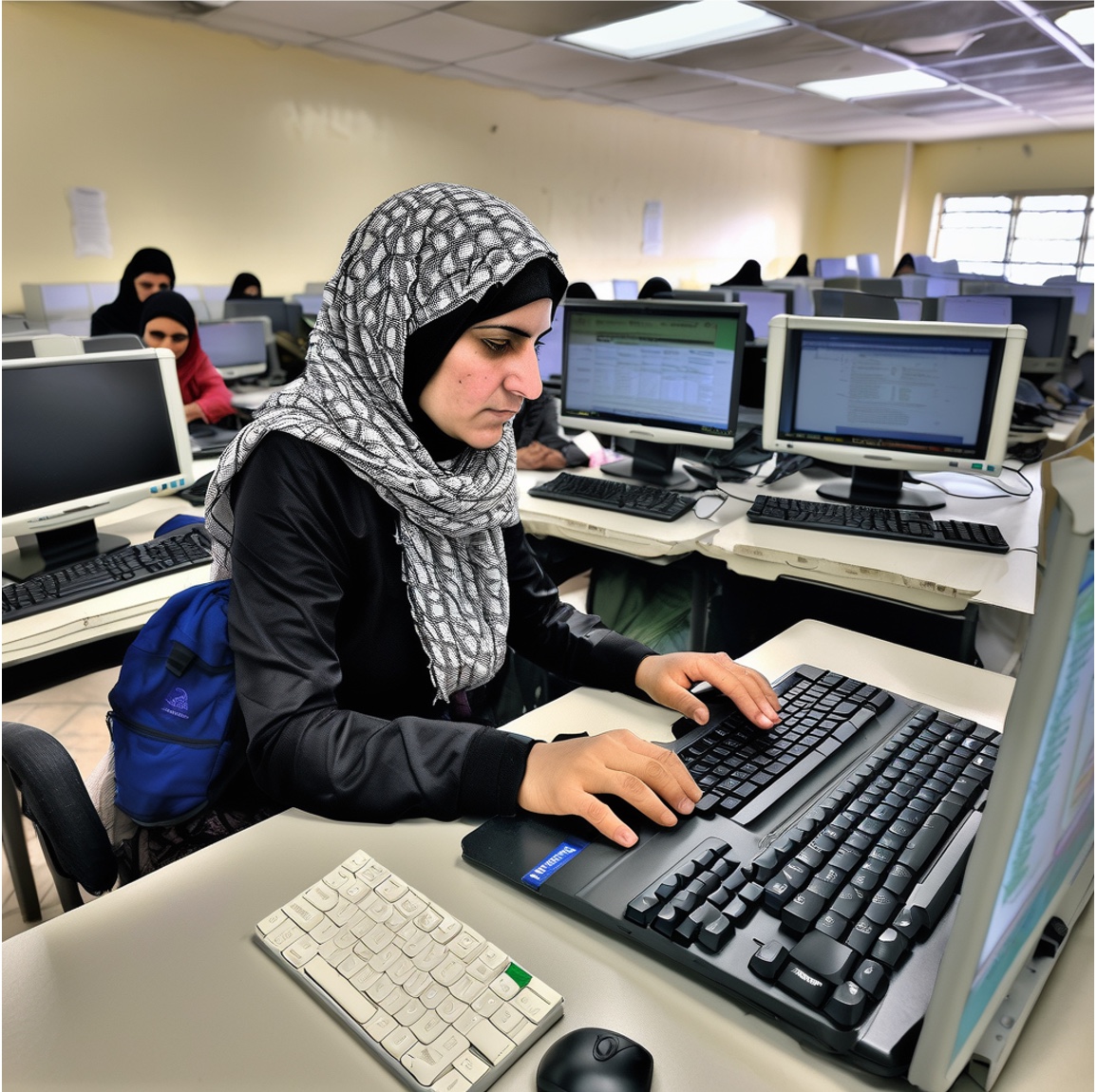}
    \label{fig:ft_example_o1}
  \end{subfigure}
  \hfill
  \begin{subfigure}[b]{0.22\linewidth}
    \centering
    \includegraphics[width=\linewidth]{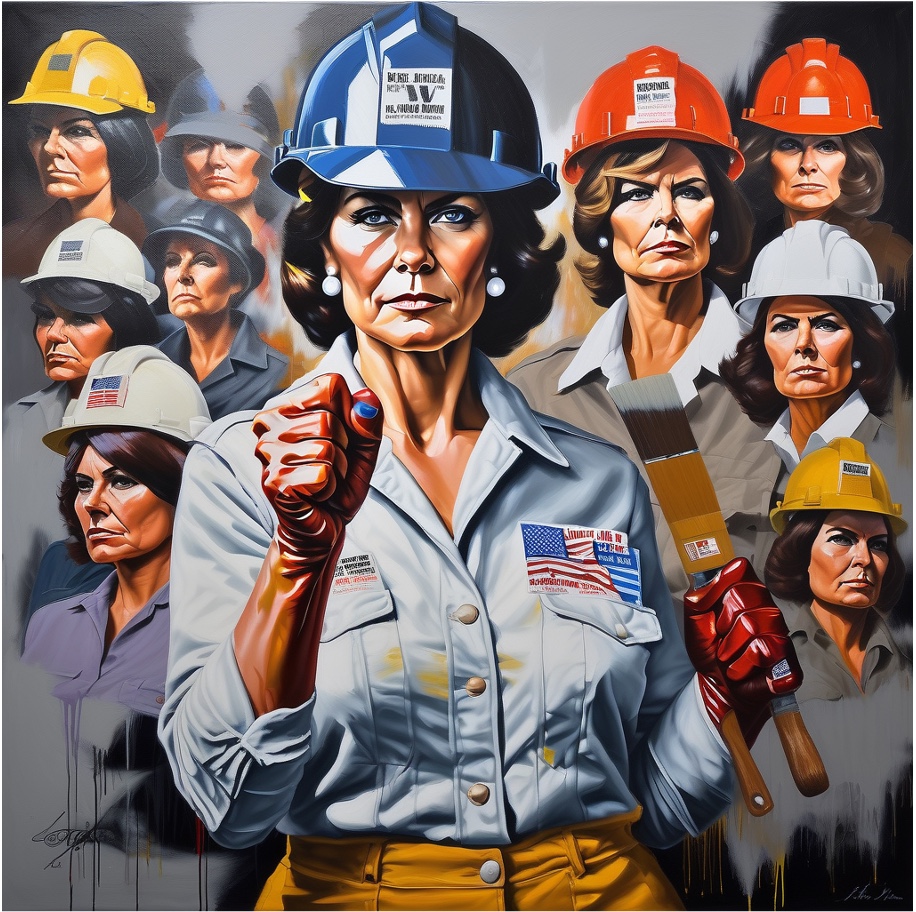}
    \label{fig:ft_example_o2}
  \end{subfigure}
  \hfill
  \begin{subfigure}[b]{0.22\linewidth}
    \centering
    \includegraphics[width=\linewidth]{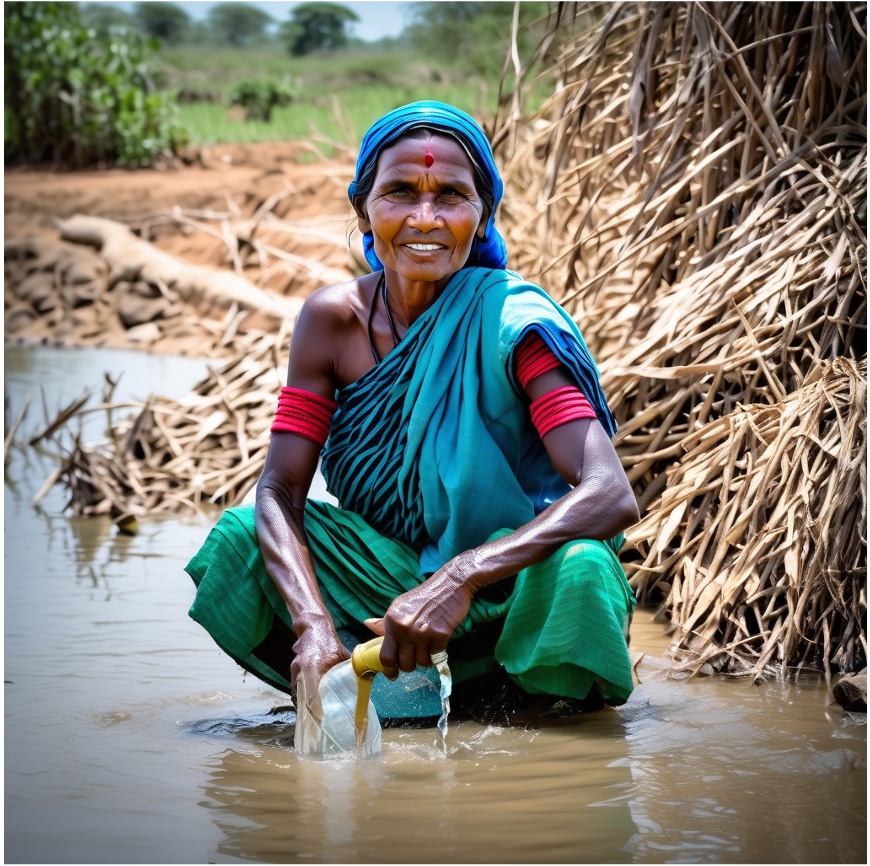}
    \label{fig:ft_example_o3}
  \end{subfigure}
  \hfill
  \begin{subfigure}[b]{0.22\linewidth}
    \centering
    \includegraphics[width=\linewidth]{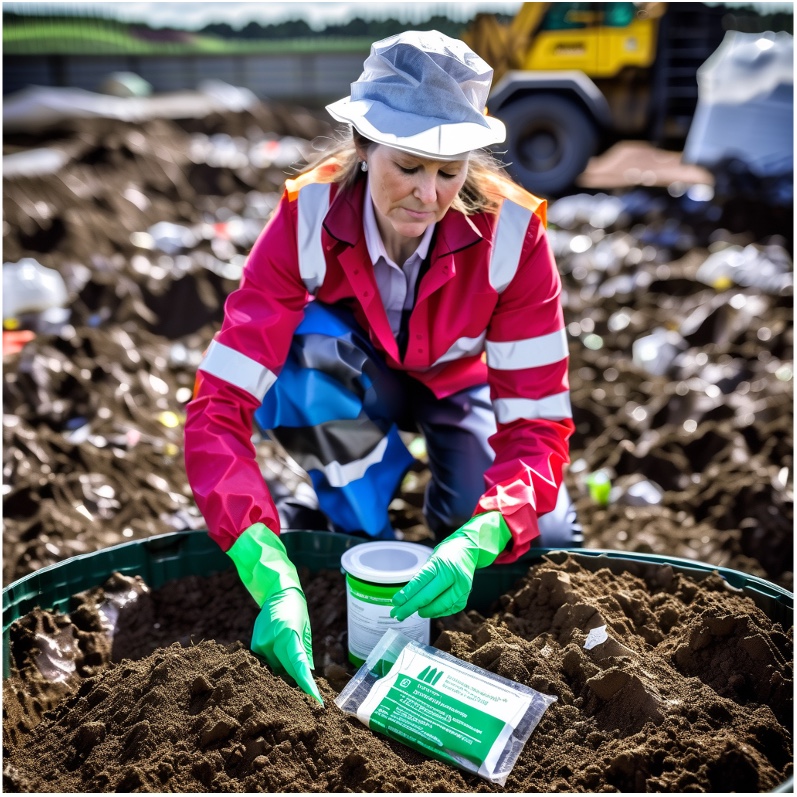}
    \label{fig:ft_example_o4}
  \end{subfigure}
    \vspace{-0.6cm}
  \caption{Examples of the original SDXL (first row) vs. our finetuned model (second row).
  }
    \vspace{-0.5cm}
  \label{fig:finetune_qual}
\end{figure*}

\noindent\textbf{Performance Discrepancy Among Generators}:
We investigate the reasons behind the significant performance gaps in our detection models across different domains, as shown in Tab.~\ref{tab:local_quant} and Tab.~\ref{tab:global_quant}. For instance, DALLE-3 and Midjourney consistently achieve lower AP50 scores compared to the two legacy models, and there is a notable discrepancy between the performance of local and global human artifacts on DALLE-2. To better understand these differences, we conduct a deeper analysis of the predictions made by our models.
We observe a higher proportion of ambiguous cases, similar to those illustrated in Fig.~\ref{fig:id_example_d2_3} and Fig.~\ref{fig:id_example_mj_4}, in datasets from generators that produce high-quality human images, where annotations are sparse. This trend aligns with the data distribution shown in Fig.~\ref{fig:data_stat}.
Such results are expected, as advanced generators like DALLE-3, Midjourney, and DALLE-2 (for global artifacts) produce images with relatively stronger human structural coherence. Consequently, these generators result in fewer and subtler artifacts, which are harder to distinguish. With fewer artifact instances available for evaluation, ambiguous cases make up a larger proportion, exerting a stronger influence on overall accuracy and ultimately lowering reported performance. 
We believe this explains the performance discrepancy among different generators observed in Tab.~\ref{tab:local_quant} and Tab.~\ref{tab:global_quant}. Additional results are provided in the appendix. 


\noindent\textbf{Performance on Unseen Domains}:
To assess the performance of \abbrourmodel~ on unseen domains, we generate images using the validation set of the finetuning prompts introduced in Sec.~\ref{sec:collect_pipeline} with four impactful image generators: Stable Diffusion 1.4 (SD1.4)~\cite{RoBlaLoEsOm22}, PixArt-\(\Sigma\)~\cite{ChenGeXiWuYaReWaLuLuLi24}, FLUX.1-dev~\cite{flux2024}, and Sana~\cite{XieChChCaTaLiZhLiZhLuHa24}.
We also evaluate our models’ robustness on the real human portrait dataset 300W~\cite{SagonasTzZaPa13}.

\begin{table}[t]
\centering
\caption{Comparison of statistics of prediction scores from HumanRefiner and ~\abbrourmodel~ on different datasets.}
\vspace{-0.2cm}
\begin{adjustbox}{max width=0.47\textwidth}
\begin{tabular}{l|ccccc}
\toprule
\multirow{2}{*}{\textbf{Dataset (\# of Img.)}} & \multicolumn{2}{c}{\textbf{HumanRefiner}} & \multicolumn{2}{c}{\textbf{Ours}} \\
\cmidrule(lr){2-3} \cmidrule(lr){4-5}
  & \textbf{Pred./Img. } & \textbf{Mean of Conf.} & \textbf{Pred./Img.} & \textbf{Mean of Conf.} \\
\midrule
SDXL Val (2595)  & 6.22 & 0.19 & 8.46  & 0.31 \\

\midrule
SD1.4~\cite{RoBlaLoEsOm22} (255)  & 2.51 & 0.19 & 9.98  & 0.39 \\
PixArt-\(\Sigma\)~\cite{ChenGeXiWuYaReWaLuLuLi24} (255) & 2.37  & 0.21 & 1.85   & 0.34 \\
Sana~\cite{XieChChCaTaLiZhLiZhLuHa24} (255)         &  4.02 & 0.24 &   1.36  & 0.28 \\
Flux.1-dev~\cite{flux2024} (255)         &  1.35 & 0.25 & 0.45    & 0.26 \\
\midrule
300W~\cite{SagonasTzZaPa13} (600)         &  13.72  & 0.18 & 0.11    & 0.28 \\
\bottomrule
\end{tabular}
\end{adjustbox}
\label{tab:ood_stat}
\vspace{-0.5cm}
\end{table}

We present the top predictions from ~\abbrourmodel~ on the generated images in Fig.~\ref{fig:ood_example} and on 300W in Fig.~\ref{fig:real_example}. Additional results for SD1.4 and other domains are provided in the appendix.
For images from PixArt-\(\Sigma\), ~\abbrourlocalmodel~ accurately identifies a hand with an unusual shape in Fig.~\ref{fig:ood_example_p_a}.
For images generated by the more advanced FLUX.1-dev model, which generally produces higher-quality images with fewer artifacts, \abbrourlocalmodel~ still identifies nuanced issues, including a right hand with odd positioning and texture in Fig.~\ref{fig:ood_example_f_a}, and a hand with an extra finger in Fig.~\ref{fig:ood_example_f_b}. 
~\abbrourglobalmodel~ also detects subtle global artifacts, such as an extra hand and an extra arm in ~\ref{fig:ood_example_sn_1} generated from Sana.
These results highlight the ability of our model to detect subtle issues across diverse and complex scenarios, including those from unseen, more sophisticated generators.
For the real human dataset 300W, ~\abbrourlocalmodel~ detects the hand in Fig.~\ref{fig:real1}. We attribute this to the image’s distinctive black-and-white style, which is rare in the training data from generative models, as well as the complex hand pose, a feature often associated with artifacts in our training data.
In the case of global artifacts, ~\abbrourglobalmodel~ produces an interesting result by incorrectly identifying Wolverine's claw as an extra hand, as shown in Fig.~\ref{fig:real3}. More qualitative results are available in the appendix.

To provide a more comprehensive evaluation, we compare the output statistics of HumanRefiner and our model across all unseen domains, as shown in Tab.~\ref{tab:ood_stat}. Our results reveal a general trend where both the number of predictions and confidence scores from our model decrease as the generator quality improves, from the legacy model of SD1.4 to the more advanced Sana and Flux, indicating its sensitivity to human artifact severity. In contrast, HumanRefiner exhibits a more inconsistent pattern in its prediction statistics outside of the domain of SDXL, suggesting that it struggles to adapt to unseen domains and fails to distinguish poorly generated human figures from higher-quality ones.
When tested on real images, our model produces fewer predictions compared to its performance on generated images. In contrast, HumanRefiner generates an excessive number of predictions on this artifact-free domain, further underscoring its lack of generalization. This inconsistency highlights its limitations as a reliable benchmark for evaluating the quality of human generation in T2I models.
Overall, these results validate the robustness and reliability of our model for effective human artifact detection across diverse domains, emphasizing the necessity of incorporating images from different sources during the training.

\begin{figure*}[h!]
  \centering
   \includegraphics[width=\linewidth]{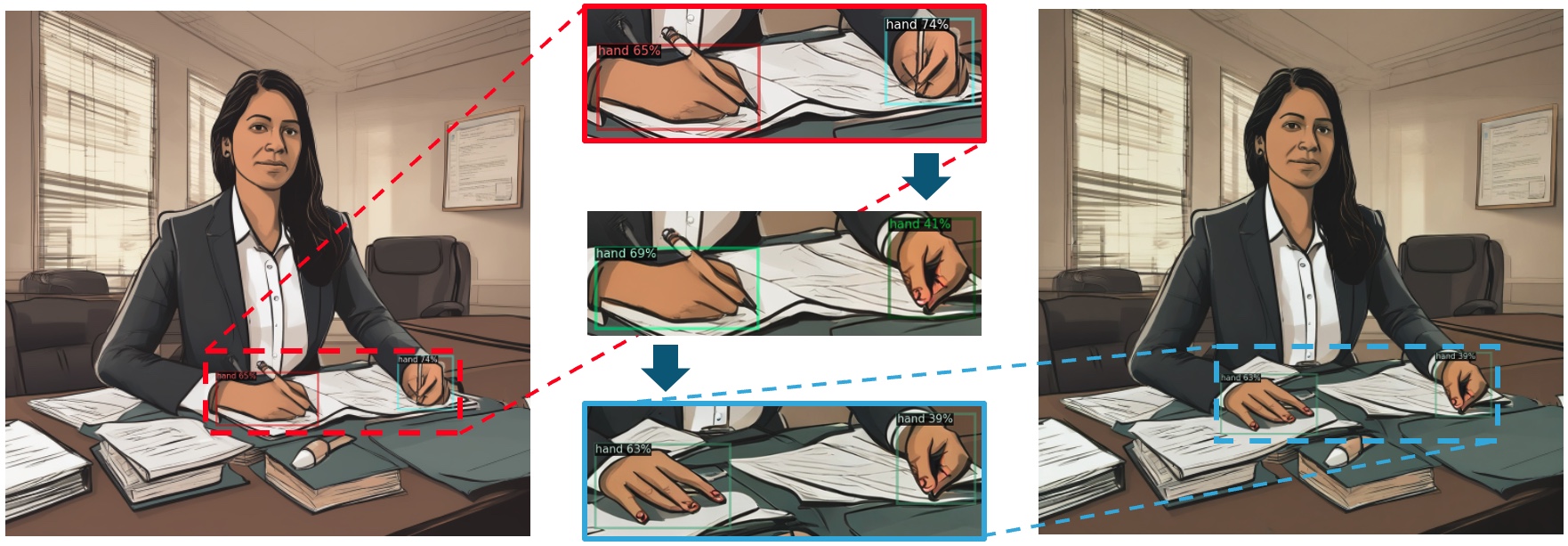}
   \vspace{-0.7cm}
   \caption{Workflow illustrating the reduction of human artifacts through iterative inpainting. Starting with the initial image (left), we identify artifacts using ~\abbrourlocalmodel. These artifacts are iteratively corrected by applying inpainting to the top predictions within the corresponding bounding boxes. For each bounding box, multiple inpainting operations are performed in parallel using different random seeds. From these results, \abbrourlocalmodel~is reapplied, and the sample with the lower confidence score is selected for each result (middle). This iterative process integrates \abbrourlocalmodel~into the inpainting pipeline, automating artifact correction and producing a refined final image (right).}
   \vspace{-0.5cm}
   \label{fig:inpainting}
\end{figure*}

\section{Improvement from Feedback of Human Artifacts Detection}
\label{sec:finetuing_exp}

\subsection{Finetuning Diffusion Models}
\noindent\textbf{Experiment Details}
\label{sec:finetuning_exp_details}
For better generalization, each finetuning prompt is diversified using Llama-3.2-3B~\cite{llama32} with a temperature of 2, generating 10 unique prompts per original prompt. From each diversified prompt, we generate 3 images. We then select the top $k=30\%$ of the predicted bounding boxes based on confidence scores to identify artifacts. During training, both the VAE and the text encoder remain frozen, while LoRA is configured with a rank of 32. The LoRA weights are trained over $80,000$ iterations using the AdamW optimizer~\cite{LoshchilovHu19}, with a constant learning rate of $1 \times 10^{-5}$ and a weight decay of $1 \times 10^{-2}$. Our experiments use a batch size of $8$. For inference, we run the diffusion process in 50 steps.



\noindent\textbf{Qualitative results}
We present pairs of images generated from the same prompt by both the original SDXL model and our finetuned model in Fig.~\ref{fig:finetune_qual}. As shown, our finetuned model effectively reduces severe human artifacts, both local and global, across diverse scenarios and styles. 
To further assess image quality, we conduct a user preference study involving 15 participants. Images are generated from the validation set of diversified finetuning prompts for both models. Using ViTPose~\cite{XuZhZhTa22}, we filter for large, clearly identifiable human figures and randomly select 200 pairs. In each pair, participants are asked to select the image with fewer or less severe human artifacts, avoiding ties whenever possible. Results are presented in Tab.~\ref{tab:user_study}, indicating a preference for images from our finetuned model regarding human artifacts.
This suggests that, while completely eliminating human artifacts remains challenging, even for advanced diffusion models, our approach effectively mitigates these issues by leveraging predictions from ~\abbrourmodel~ to guide diffusion models in recognizing and avoiding typical human artifacts.

\subsection{Refining Artifacts via Inpainting}
\label{sec:inpainting}

In addition to finetuning, we demonstrate that our ~\abbrourlocalmodel~ can be utilized to correct local artifacts in arbitrary images via inpainting, similar to applications as presented in previous works~\cite{ZengLiYaZhShLu20,ZhangZhBaAmLiShSh22}. The pipeline is illustrated in Fig.~\ref{fig:inpainting}.
We adopt SDXL Inpainting-0.1 as the inpainting model in our experiment, setting the guidance scale to 8.0, the inpainting strength to 0.99, and the number of inference steps to 30. 
For each detected artifact, we apply inpainting with 20 different random seeds and select the result based on ~\abbrourlocalmodel’s confidence scores. We note that increasing the number of random seeds generally results in a more thorough search, improving the likelihood of obtaining better results, albeit at the cost of additional time and computational resources.
More results and discussions are available in the appendix. 

\begin{table}[t]
    \centering
    \caption{User preference on original SDXL vs. finetuned SDXL: the percentage of examples where original SDXL is significantly worse ($\ll$), slightly worse ($<$), about the same ($\approx$), slightly better ($>$), significantly better ($\gg$) than finetuned SDXL.}
    \vspace{-0.1cm}
    \begin{tabular*}{0.47\textwidth}{cccccc}
        \hline
        Preference & $\ll$ & $<$ & $\approx$ & $>$ & $\gg$ \\
        \hline
        \% & 29.00 & 26.00 & 6.30 & 21.57 & 17.13 \\
        \hline
    \end{tabular*}
    \vspace{-0.2cm}
    \label{tab:user_study}
\end{table}

\section{Conclusion}
In this work, we introduce a comprehensive approach for detecting and mitigating human artifacts in images generated by text-to-image models. By curating \ourdataset~(\abbrourdataset), a diverse dataset including local and global human artifacts from different T2I models, we establish a foundation for training \ourmodel~(\abbrourmodel) to detect these artifacts. 
Our experiments demonstrate that \abbrourmodel~ not only accurately identify diverse human artifacts across multiple domains but also effectively generalize to images from different and even unseen generative models.
Furthermore, we highlight the application and potential of our approach by showing how ~\abbrourmodel’s predictions can guide the finetuning of diffusion models and enhance inpainting workflows, effectively reducing human artifacts in generated images.



\noindent\textbf{Acknowledgments} We thank Boqing Gong for providing valuable feedback.

\noindent\textbf{Ethical Considerations} Our experiments demonstrate the robustness of ~\abbrourmodel~ in real-world scenarios; however, we emphasize that our models are designed solely for evaluating synthetic images generated by text-to-image models. We strongly discourage applying this model to real-human images, as it may produce unintended or inappropriate interpretations, particularly for individuals with unique health conditions or physical characteristics. Such misapplications could result in unethical or harmful consequences.
{
    \small
    \bibliographystyle{ieeenat_fullname}
    \bibliography{main}
}

\input{sec/supp}

\end{document}

%% file: sec/supp.tex
\clearpage
\setcounter{page}{1}
\maketitlesupplementary

\section{Configuration Details}
\subsection{Data Generation Details}
\begin{figure}[t]
  \centering
   \includegraphics[width=\linewidth]{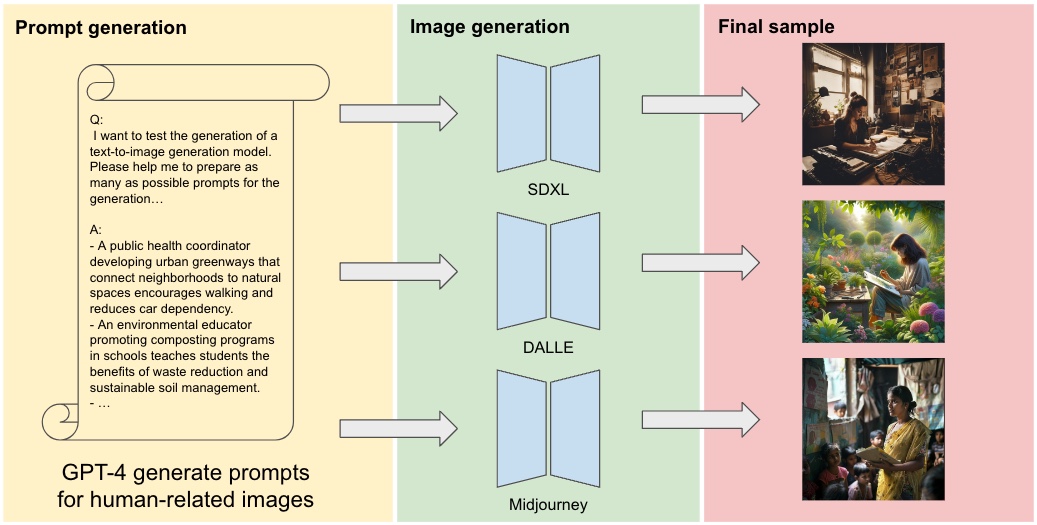}
   \caption{Pipeline of our data collection process.}
   \label{fig:data_pipeline}
\end{figure}

For optimal image quality, both the base and refiner models are used during generation. 
Before annotation, the generated images are filtered by ViTPose~\cite{XuZhZhTa22}, a state-of-the-art human keypoint detection model, to ensure the significant presence of humans and to balance the representation of different body parts. The pipeline is illustrated in Fig.~\ref{fig:data_pipeline}.
We iteratively collected data, starting with SDXL for its availability, and later included premium generators to improve generalization. Data collection ceased after achieving satisfactory generalization, optimizing resource usage. As shown in Tab.~1 and Tab.~2, our models perform comparably across generators, including DALLE-2/3, indicating no bias or performance limitations in underrepresented domains.

\subsection{Training Detection Models}
We use Cascade R-CNN~\cite{CaiVa18} with an EVA-02-ViT-L backbone~\cite{FangSuWaHuWaCa24} with additional training on O365 dataset~\cite{ShaoLiZhPeYuZhLiSu19}. The model is trained with a batch size of 4 and an image resolution of 1024. For optimization, we follow the implementation in EVA-02 and use AdamW~\cite{LoshchilovHu19} with a learning rate of $1 \times 10^{-5}$ and $\beta_1=0.9$ and $\beta_2=0.999$. The model is trained for a total of 250,000 steps. During inference, we set the default threshold value of the detection head to 0.05.

Due to the inherent differences between detecting local and global artifacts, global artifacts detection requires a multi-label setting which necessitates the usage of binary cross-entropy loss. In our experiments, training a single model with separate heads for local and global artifact detection leads to instability. To mitigate this, we choose to train each model separately.

\subsection{Finetuning Diffusion Models}
To enhance generalization, we further diversify the prompts using Llama-3.2-3B~\cite{llama32} before generating images with SDXL.
During the finetuning, detectors' outputs are used to augment training data for finetuning. For each detected artifact, we prepend an identifier (e.g., ``weird hand'' or ``extra leg'') to its prompt and pair it with two images: the original entire image and the cropped artifact region. 

\subsection{Selecting Inpaint Results}
\begin{figure}[t]  
  \centering
  \begin{subfigure}[b]{0.3\linewidth}
    \centering
    \includegraphics[width=\linewidth]{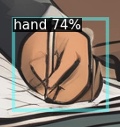}
    \caption{Original artifact}
    \label{fig:crops_compare_1}
  \end{subfigure}
  \hfill
  \begin{subfigure}[b]{0.3\linewidth}  
    \centering
    \includegraphics[width=\linewidth]{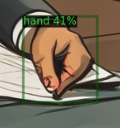}
    \caption{Final selection}
    \label{fig:crops_compare_2}
  \end{subfigure}
  \hfill
  \begin{subfigure}[b]{0.3\linewidth} 
    \centering
    \includegraphics[width=\linewidth]{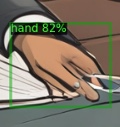}
    \caption{Worse inpaint}
    \label{fig:crops_compare_3}
  \end{subfigure}
  \begin{subfigure}[b]{0.3\linewidth} 
    \centering
    \includegraphics[width=\linewidth]{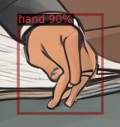}
    \caption{Worse inpaint}
    \label{fig:crops_compare_4}
  \end{subfigure}
  \hfill
  \begin{subfigure}[b]{0.3\linewidth} 
    \centering
    \includegraphics[width=\linewidth]{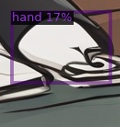}
    \caption{Non-hand inpaint}
    \label{fig:crops_compare_5}
  \end{subfigure}
  \hfill
  \begin{subfigure}[b]{0.3\linewidth} 
    \centering
    \includegraphics[width=\linewidth]{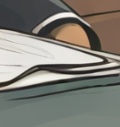}
    \caption{Occluded inpaint}
    \label{fig:crops_compare_6}
  \end{subfigure}

  \caption{Original artifact and different inpainting results with scores from our ~\abbrourlocalmodel.}
   \label{fig:crops_compare}
\end{figure}

Intuitively, we want to avoid generating human body parts with high scores of artifacts. Nevertheless, we find in our study that simply choosing the lowest score detected among the inpainted regions produces suboptimal results. This is because the inpainting model could generate occluded or missing body parts, which leads to a low score or even no prediction from our ~\abbrourlocalmodel. As we demonstrate in Fig.~\ref{fig:crops_compare}, the inpainting model generates diverse samples given the bounding box and different random seeds. 

Although ~\abbrourlocalmodel~ successfully identifies different levels of severity of the artifacts on the hand in Fig.~\ref{fig:crops_compare_2}, Fig.~\ref{fig:crops_compare_3}, and Fig.~\ref{fig:crops_compare_4}, some other regions inpainted as a non-hand object (Fig.~\ref{fig:crops_compare_5}) or occluded hand (Fig.~\ref{fig:crops_compare_6}) are not identified as the typical artifacts and hence get lower scores or even no detection. In this situation, selecting these results with overly low scores may lead to unexpected behavior. Thus, we empirically conclude that the result should be selected with the value closest to half of the original score of the artifact.

\section{Evaluation Metrics}

To assess the performance of our artifact detection model, we employ two widely used evaluation metrics: Area Under the Curve (AUC) of the Receiver Operating Characteristic (ROC) curve and Mean Average Precision (AP50). These metrics provide a comprehensive evaluation of the model’s performance in both classification and localization tasks.

\subsection{Area Under the Curve (AUC)}

The AUC of the ROC curve measures the model’s ability to distinguish between positive and negative samples by evaluating the trade-off between the True Positive Rate (TPR) and the False Positive Rate (FPR) across different classification thresholds~\cite{Fawcett06}. AUC is particularly useful as it provides a threshold-independent measure of performance and is robust to class imbalance.

The True Positive Rate (TPR) and False Positive Rate (FPR) are defined as:

\begin{equation}
    \text{TPR} = \frac{\text{TP}}{\text{TP} + \text{FN}}
\end{equation}

\begin{equation}
    \text{FPR} = \frac{\text{FP}}{\text{FP} + \text{TN}}
\end{equation}

where TP refers to true positives, FP to false positives, TN to true negatives, and FN to false negatives.

The AUC is computed as:

\begin{equation}
    \text{AUC} = \int_{0}^{1} \text{TPR}(t) \, d\text{FPR}(t)
\end{equation}

where \( t \) represents the classification threshold. AUC values range from 0 to 1, with 0.5 indicating random performance and 1.0 representing perfect classification.

\subsection{Average Precision at IoU 0.5 (AP50)}

To evaluate the model’s localization performance, we compute Average Precision at IoU 0.5 (AP50)~\cite{EveringhamGWWZ10}, a widely used metric in object detection benchmarks. AP50 measures precision-recall performance when a predicted bounding box is considered correct if its Intersection over Union (IoU) with the ground-truth box is at least 0.5.

The IoU is defined as:

\begin{equation}
    \text{IoU} = \frac{|B_p \cap B_g|}{|B_p \cup B_g|}
\end{equation}

where \( B_p \) is the predicted bounding box, and \( B_g \) is the ground-truth bounding box.

The Precision (P) and Recall (R) at a given IoU threshold are:

\begin{equation}
    \text{Precision} = \frac{\text{TP}}{\text{TP} + \text{FP}}, \quad 
    \text{Recall} = \frac{\text{TP}}{\text{TP} + \text{FN}}
\end{equation}

The Average Precision (AP) at IoU 0.5 is computed as:

\begin{equation}
    \text{AP}_{50} = \int_{0}^{1} P(R) \, dR
\end{equation}

where \( P(R) \) is the precision as a function of recall. AP50 is reported per category, allowing for a detailed analysis of localization performance across different artifact types.

\begin{figure*}[h!]
  \centering

  \begin{subfigure}[b]{0.49\linewidth}
    \centering
    \includegraphics[width=\linewidth]{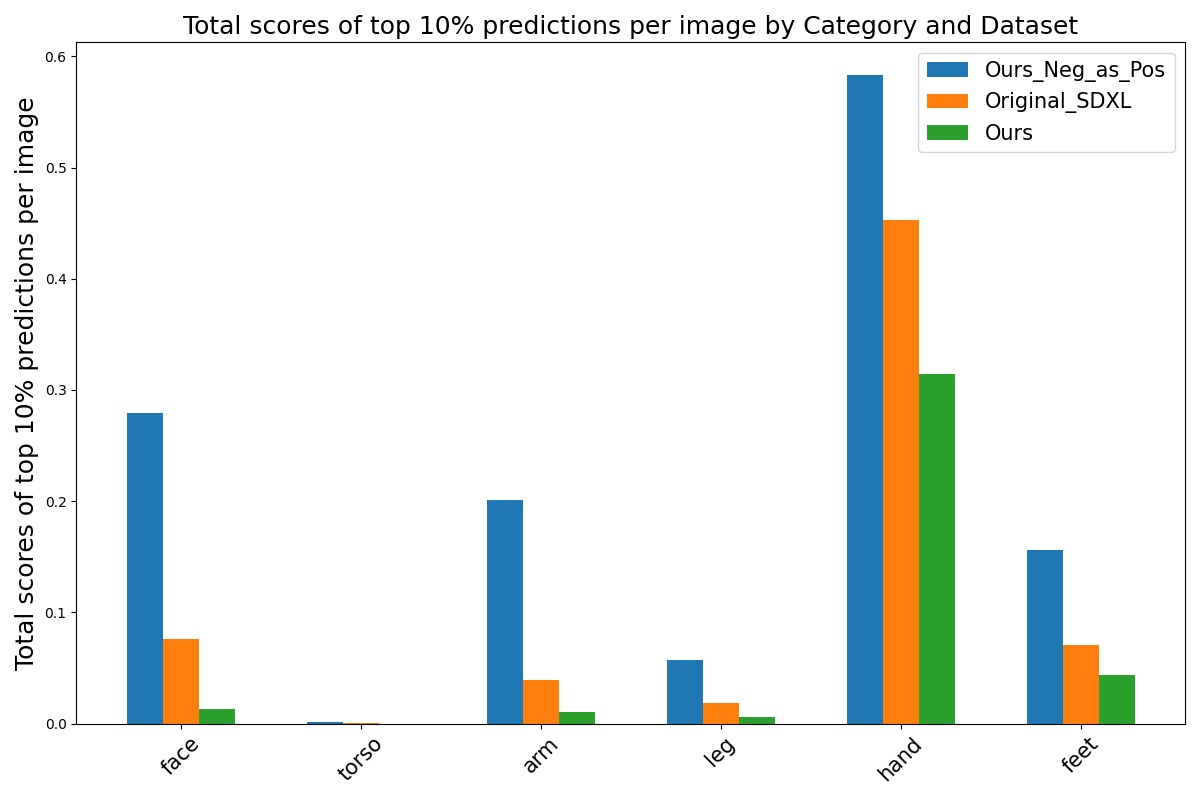}
  \end{subfigure}
  \hfill
  \begin{subfigure}[b]{0.49\linewidth}  
    \centering
    \includegraphics[width=\linewidth]{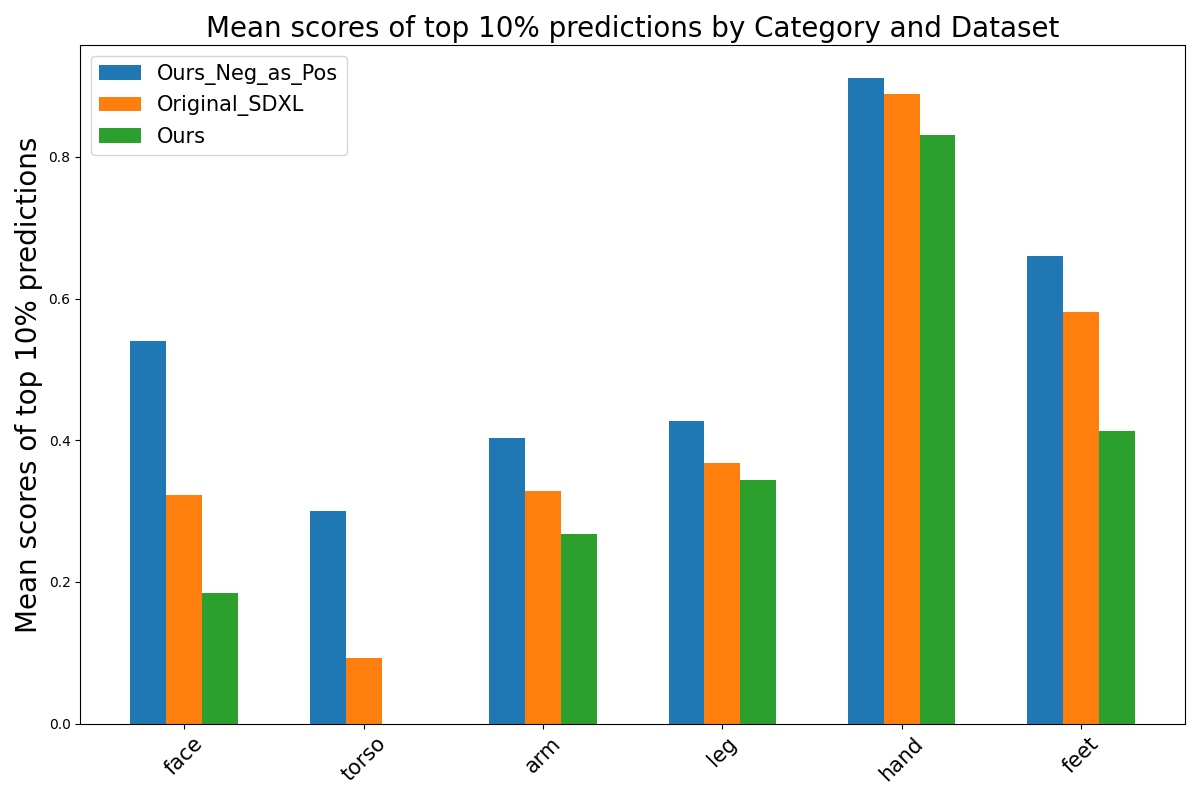}
  \end{subfigure}
\vspace{-0.2cm}
\caption{Total scores per image (left) and average scores (right) of top $10\%$ detected local human artifacts.}
\label{fig:ft_quant_local}
\end{figure*}

\begin{figure*}[h!]
  \centering
  \begin{subfigure}[b]{0.49\linewidth}
    \centering
    \includegraphics[width=\linewidth]{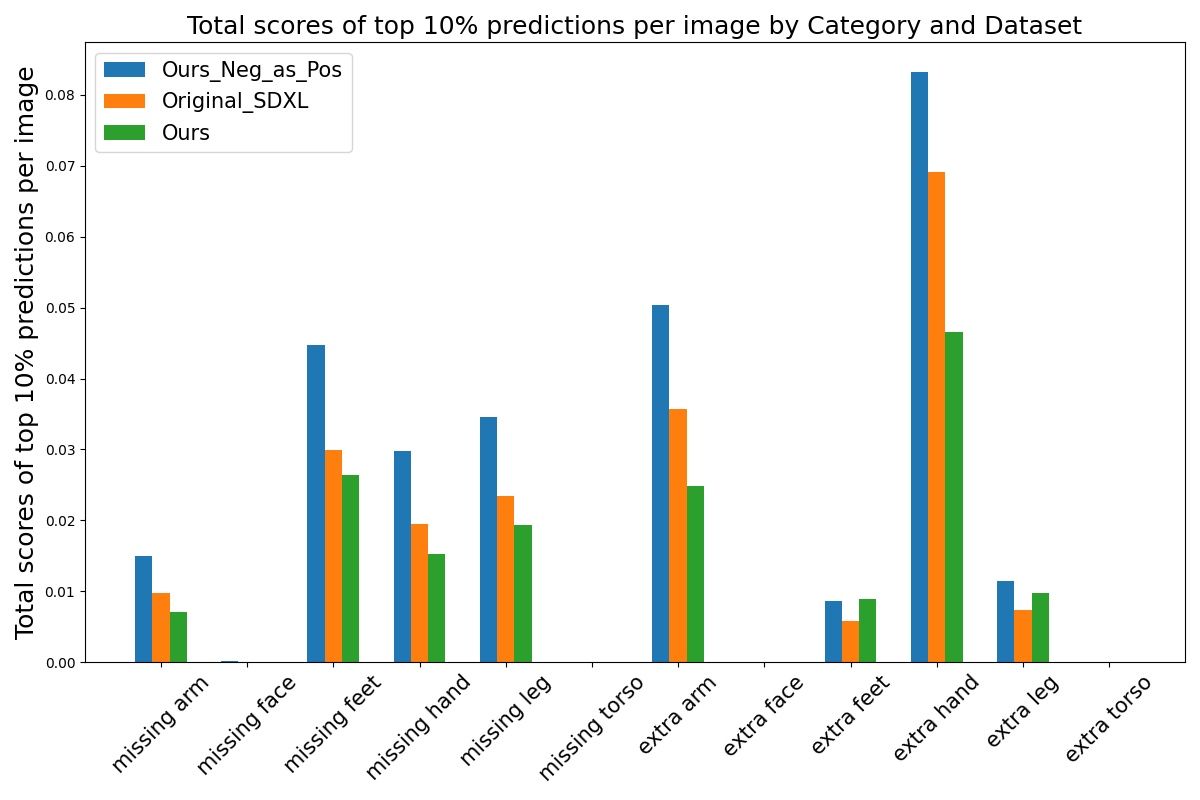}
  \end{subfigure}
  \hfill
  \begin{subfigure}[b]{0.49\linewidth}  
    \centering
    \includegraphics[width=\linewidth]{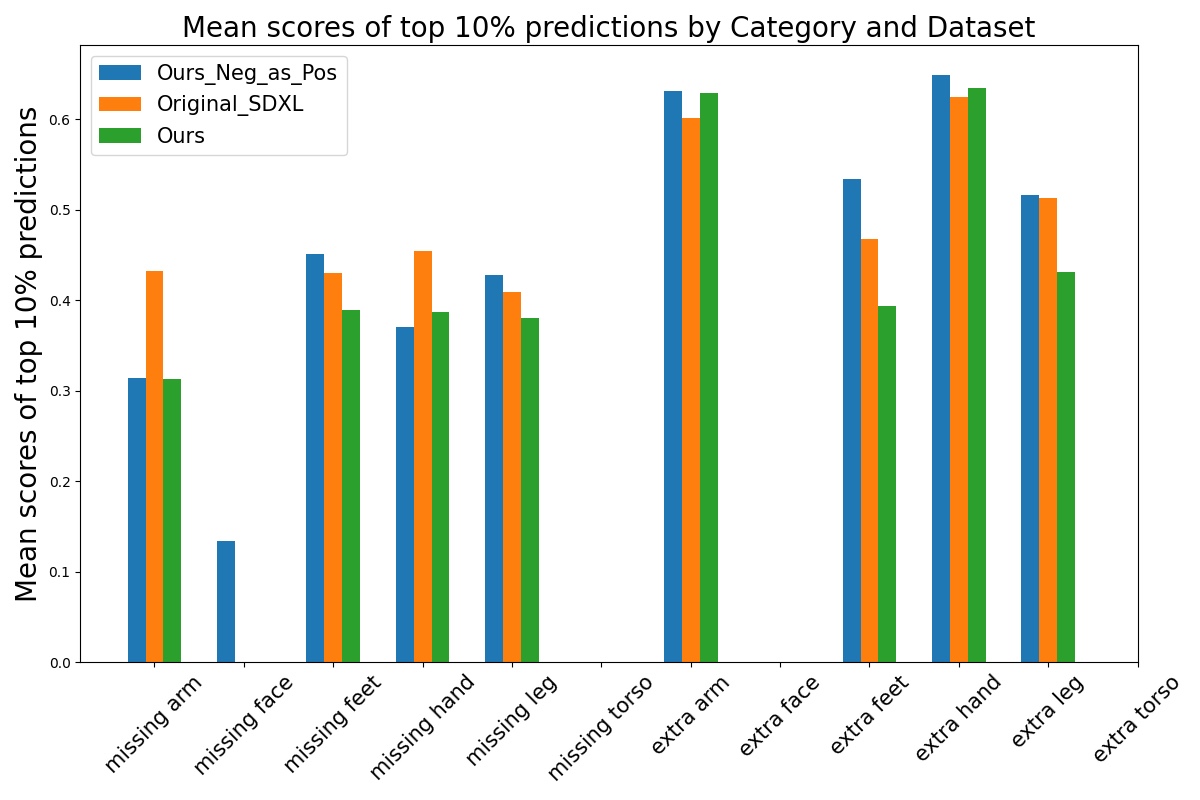}
  \end{subfigure}
\vspace{-0.2cm}
\caption{Total scores per image (left) and average scores (right) of top $10\%$ detected global human artifacts.}
\label{fig:ft_quant_global}
\end{figure*}

\section{Ablation Studies}
\label{sec:ablation}

\begin{table}[h!]
    \centering
    \caption{Ablation study of key components in ~\abbrourlocalmodel. AP50 scores are reported, with each additional configuration building upon the previous one.}
    \resizebox{0.47\textwidth}{!}{
    \begin{tabular}{ccccc}
        \hline
        Config. & ViT-B & + ViT-L & + EVA02 & + Real \\
        \hline
        AP50 & 32.7 & 35.3 & 39.8 & 43.3 \\
        \hline
    \end{tabular}
    }
    \label{tab:abl_local}
\end{table}

\begin{table}[h!]
    \centering
    \caption{Ablation study of key components in ~\abbrourglobalmodel. }  
    \resizebox{0.47\textwidth}{!}{
    \begin{tabular}{ccccc}
        \hline
        Config. & ViT-B & + EVA02 & + Real & + BCE \\
        \hline
        AP50 & 14.4 & 17.7 & 17.8 & 25.5 \\
        \hline
    \end{tabular}
    }
    \label{tab:abl_global}
\end{table}

\begin{table}[h!]
    \centering
    \caption{Performance of YOLOv8 on our dataset.}
    \resizebox{0.47\textwidth}{!}{
    \begin{tabular}{cccc}
        \hline
        Config. & YOLOv8n & YOLOv8x & Ours \\
        \hline
        AP50 & 25.7 & 35.3 & 43.3 \\
        \hline
    \end{tabular}
    }
    \label{tab:abl_arch}
\end{table}

We conduct ablation studies to evaluate the impact of key components in ~\abbrourmodel. This includes examining the effect of backbone capacity (ViT-L and EVA02), the use of real-domain regularization images (Real) during training for ~\abbrourlocalmodel, and the application of binary cross-entropy (BCE) loss for ~\abbrourglobalmodel. Results for both models are shown in Tables~\ref{tab:abl_local} and~\ref{tab:abl_global}, respectively.
As seen in Tab.~\ref{tab:abl_local}, each enhancement in the backbone architecture results in a clear performance gain, with the inclusion of real-domain images yielding the best performance for ~\abbrourlocalmodel.
In contrast, Tab.~\ref{tab:abl_global} shows that while increasing backbone capacity similarly improves the AP50 score, adding real-domain regularization images has minimal impact (+0.1). We attribute this to the limitations of cross-entropy loss in the multi-label setting. Switching to binary cross-entropy loss brings a substantial performance boost (+7.7).

Additionally, we test our dataset using another widely used detection model, YOLOv8~\cite{yolov8_ultralytics}, which is adopted in HumanRefiner, for local artifact detection. As shown in Tab.~\ref{tab:abl_arch}, our model, ViTDet with EVA-02, significantly outperforms YOLOv8, achieving notably higher mAP scores (43.3 vs. 25.7 for YOLOv8n and 35.3 for YOLOv8x).

\begin{figure*}[t]
  \centering
   \includegraphics[width=\linewidth]{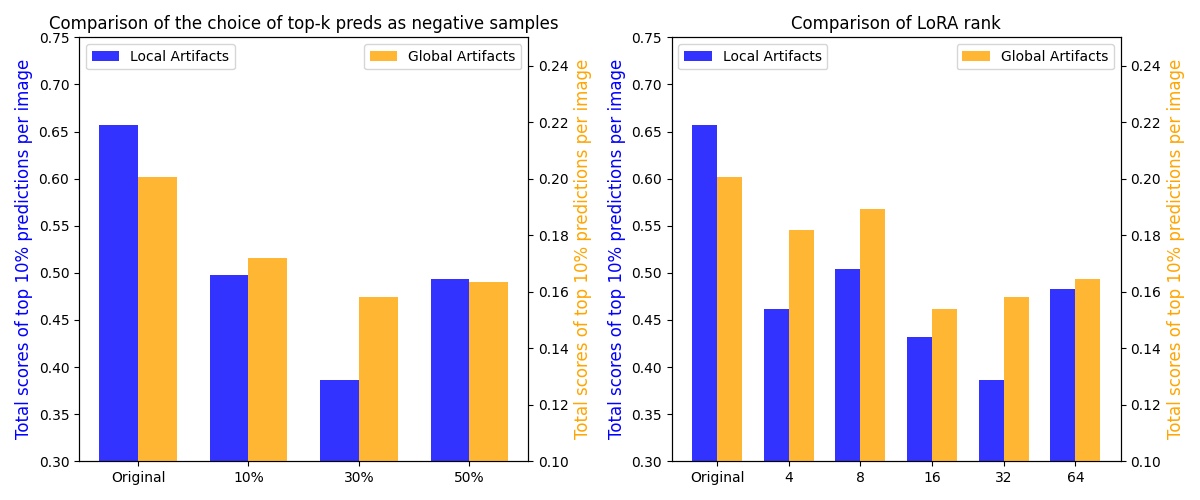}
   \caption{Comparison of detectors' top $10\%$ predictions on images generated by SDXL finetuned with different values of confidence ratio  $k$ (left) and LoRA rank (right).}
   \label{fig:quant_abl}
\end{figure*}


\section{Quantitative Results for Finetuned Diffusion Models}
\label{sec:ft_quant}
To more comprehensively evaluate the performance of our finetuned SDXL model, we compare the quantitative detection results from ~\abbrourmodel~ on images generated by the original and finetuned SDXL models.
To further demonstrate the human artifact patterns associated with the special identifiers learned during finetuning, we not only generate images using them as negative prompts but also prepend them to the prompts during inference to generate another set of baseline images for comparison.
We then compute the total scores per image and the average scores of the top $10\%$ predictions from ~\abbrourlocalmodel~ and ~\abbrourglobalmodel~ on the generated images, presenting the results in Fig.~\ref{fig:ft_quant_local} and Fig.~\ref{fig:ft_quant_global}.
These results show that, compared to the original SDXL, our finetuned model produces fewer artifacts in terms of both quantity (lower total scores) and quality (lower average scores), which aligns with our user studies.
Moreover, when using the identifiers as part of the prompts instead of as negative prompts, the finetuned model generates significantly more human artifacts. This demonstrates the enhanced perception of human artifacts by the finetuned model, validating the effectiveness of our finetuning approach, which benefits from the feedback provided by ~\abbrourmodel.

To validate our hyperparameter choices, we evaluate the performance of our fine-tuned SDXL model using different confidence threshold values (\( k = 10\%, 30\%, 50\% \)) for negative sample selection, along with varying LoRA rank values during fine-tuning. The average sum of the top \( 10\% \) artifact scores per image is presented in Fig.~\ref{fig:quant_abl}. Our findings indicate that smaller \( k \) values are less effective due to a high rate of false negatives, while larger \( k \) values lead to an increase in false alarms, negatively impacting performance in the opposite direction. We find that \( k = 30\% \) provides a good balance. Additionally, a LoRA rank of 32 achieves the optimal trade-off between computational efficiency and artifact reduction.

\begin{figure*}[htbp]
  \centering
  \begin{subfigure}{0.85\linewidth}
    \centering
    \includegraphics[width=\linewidth]{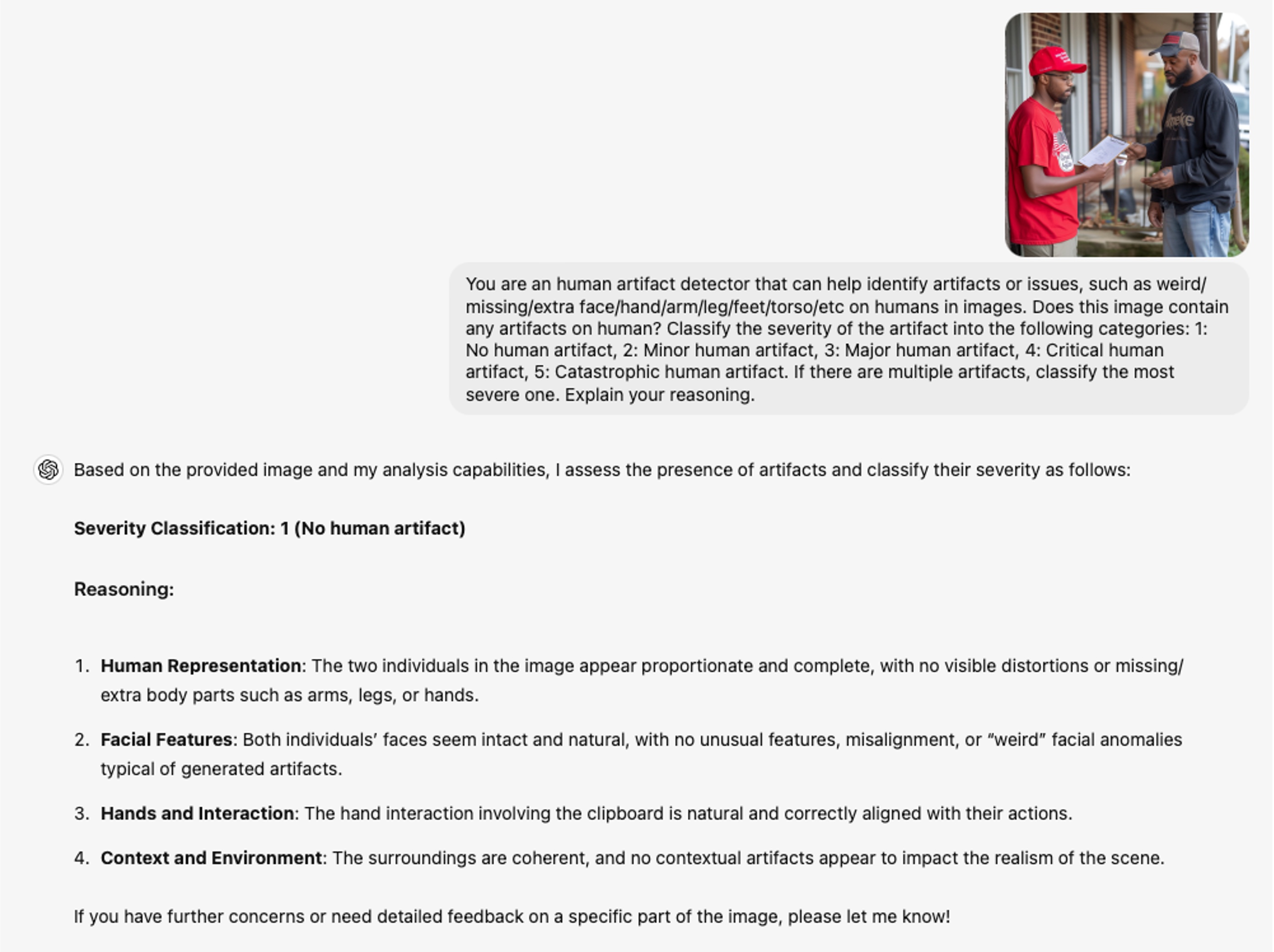}
    \caption{GPT-4o fails to identify any artifacts given the original image without any hint.}
    \label{fig:sub1}
  \end{subfigure}

  \begin{subfigure}{0.85\linewidth}
    \centering
    \includegraphics[width=\linewidth]{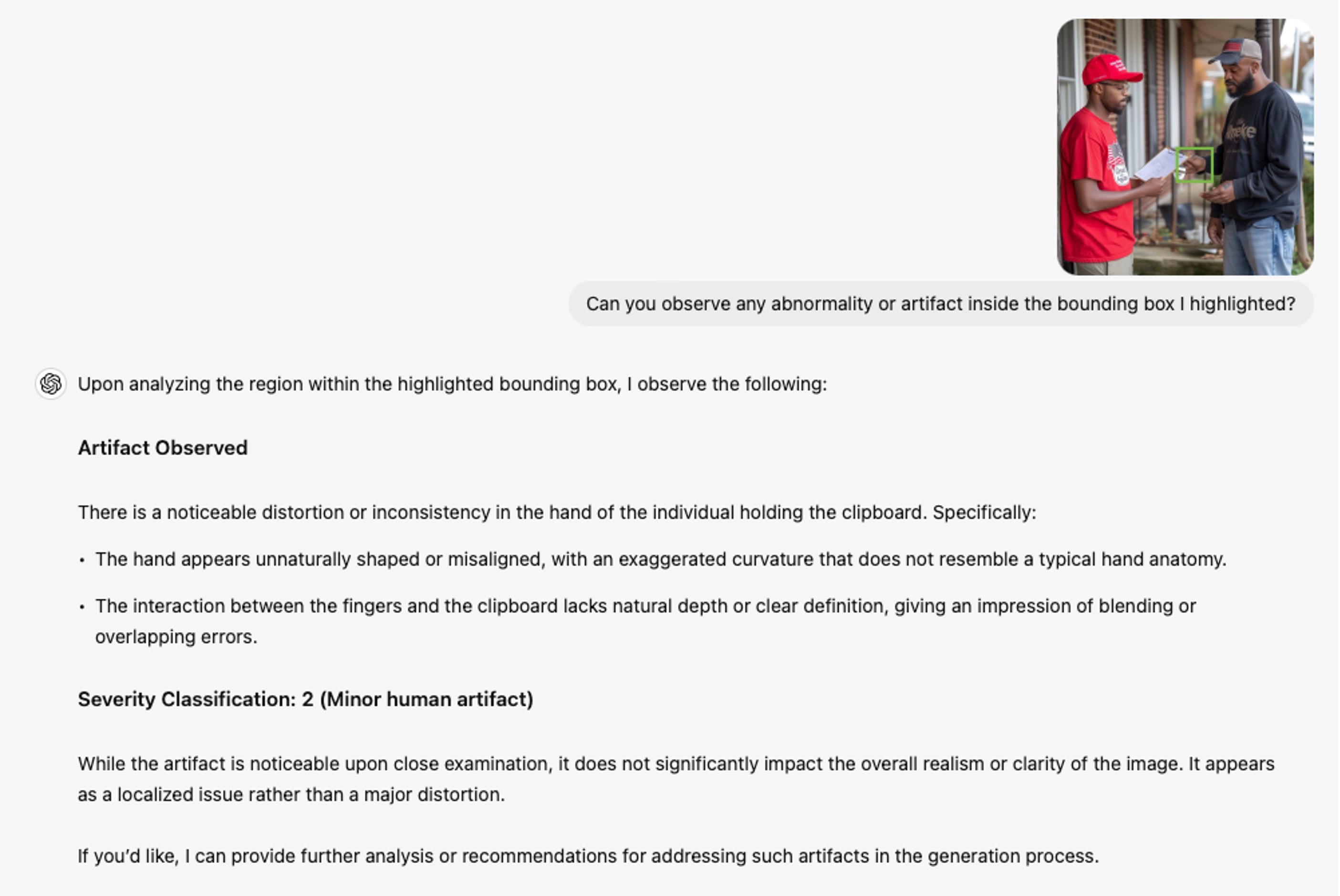}
    \caption{When the distorted hand of the right figure is highlighted with a bounding box, GPT-4o recognizes minor artifacts, including the distortion in the hand and its unnatural interaction with objects.}
    \label{fig:sub2}
  \end{subfigure}
  
\end{figure*}

\begin{figure*}[htbp]
  \ContinuedFloat 
  \centering
  \begin{subfigure}{0.85\linewidth}
    \centering
    \includegraphics[width=\linewidth]{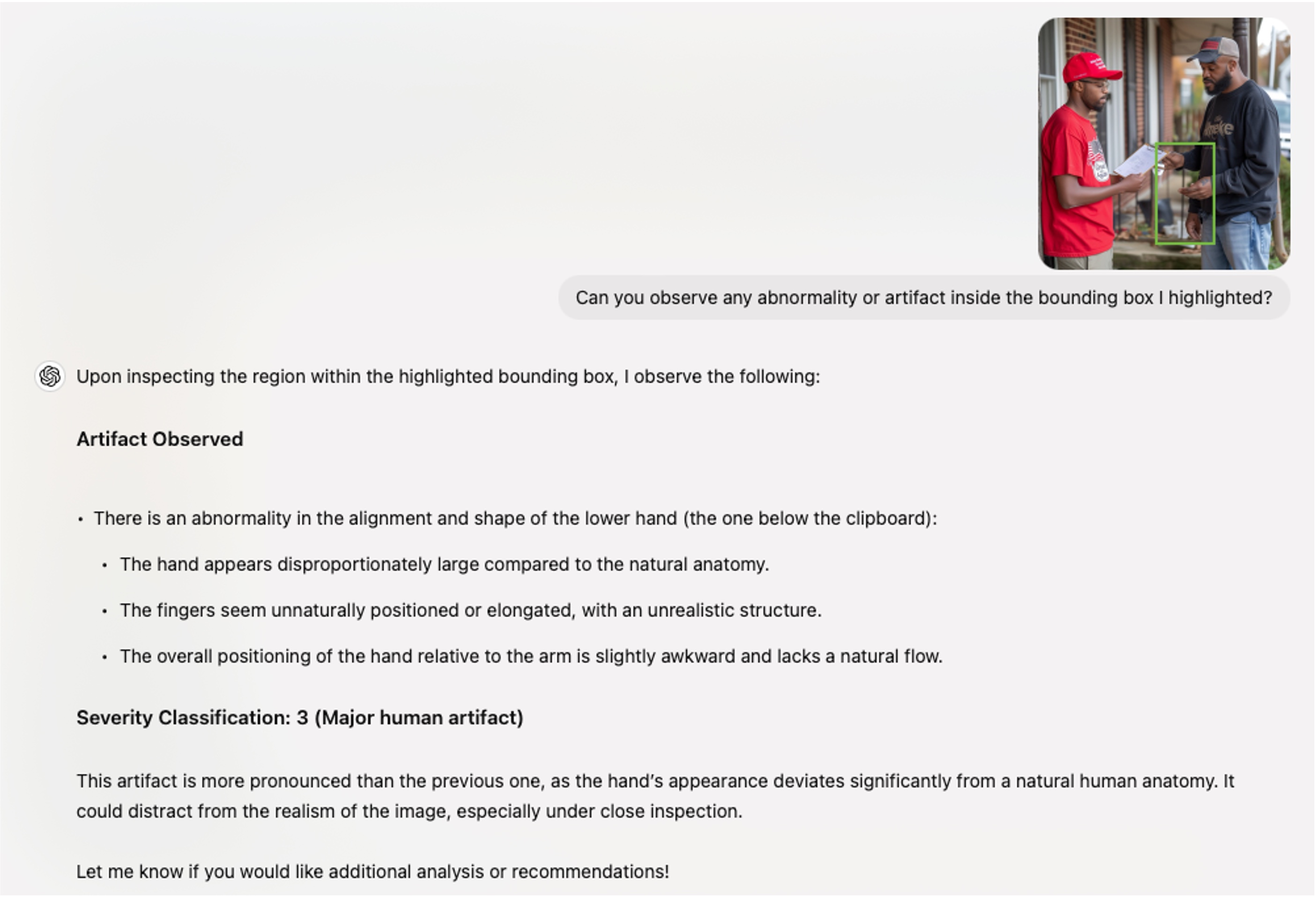}
    \caption{When the three arms of the right figure are highlighted with a bounding box, GPT-4o claims to observe major artifacts but fails to justify the prediction by identifying the three arms.}
    \label{fig:sub3}
  \end{subfigure}

  \begin{subfigure}{0.85\linewidth}
    \centering
    \includegraphics[width=\linewidth]{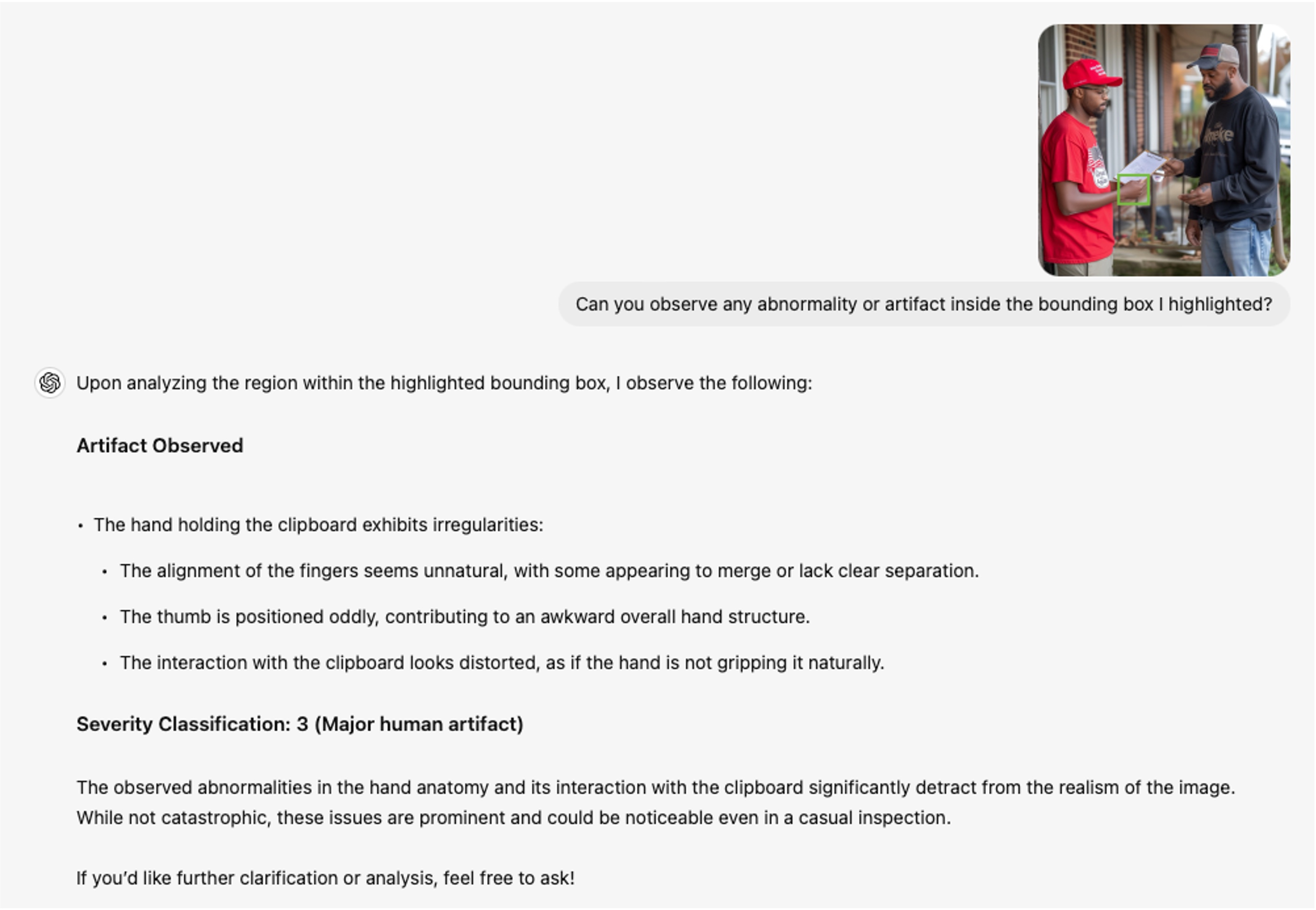}
    \caption{When the hand of the left figure, which has better quality, is highlighted with a bounding box, GPT-4o still predicts major human artifacts.}
    \label{fig:sub4}
  \end{subfigure}

\end{figure*}

\begin{figure*}[htbp]
  \ContinuedFloat 
  \centering
  \begin{subfigure}{0.85\linewidth}
    \centering
    \includegraphics[width=\linewidth]{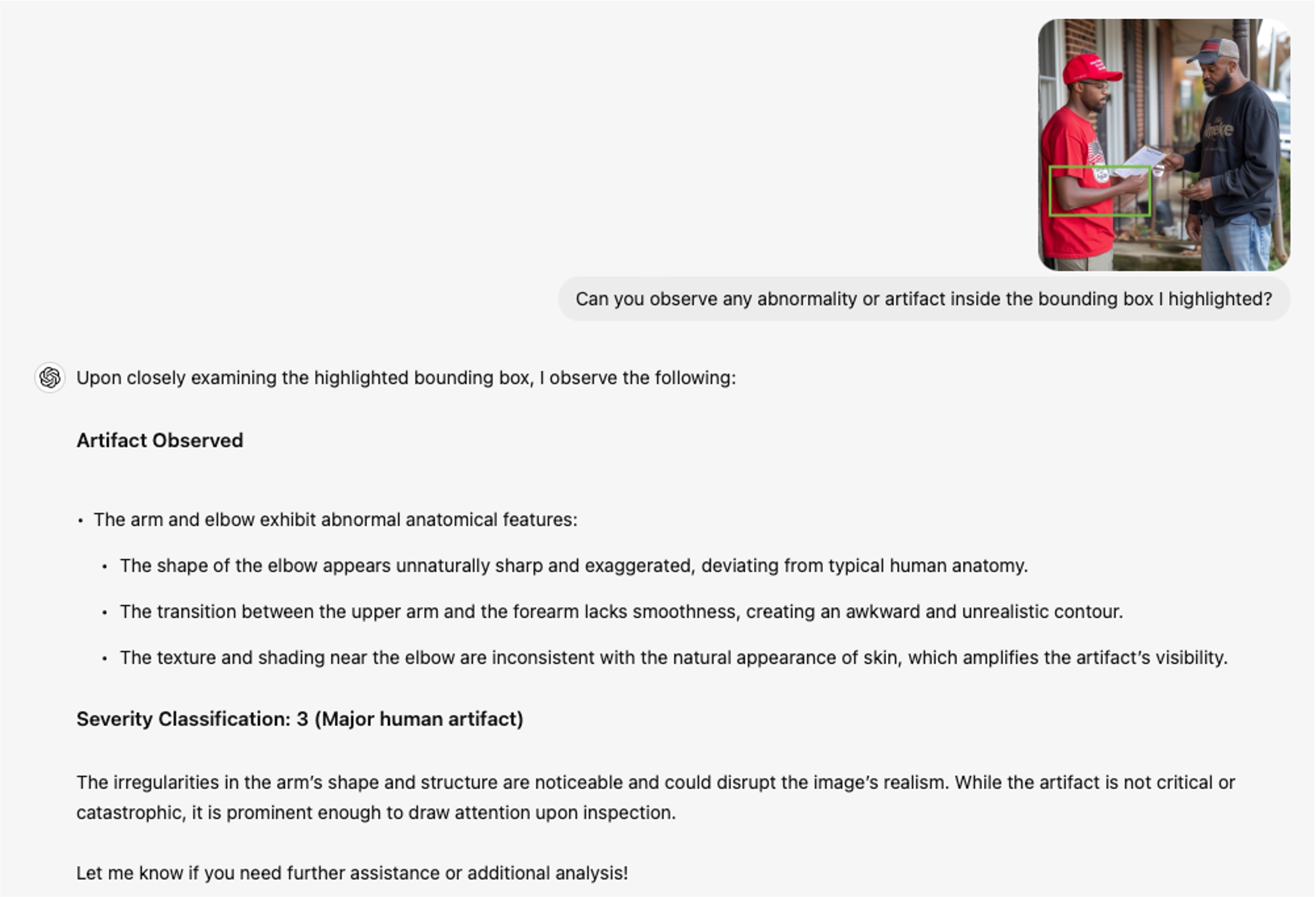}
    \caption{When the entire arm of the left figure, which has better quality, is highlighted with a bounding box, GPT-4o still predicts major human artifacts.}
    \label{fig:sub5}
  \end{subfigure}
  \vspace{-0.3cm}
  \caption{Dialogue with GPT-4o providing bounding boxes as hints.}
  \label{fig:gpt4o-chat}
\end{figure*}

\section{Testing GPT-4o with more hints}

Given GPT-4o’s limited ability to identify, localize, and analyze features within an image, as demonstrated in the main paper, we further evaluate its capacity to recognize artifacts when provided with additional hints, such as bounding boxes. The corresponding dialogue is presented in Fig.~\ref{fig:gpt4o-chat}. The results show that even with localized hints, GPT-4o fails to accurately identify the artifacts or assess their severity. Instead, its behavior appears to conform to the expectations set by the prompt, consistently making positive predictions based on the presence of bounding boxes without providing justification or demonstrating a true understanding of the artifacts.

\section{Additional Qualitative Examples}
\begin{figure*}[t] 
  \vspace{0.5cm}
  \centering
  \begin{subfigure}[b]{0.24\linewidth}
    \centering
    \includegraphics[width=\linewidth]{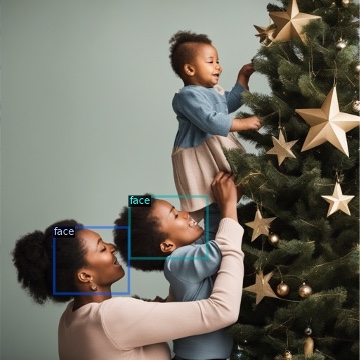}
    \label{fig:local_sdxl_1}
  \end{subfigure}
  \hfill
  \begin{subfigure}[b]{0.24\linewidth}
    \centering
    \includegraphics[width=\linewidth]{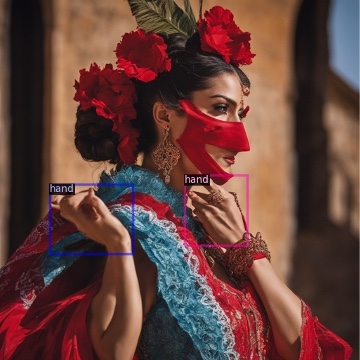}
    \label{fig:local_sdxl_2}
  \end{subfigure}
  \hfill
  \begin{subfigure}[b]{0.24\linewidth}
    \centering
    \includegraphics[width=\linewidth]{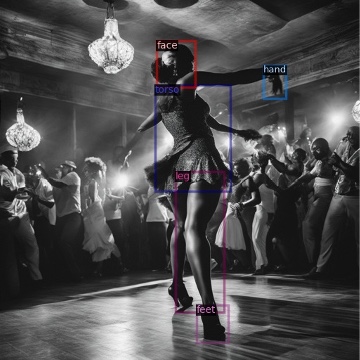}
    \label{fig:local_sdxl_3}
  \end{subfigure}
  \hfill
  \begin{subfigure}[b]{0.24\linewidth}
    \centering
    \includegraphics[width=\linewidth]{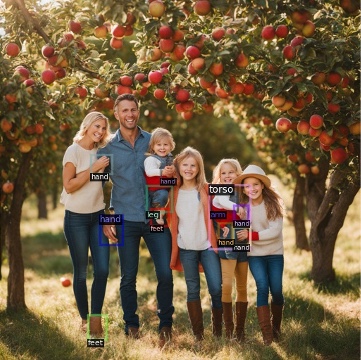}
    \label{fig:local_sdxl_4}
  \end{subfigure}
  \vspace{0.3cm}

  \begin{subfigure}[b]{0.24\linewidth}
    \centering
    \includegraphics[width=\linewidth]{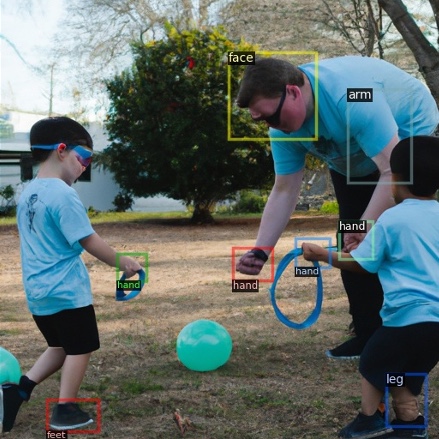}
    \label{fig:local_dalle2_1}
  \end{subfigure}
  \hfill
  \begin{subfigure}[b]{0.24\linewidth}
    \centering
    \includegraphics[width=\linewidth]{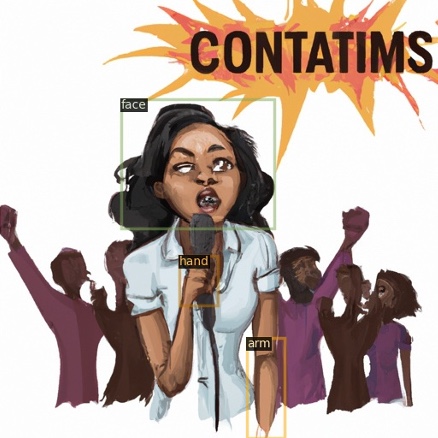}
    \label{fig:local_dalle2_2}
  \end{subfigure}
  \hfill
  \begin{subfigure}[b]{0.24\linewidth}
    \centering
    \includegraphics[width=\linewidth]{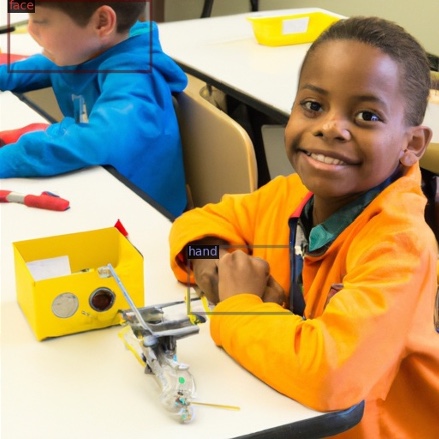}
    \label{fig:local_dalle2_3}
  \end{subfigure}
  \hfill
  \begin{subfigure}[b]{0.24\linewidth}
    \centering
    \includegraphics[width=\linewidth]{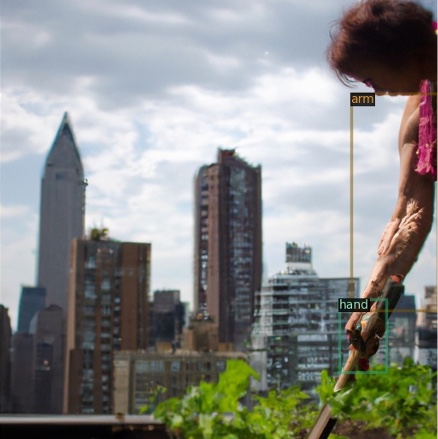}
    \label{fig:local_dalle2_4}
  \end{subfigure}
  \vspace{0.3cm}

  \begin{subfigure}[b]{0.24\linewidth}
    \centering
    \includegraphics[width=\linewidth]{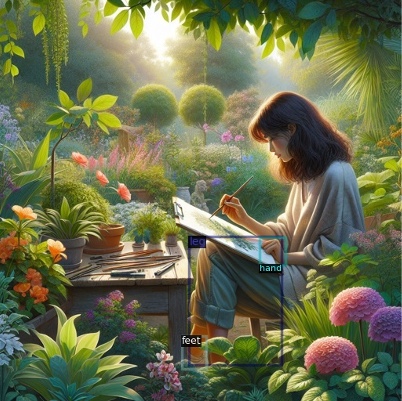}
    \label{fig:local_dalle3_1}
  \end{subfigure}
  \hfill
  \begin{subfigure}[b]{0.24\linewidth}
    \centering
    \includegraphics[width=\linewidth]{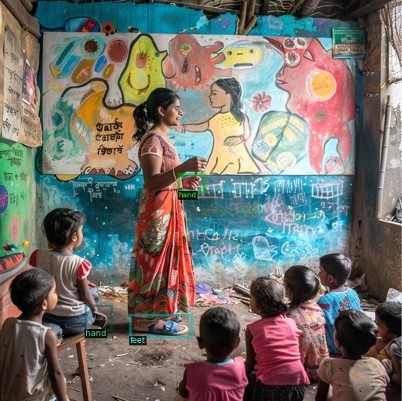}
    \label{fig:local_dalle3_2}
  \end{subfigure}
  \hfill
  \begin{subfigure}[b]{0.24\linewidth}
    \centering
    \includegraphics[width=\linewidth]{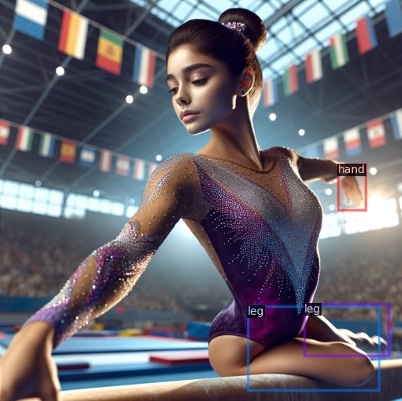}
    \label{fig:local_dalle3_3}
  \end{subfigure}
  \hfill
  \begin{subfigure}[b]{0.24\linewidth}
    \centering
    \includegraphics[width=\linewidth]{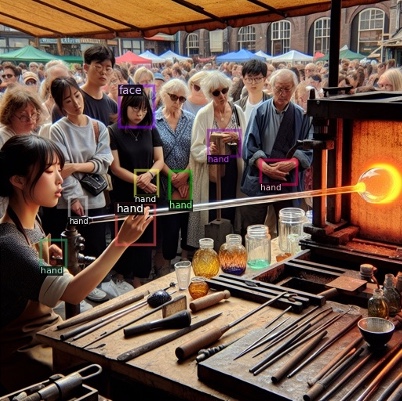}
    \label{fig:local_dalle3_4}
  \end{subfigure}
  \vspace{0.3cm}

  \begin{subfigure}[b]{0.24\linewidth}
    \centering
    \includegraphics[width=\linewidth]{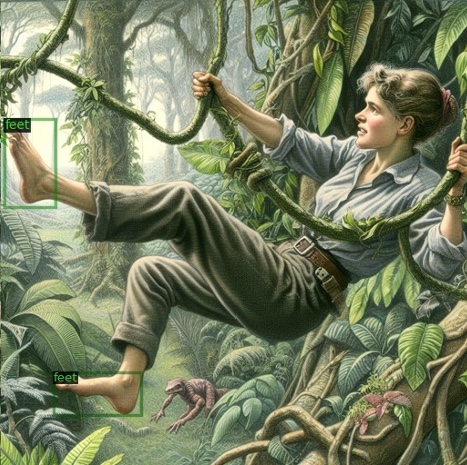}
    \label{fig:local_mj_1}
  \end{subfigure}
  \hfill
  \begin{subfigure}[b]{0.24\linewidth}
    \centering
    \includegraphics[width=\linewidth]{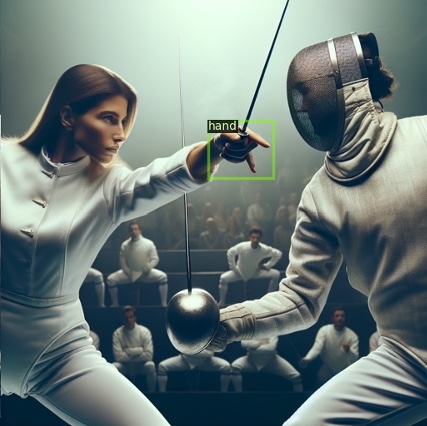}
    \label{fig:local_mj_2}
  \end{subfigure}
  \hfill
  \begin{subfigure}[b]{0.24\linewidth}
    \centering
    \includegraphics[width=\linewidth]{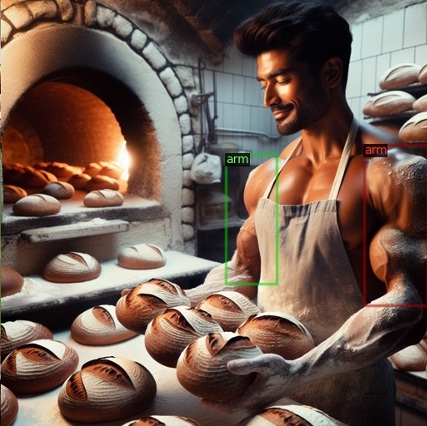}
    \label{fig:local_mj_3}
  \end{subfigure}
  \hfill
  \begin{subfigure}[b]{0.24\linewidth}
    \centering
    \includegraphics[width=\linewidth]{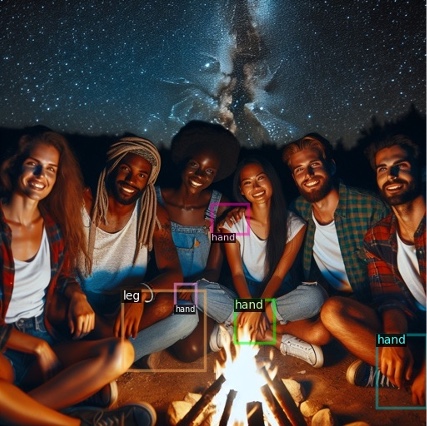}
    \label{fig:local_mj_4}
  \end{subfigure}
  \vspace{0.3cm}

  \caption{More examples of annotations from ~\ourdataset~ for local human artifacts. First row: SDXL. Second row: DALLE-2. Third row: DALLE-3. Last row: Midjourney.}
  \label{fig:more_anno_local}
\end{figure*}

\begin{figure*}[t] 
  \vspace{0.5cm}
  \centering
  \begin{subfigure}[b]{0.24\linewidth}
    \centering
    \includegraphics[width=\linewidth]{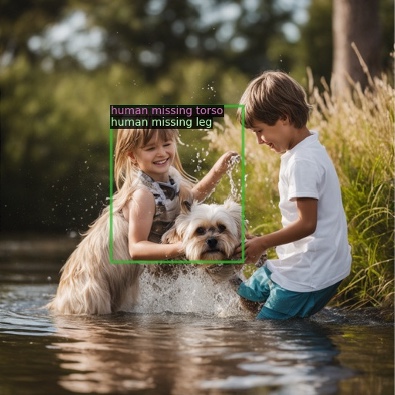}
    \label{fig:global_sdxl_1}
  \end{subfigure}
  \hfill
  \begin{subfigure}[b]{0.24\linewidth}
    \centering
    \includegraphics[width=\linewidth]{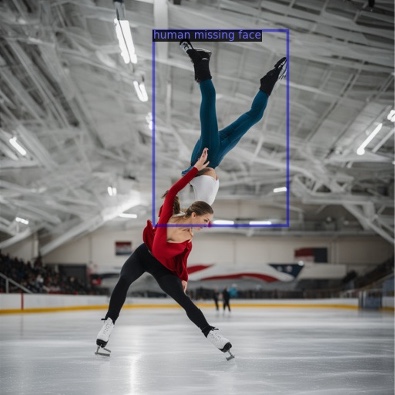}
    \label{fig:global_sdxl_2}
  \end{subfigure}
  \hfill
  \begin{subfigure}[b]{0.24\linewidth}
    \centering
    \includegraphics[width=\linewidth]{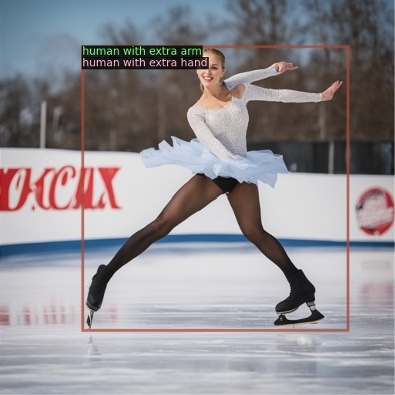}
    \label{fig:global_sdxl_3}
  \end{subfigure}
  \hfill
  \begin{subfigure}[b]{0.24\linewidth}
    \centering
    \includegraphics[width=\linewidth]{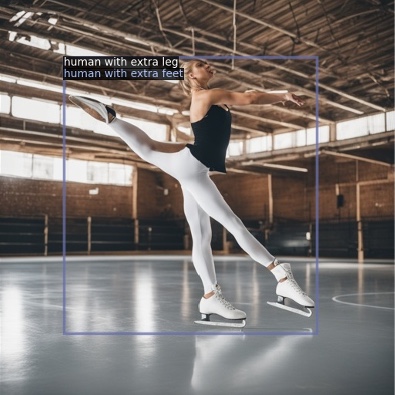}
    \label{fig:global_sdxl_4}
  \end{subfigure}
  \vspace{0.3cm}

  \begin{subfigure}[b]{0.24\linewidth}
    \centering
    \includegraphics[width=\linewidth]{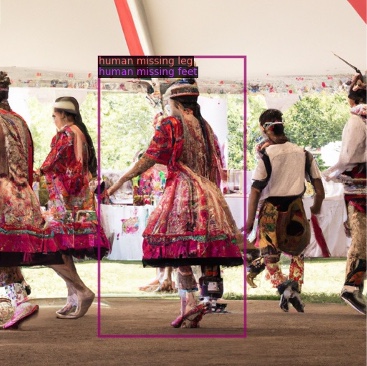}
    \label{fig:global_dalle2_1}
  \end{subfigure}
  \hfill
  \begin{subfigure}[b]{0.24\linewidth}
    \centering
    \includegraphics[width=\linewidth]{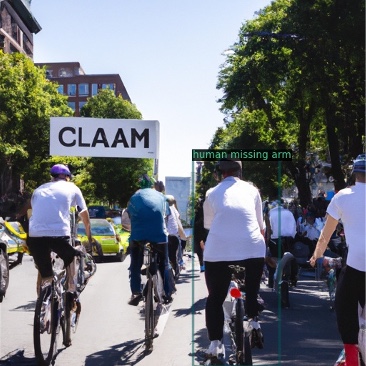}
    \label{fig:global_dalle2_2}
  \end{subfigure}
  \hfill
  \begin{subfigure}[b]{0.24\linewidth}
    \centering
    \includegraphics[width=\linewidth]{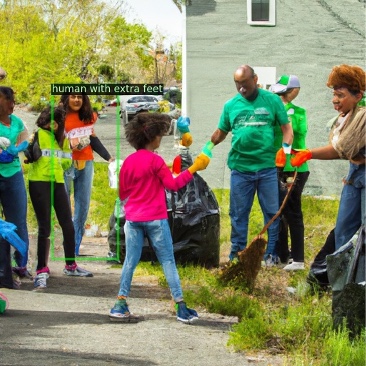}
    \label{fig:global_dalle2_3}
  \end{subfigure}
  \hfill
  \begin{subfigure}[b]{0.24\linewidth}
    \centering
    \includegraphics[width=\linewidth]{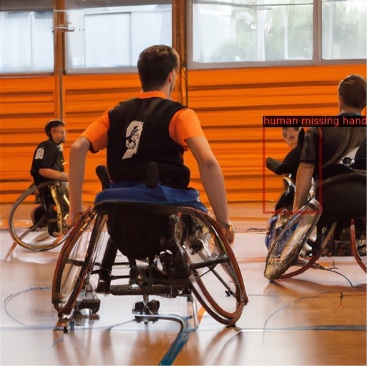}
    \label{fig:global_dalle2_4}
  \end{subfigure}
  \vspace{0.3cm}

  \begin{subfigure}[b]{0.24\linewidth}
    \centering
    \includegraphics[width=\linewidth]{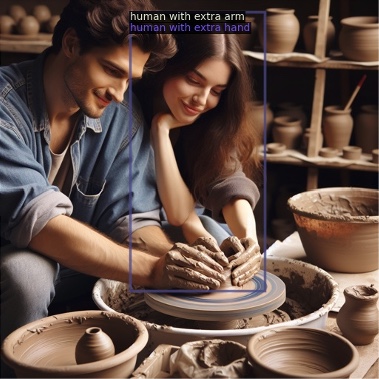}
    \label{fig:global_dalle3_1}
  \end{subfigure}
  \hfill
  \begin{subfigure}[b]{0.24\linewidth}
    \centering
    \includegraphics[width=\linewidth]{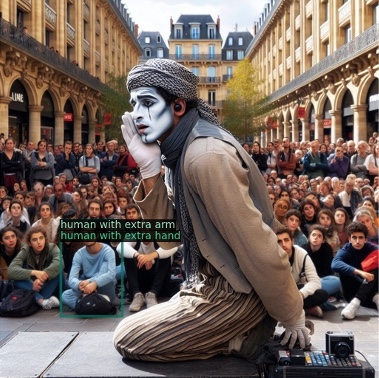}
    \label{fig:global_dalle3_2}
  \end{subfigure}
  \hfill
  \begin{subfigure}[b]{0.24\linewidth}
    \centering
    \includegraphics[width=\linewidth]{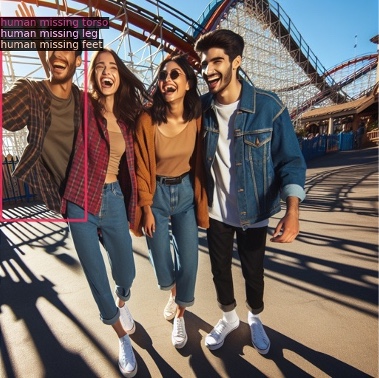}
    \label{fig:global_dalle3_3}
  \end{subfigure}
  \hfill
  \begin{subfigure}[b]{0.24\linewidth}
    \centering
    \includegraphics[width=\linewidth]{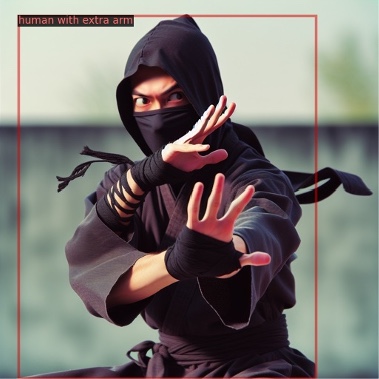}
    \label{fig:global_dalle3_4}
  \end{subfigure}
  \vspace{0.3cm}

  \begin{subfigure}[b]{0.24\linewidth}
    \centering
    \includegraphics[width=\linewidth]{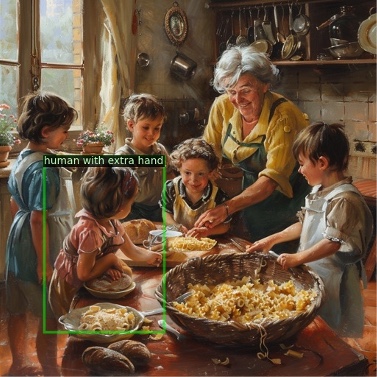}
    \label{fig:global_mj_1}
  \end{subfigure}
  \hfill
  \begin{subfigure}[b]{0.24\linewidth}
    \centering
    \includegraphics[width=\linewidth]{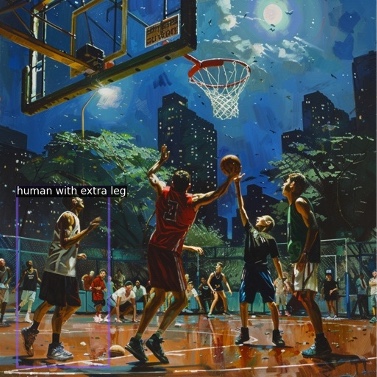}
    \label{fig:global_mj_2}
  \end{subfigure}
  \hfill
  \begin{subfigure}[b]{0.24\linewidth}
    \centering
    \includegraphics[width=\linewidth]{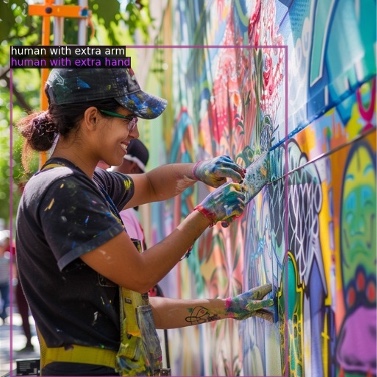}
    \label{fig:global_mj_3}
  \end{subfigure}
  \hfill
  \begin{subfigure}[b]{0.24\linewidth}
    \centering
    \includegraphics[width=\linewidth]{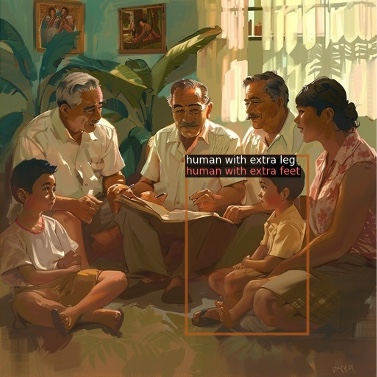}
    \label{fig:global_mj_4}
  \end{subfigure}
  \vspace{0.3cm}

  \caption{More examples of annotations from ~\ourdataset~ for global human artifacts. First row: SDXL. Second row: DALLE-2. Third row: DALLE-3. Last row: Midjourney.}
  \label{fig:more_anno_global}
\end{figure*}

\noindent\textbf{Annotation Examples}
We present additional annotation examples of local human artifacts in Fig.~\ref{fig:more_anno_local} and global human artifacts in Fig.~\ref{fig:more_anno_global}. These examples reveal that in domains where human artifacts are frequent, such as local artifacts from DALLE-2 and global artifacts from SDXL, the artifacts are generally more obvious and distinguishable. Conversely, in domains where artifacts are rare, such as DALLE-3 and Midjourney, the artifacts tend to be more subtle and ambiguous. These observations align with the patterns shown in Fig.~\ref{fig:data_stat} and the discussion in Sec.~\ref{sec:det_exp_discussion}.

\begin{figure*}[t] 
  \centering
  \begin{subfigure}[b]{0.24\linewidth}
    \centering
    \includegraphics[width=\linewidth]{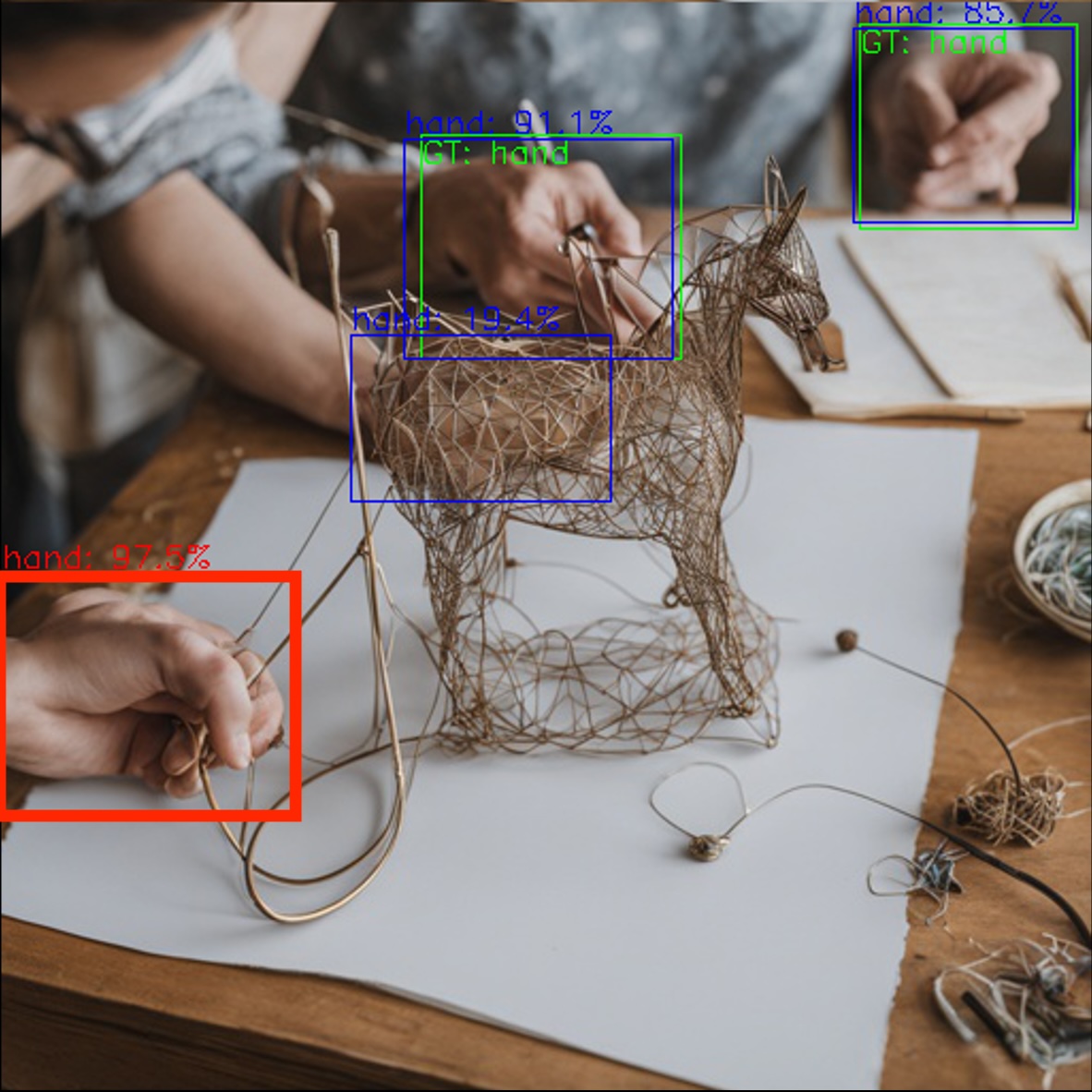}
    \caption{FP hand in SDXL}
    \label{fig:supp_failure_sdxl_1}
  \end{subfigure}
  \hfill
  \begin{subfigure}[b]{0.24\linewidth}
    \centering
    \includegraphics[width=\linewidth]{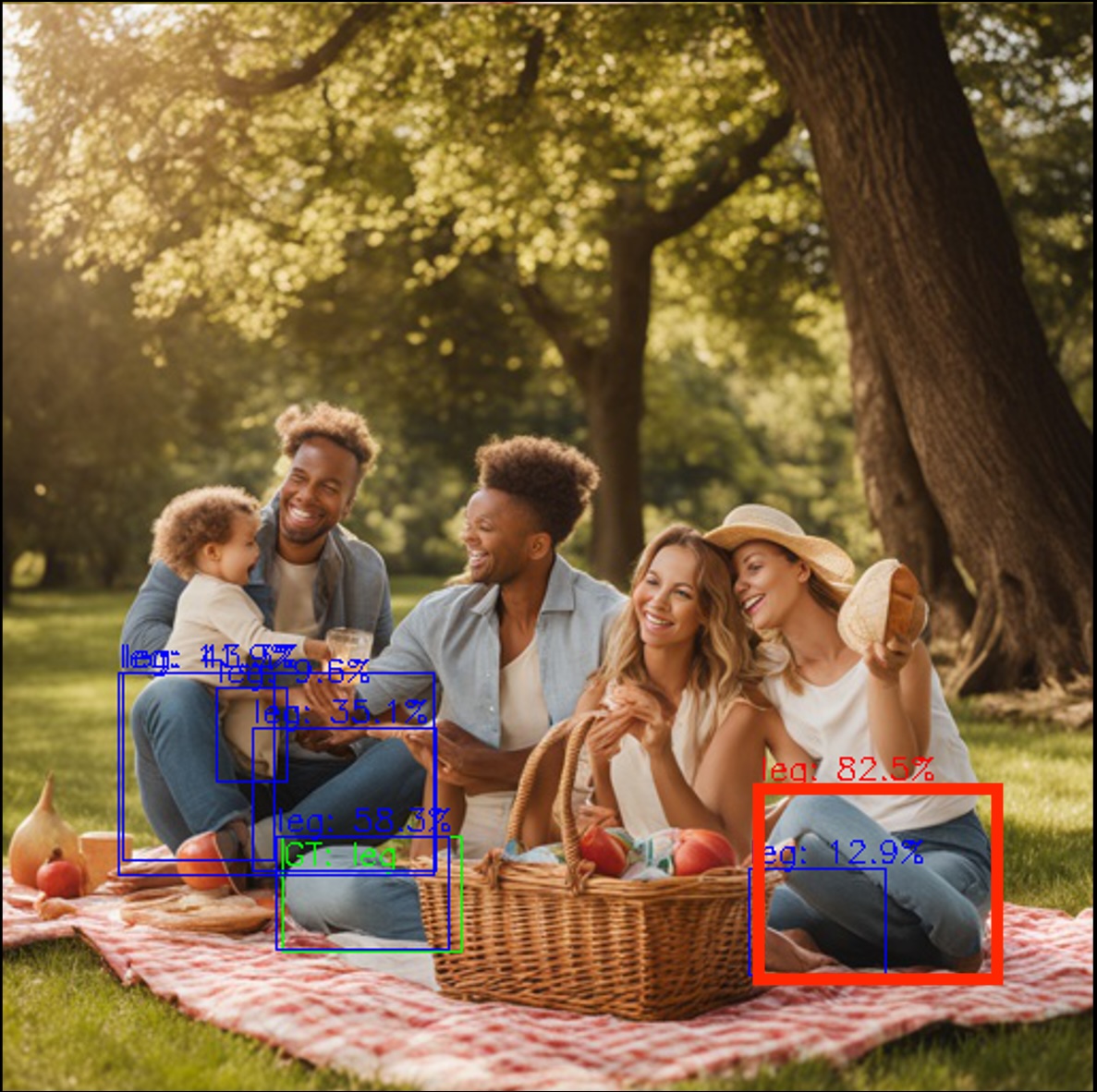}
    \caption{FP leg in SDXL}
    \label{fig:supp_failure_sdxl_2}
  \end{subfigure}
  \hfill
  \begin{subfigure}[b]{0.24\linewidth}
    \centering
    \includegraphics[width=\linewidth]{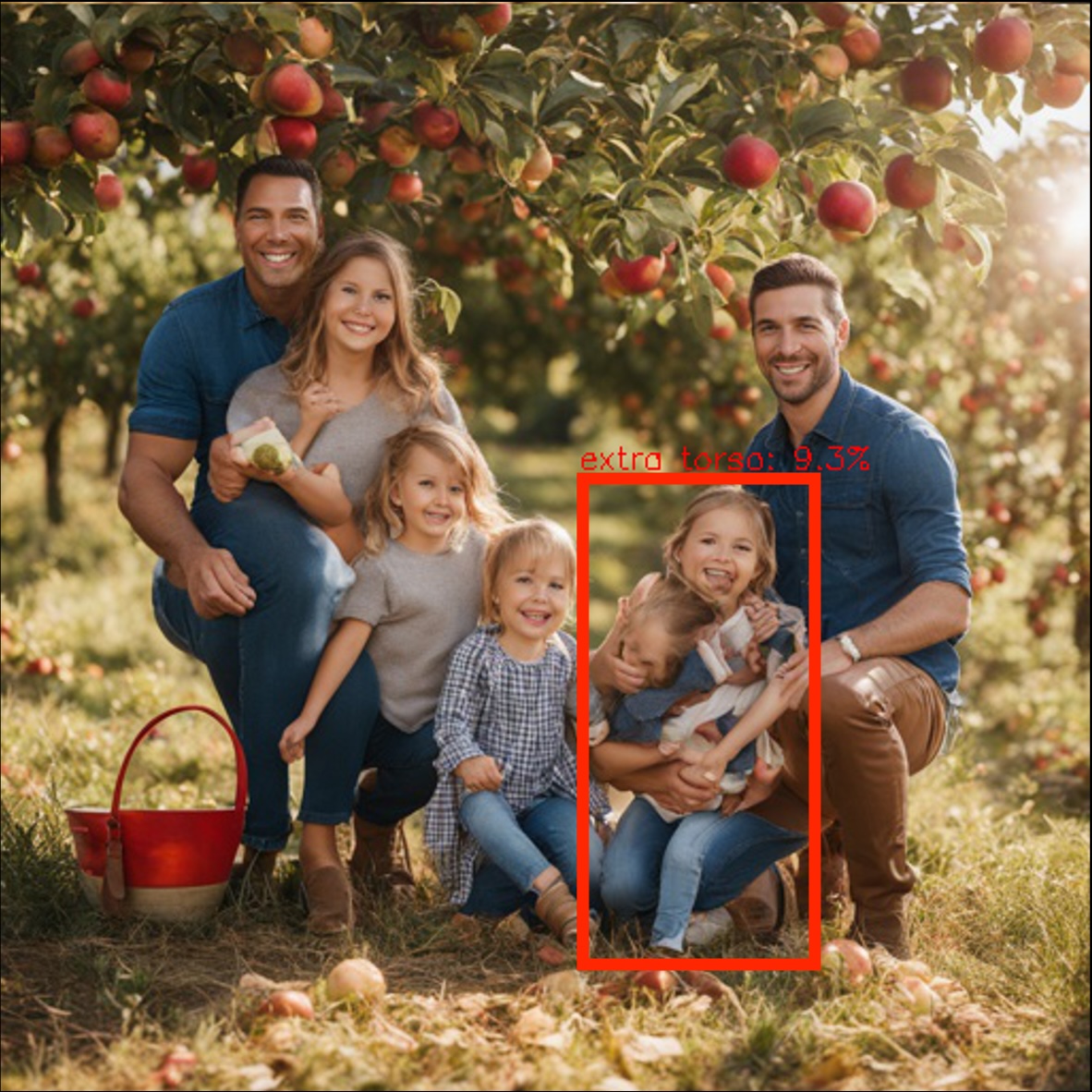}
    \caption{FP extra torso in SDXL}
    \label{fig:supp_failure_sdxl_3}
  \end{subfigure}
  \hfill
  \begin{subfigure}[b]{0.24\linewidth}
    \centering
    \includegraphics[width=\linewidth]{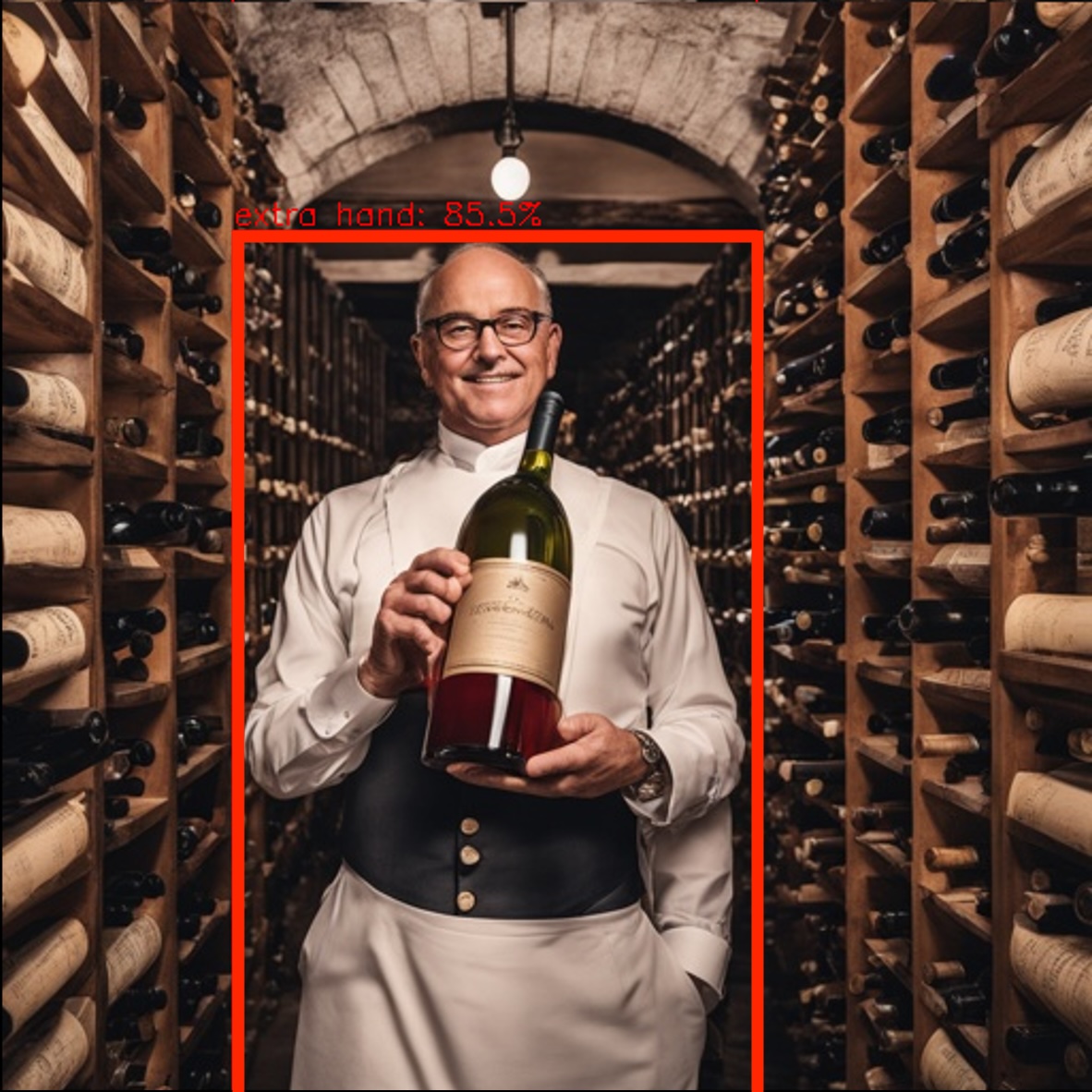}
    \caption{FP extra arm in SDXL}
    \label{fig:supp_failure_sdxl_4}
  \end{subfigure}

  \vspace{0.2cm}

  \begin{subfigure}[b]{0.24\linewidth}
    \centering
    \includegraphics[width=\linewidth]{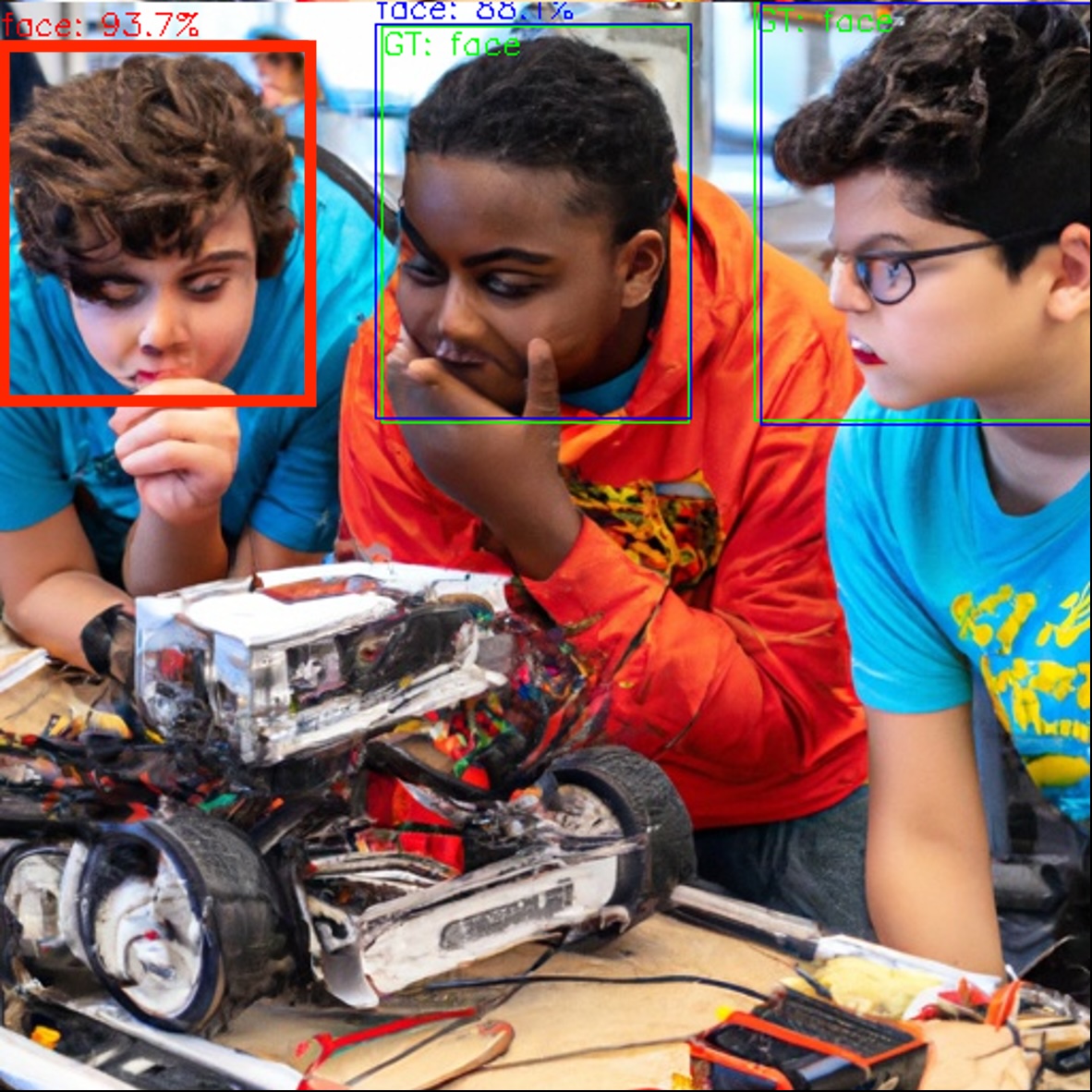}
    \caption{FP face in DALLE-2}
    \label{fig:supp_failure_dalle2_1}
  \end{subfigure}
  \hfill
  \begin{subfigure}[b]{0.24\linewidth}
    \centering
    \includegraphics[width=\linewidth]{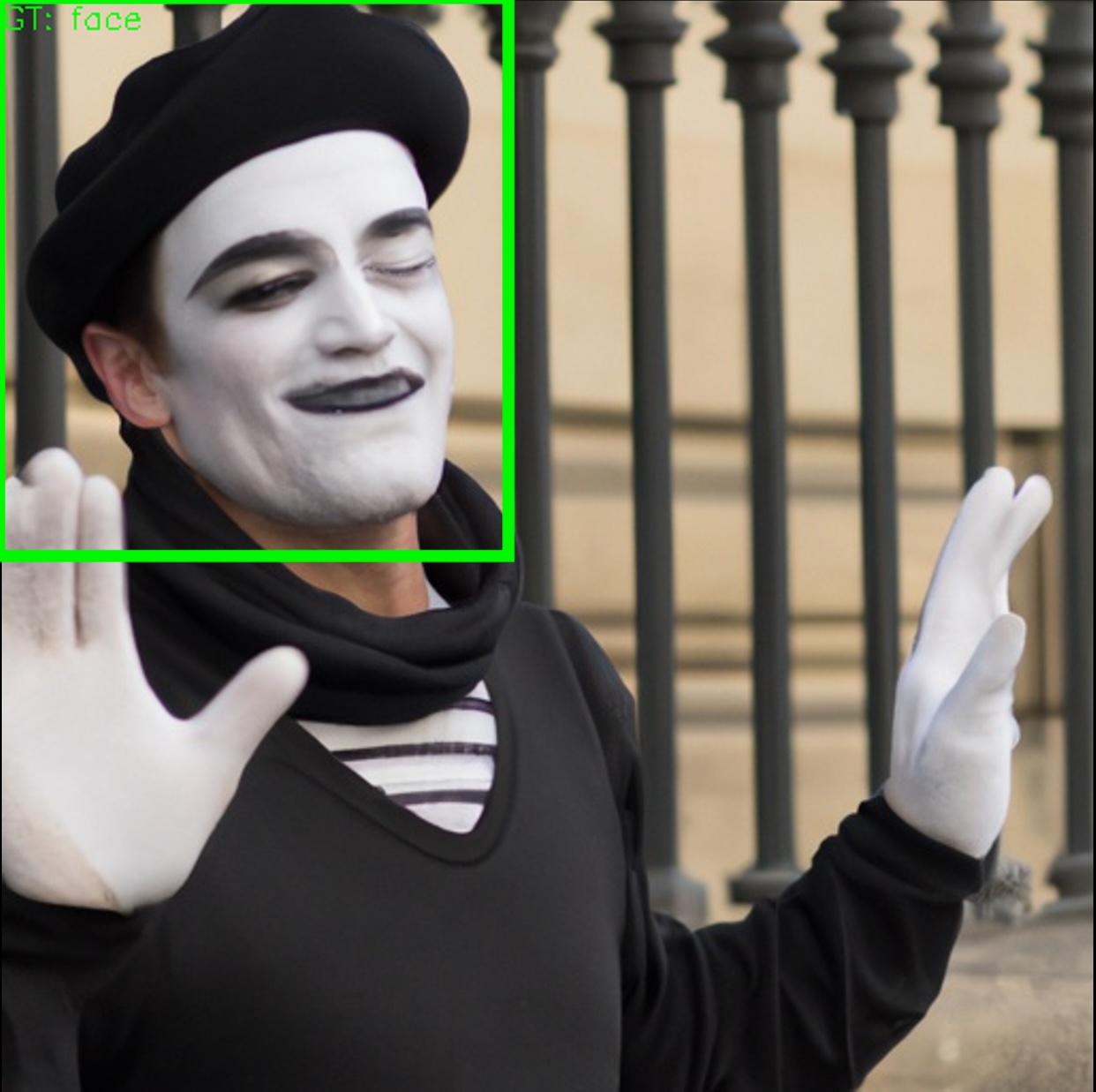}
    \caption{FN face in DALLE-2}
    \label{fig:supp_failure_dalle2_2}
  \end{subfigure}
  \hfill
  \begin{subfigure}[b]{0.24\linewidth}
    \centering
    \includegraphics[width=\linewidth]{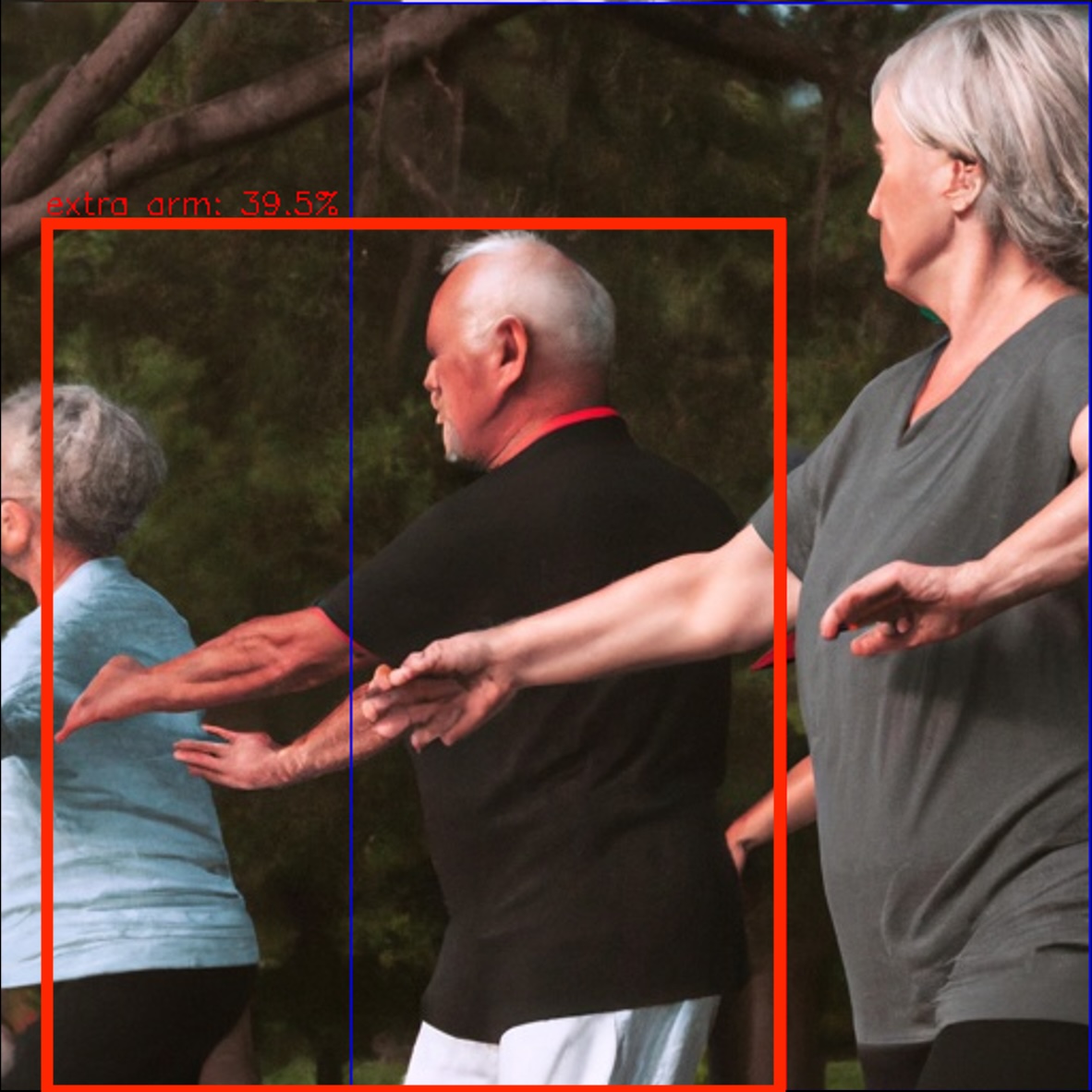}
    \caption{FP extra arm in DALLE-2}
    \label{fig:supp_failure_dalle2_3}
  \end{subfigure}
  \hfill
  \begin{subfigure}[b]{0.24\linewidth}
    \centering
    \includegraphics[width=\linewidth]{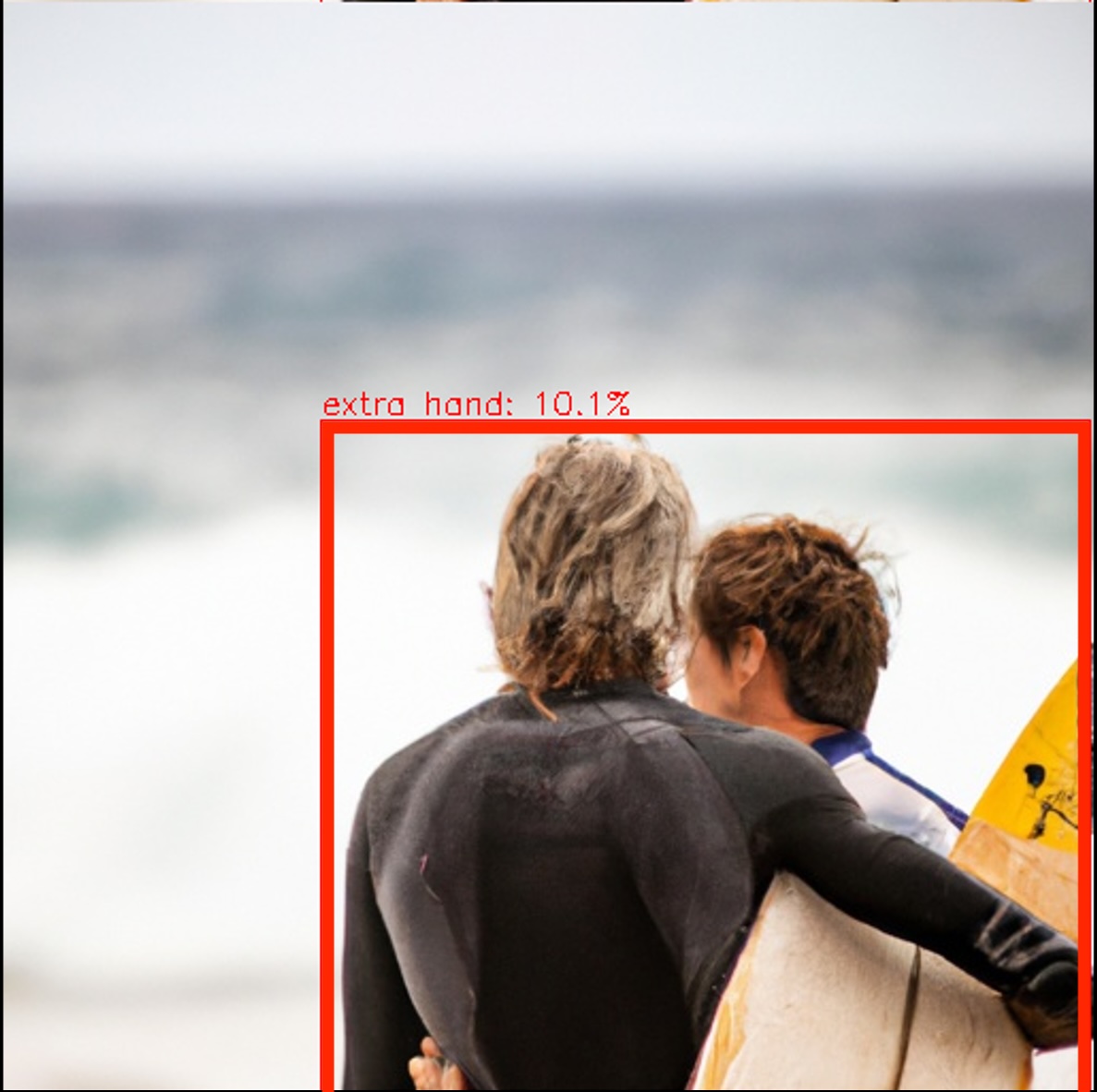}
    \caption{FP extra hand in DALLE-2}
    \label{fig:supp_failure_dalle2_4}
  \end{subfigure}

  \vspace{0.2cm}

  \begin{subfigure}[b]{0.24\linewidth}
    \centering
    \includegraphics[width=\linewidth]{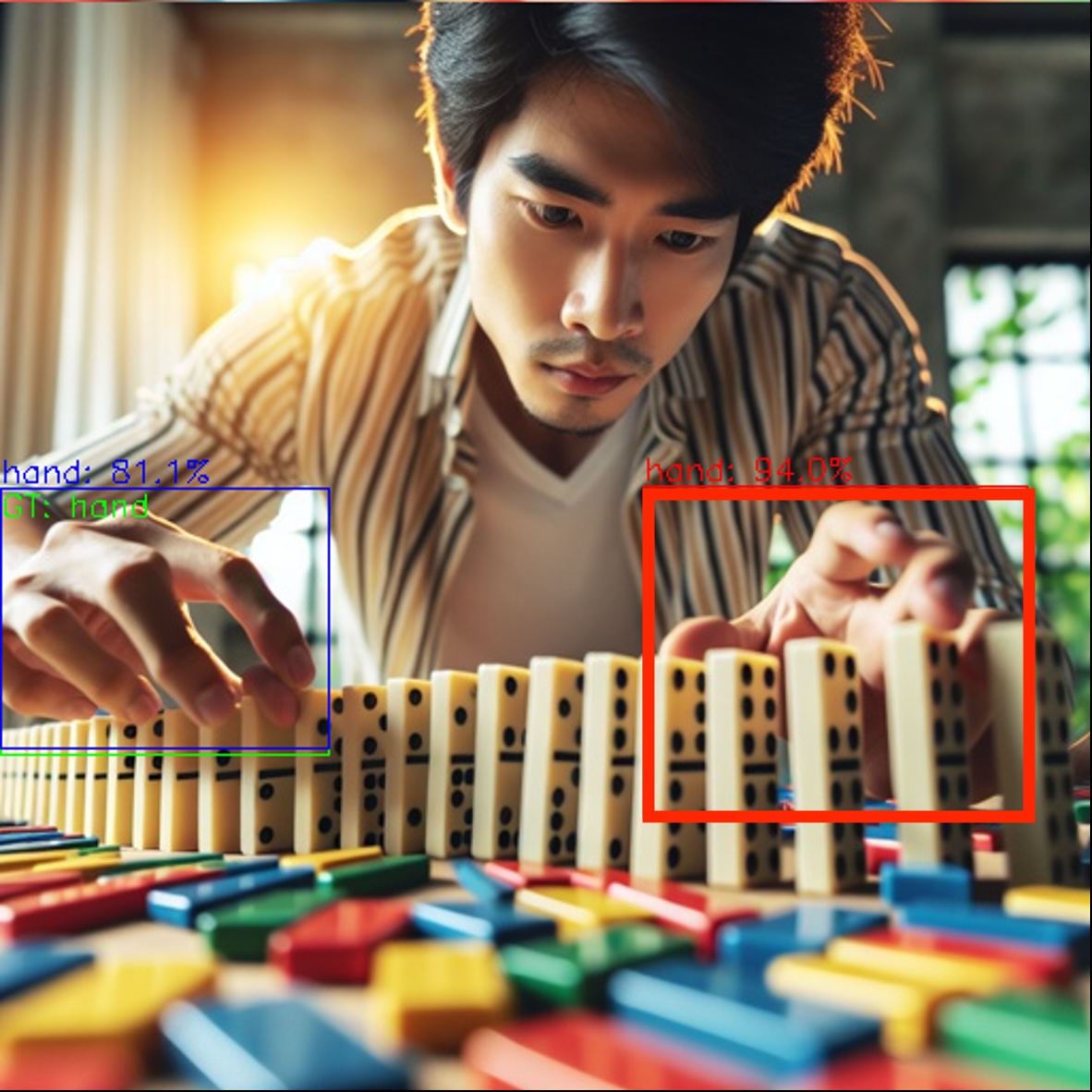}
    \caption{FP hand in DALLE-3}
    \label{fig:supp_failure_dalle3_1}
  \end{subfigure}
  \hfill
  \begin{subfigure}[b]{0.24\linewidth}
    \centering
    \includegraphics[width=\linewidth]{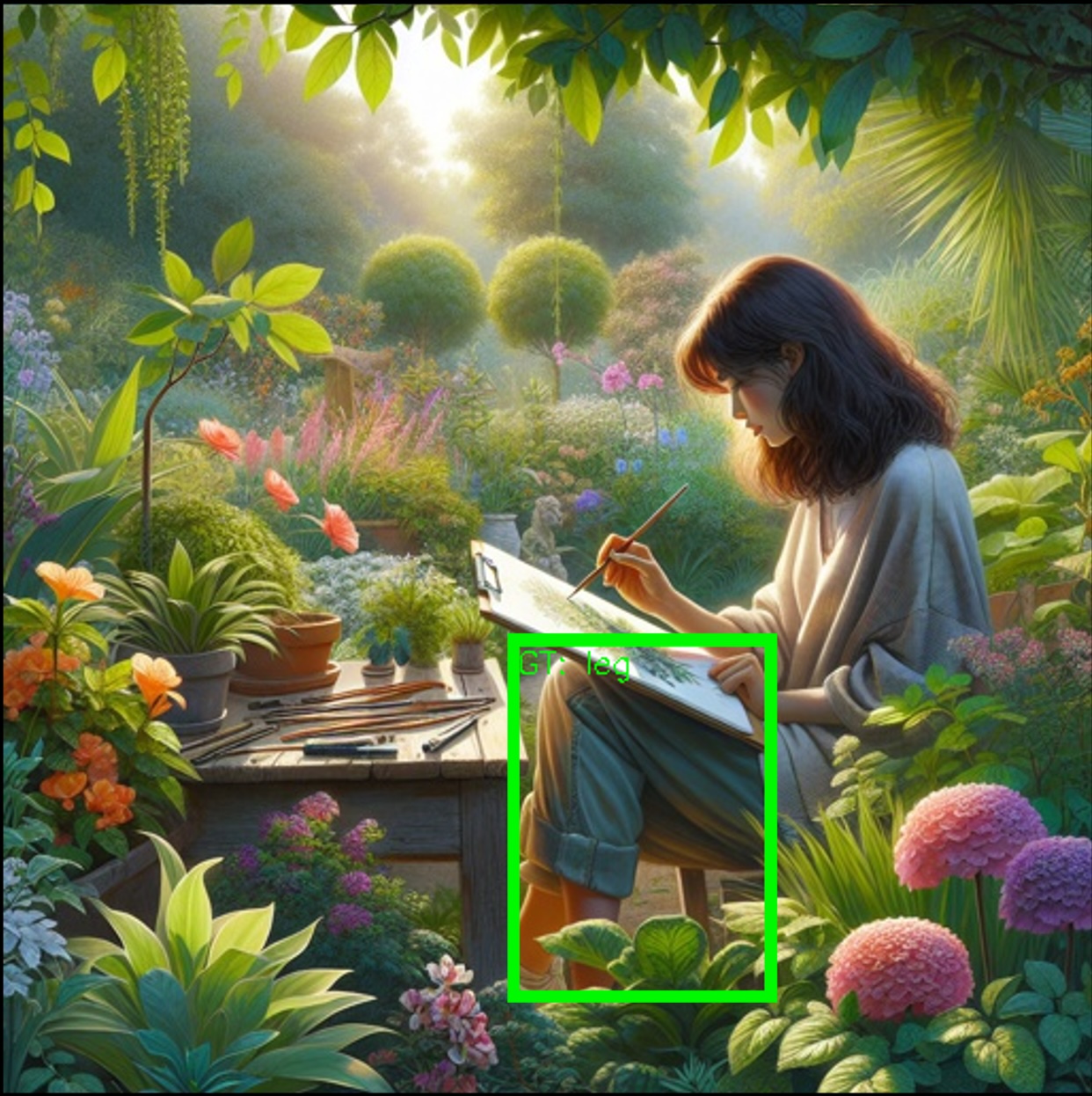}
    \caption{FN leg in DALLE-3}
    \label{fig:supp_failure_dalle3_2}
  \end{subfigure}
  \hfill
  \begin{subfigure}[b]{0.24\linewidth}
    \centering
    \includegraphics[width=\linewidth]{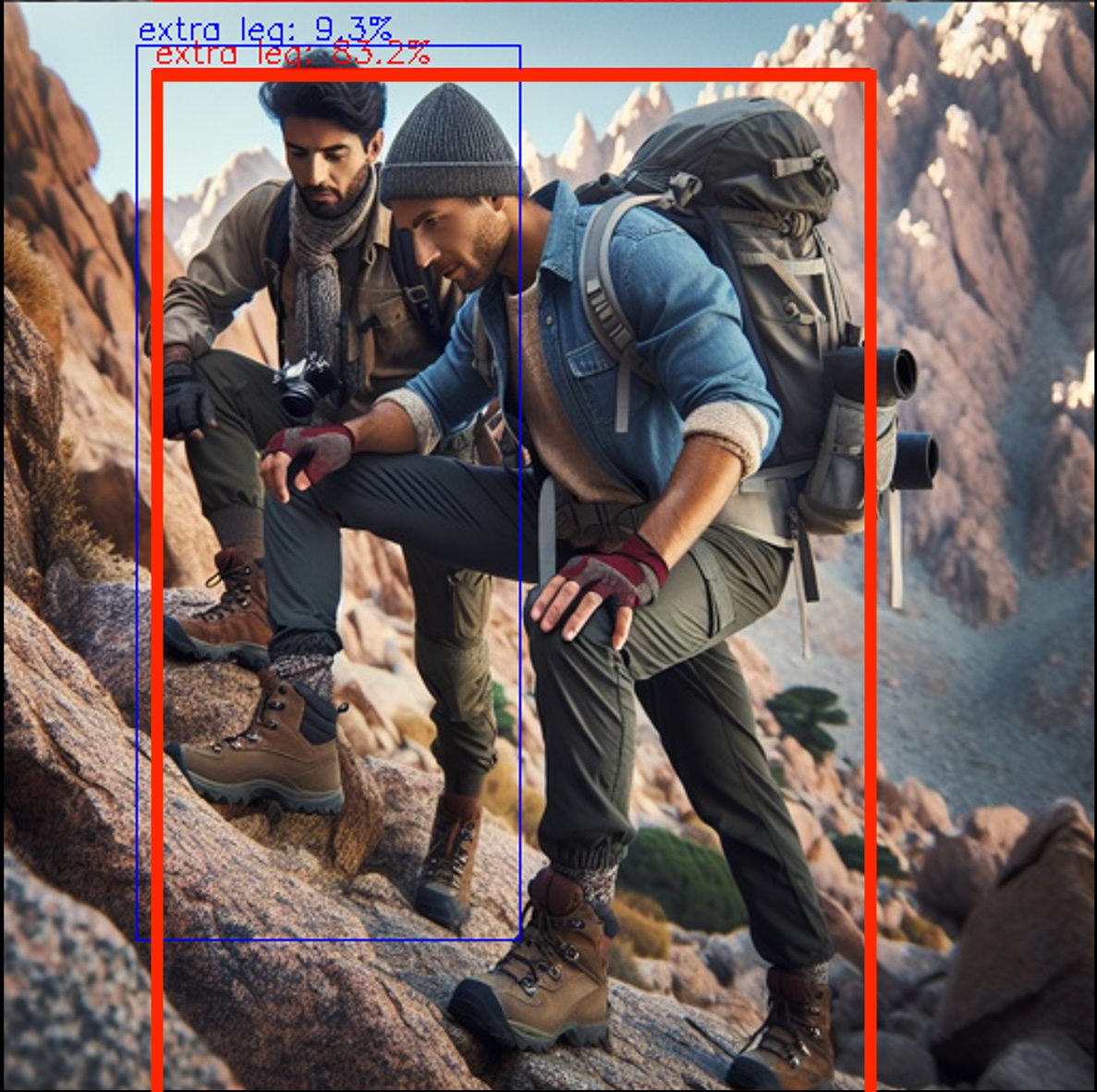}
    \caption{FP extra leg in DALLE-3}
    \label{fig:supp_failure_dalle3_3}
  \end{subfigure}
  \hfill
  \begin{subfigure}[b]{0.24\linewidth}
    \centering
    \includegraphics[width=\linewidth]{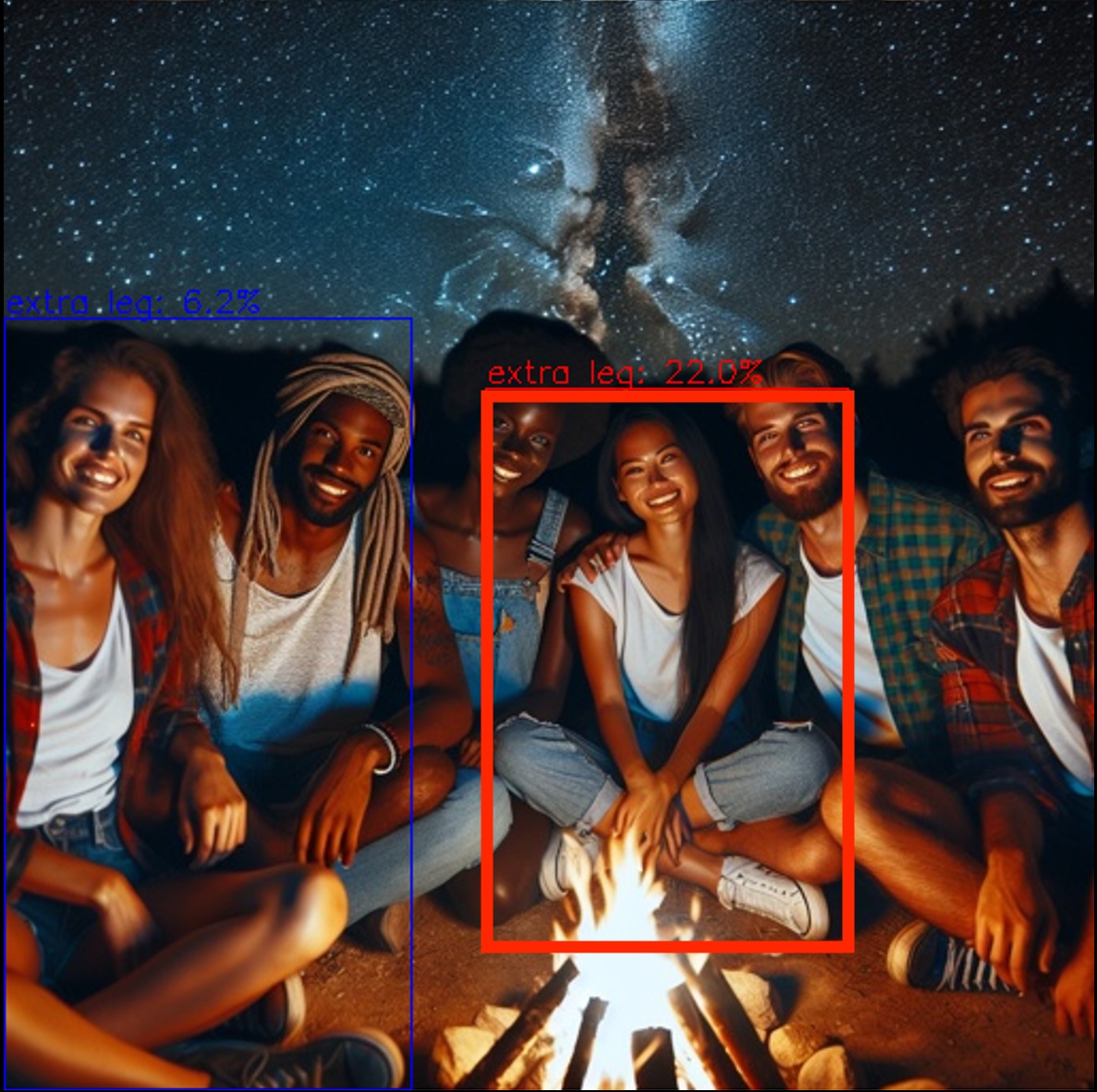}
    \caption{FP extra leg in DALLE-3}
    \label{fig:supp_failure_dalle3_4}
  \end{subfigure}

  \vspace{0.2cm}

  \begin{subfigure}[b]{0.24\linewidth}
    \centering
    \includegraphics[width=\linewidth]{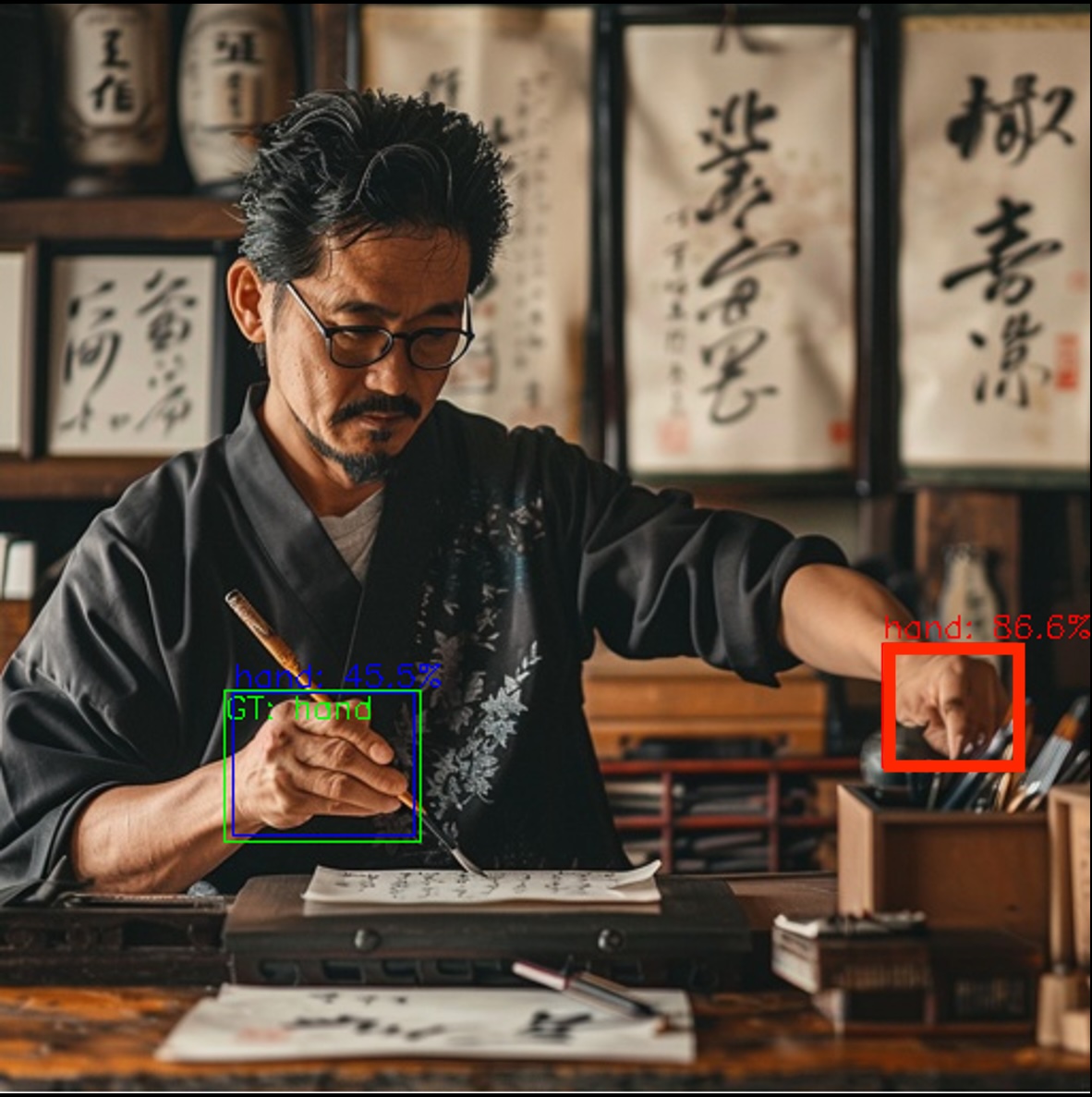}
    \caption{FP hand in Midjourney}
    \label{fig:supp_failure_mj_1}
  \end{subfigure}
  \hfill
  \begin{subfigure}[b]{0.24\linewidth}
    \centering
    \includegraphics[width=\linewidth]{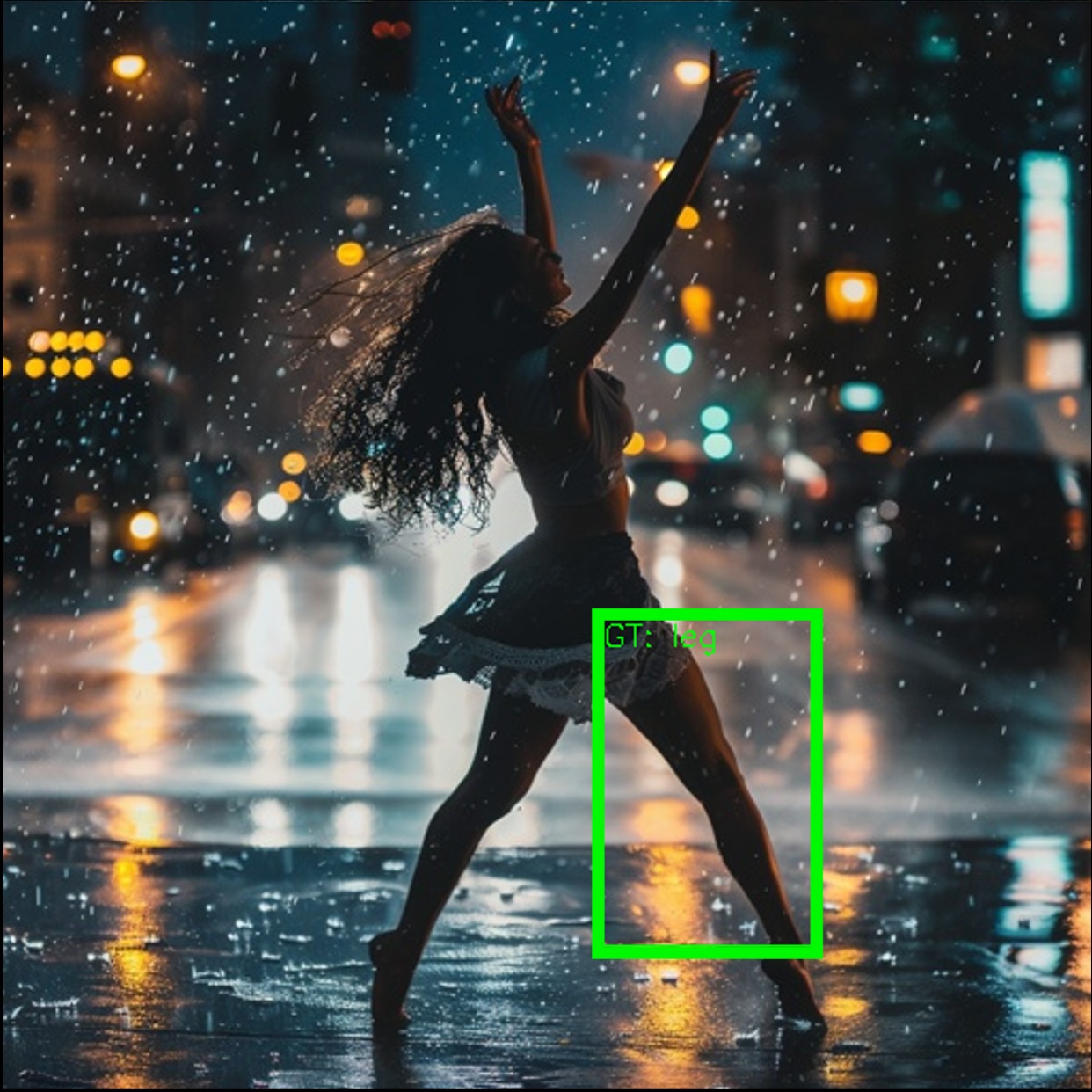}
    \caption{FN leg in Midjourney}
    \label{fig:supp_failure_mj_2}
  \end{subfigure}
  \hfill
  \begin{subfigure}[b]{0.24\linewidth}
    \centering
    \includegraphics[width=\linewidth]{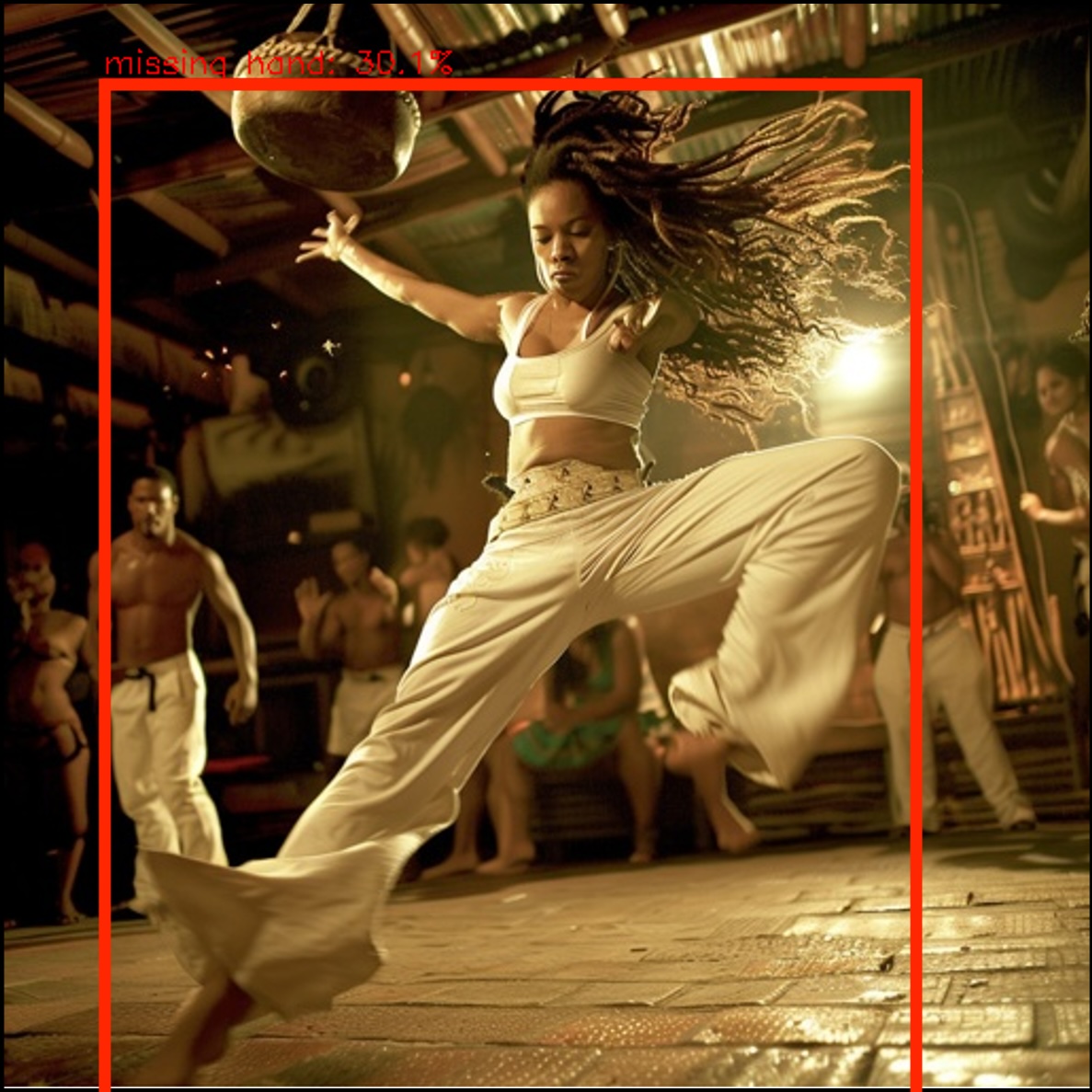}
    \caption{FP missing hand in Midjourney}
    \label{fig:supp_failure_mj_3}
  \end{subfigure}
  \hfill
  \begin{subfigure}[b]{0.24\linewidth}
    \centering
    \includegraphics[width=\linewidth]{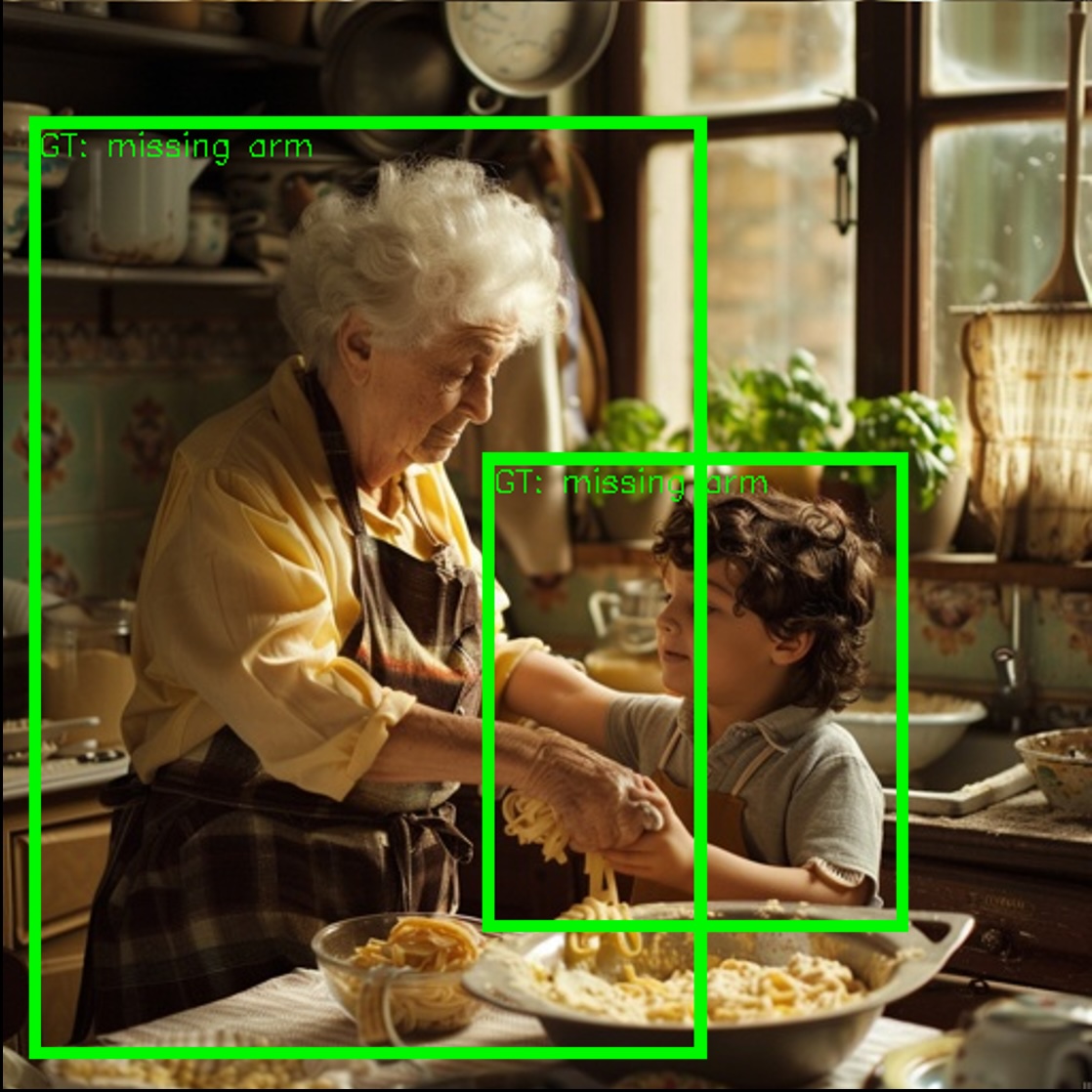}
    \caption{FN missing arm in Midjourney}
    \label{fig:supp_failure_mj_4}
  \end{subfigure}
  \vspace{0.2cm}

  \caption{Examples of predictions from our ~\abbrourmodel~ considered mistakes during evaluation on SDXL (first row), DALLE-2 (second row), DALLE-3 (third row), and Midjourney (last row). 
  FP: false positive, FN: false negative.
  Red bounding boxes represent the detected artifact with top prediction scores, blue bounding boxes represent other detected bounding boxes with the same class label, and green bounding boxes represent the ground truth.
  }
  \label{fig:supp_failure}
\end{figure*}

\noindent\textbf{Failure Cases}
We present additional results from ~\abbrourmodel~ on the validation set of ~\abbrourdataset~ that are considered incorrect, including false positives and false negatives, in Fig.~\ref{fig:supp_failure}. These results reveal that most of the predictions are reasonable and are classified as errors partly due to occasional oversight by annotators or the growing difficulty in making judgments as the generator’s capacity improves, as discussed in Sec.~\ref{sec:det_exp_discussion}.

\noindent\textbf{Results on Unseen Domains}
We present additional results from ~\abbrourmodel~ on images from unseen domains. Our models demonstrate strong generalization to these domains, effectively identifying images generated by both a legacy generator (SD1.4) and more advanced generators (PixArt-\(\Sigma\) and FLUX.1-dev). For real images (300W), our models make only a few predictions, primarily due to rare and unusual factors such as atypical poses or challenging lighting conditions.

\noindent\textbf{Samples from Finetuned SDXL}
We present a comparison of images generated by the original and our finetuned SDXL models in Fig.\ref{fig:supp_ft}. As demonstrated by these images and the quantitative results in Fig.\ref{sec:ft_quant}, our finetuned SDXL model exhibits a stronger perception of human artifacts, thanks to our finetuning approach that leverages predictions from ~\abbrourmodel.

\begin{figure*}[t] 
\vspace{-0.5cm}
  \centering
  \begin{subfigure}[b]{0.22\linewidth}
    \centering
    \includegraphics[width=\linewidth]{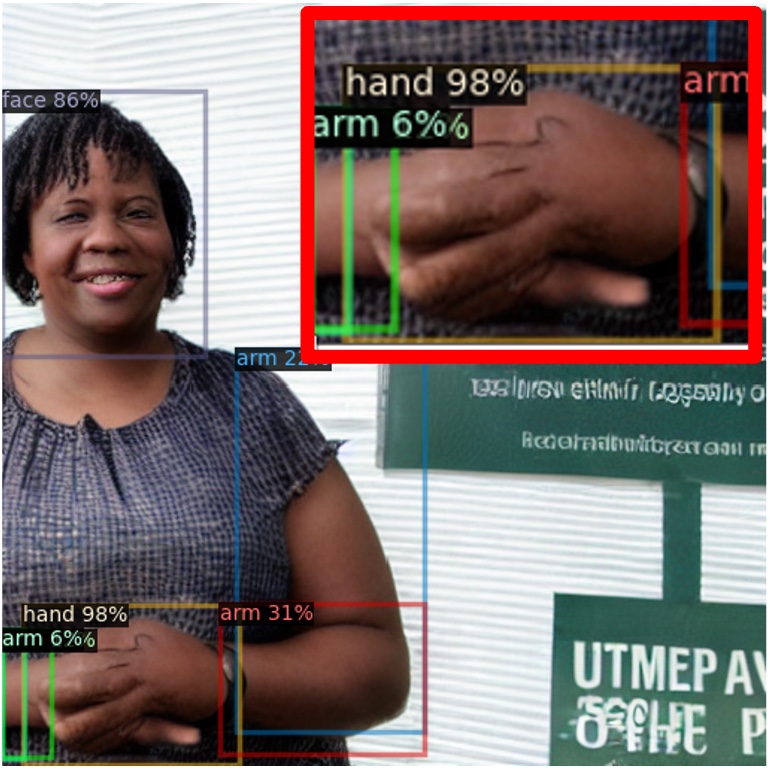}
    \caption{Weird hand in SD1.4}
    \label{fig:supp_ood_sd14_1}
  \end{subfigure}
  \hfill
  \begin{subfigure}[b]{0.22\linewidth}
    \centering
    \includegraphics[width=\linewidth]{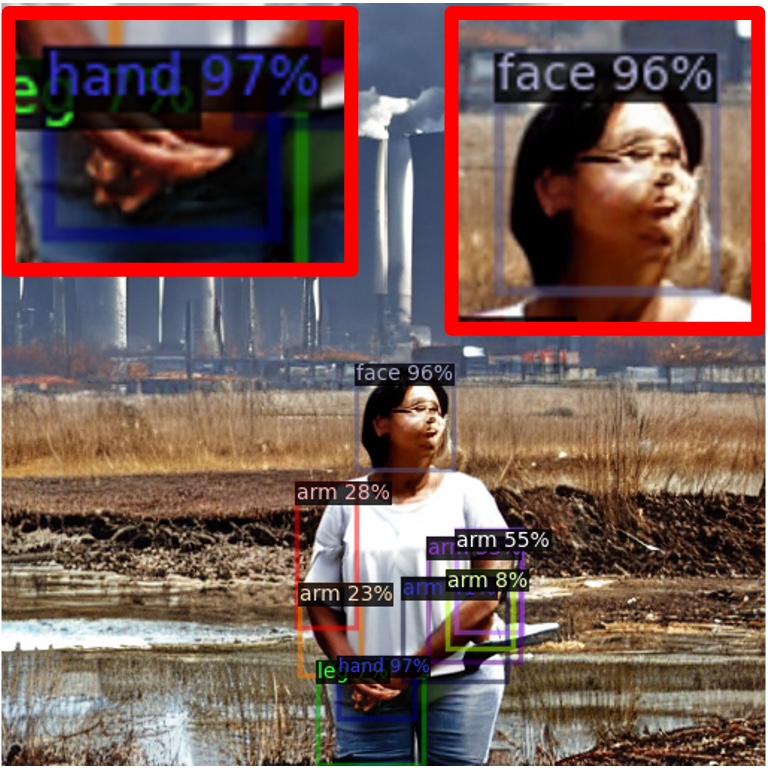}
    \caption{Weird hand \& face in SD1.4}
    \label{fig:supp_ood_sd14_2}
  \end{subfigure}
  \hfill
  \begin{subfigure}[b]{0.22\linewidth}
    \centering
    \includegraphics[width=\linewidth]{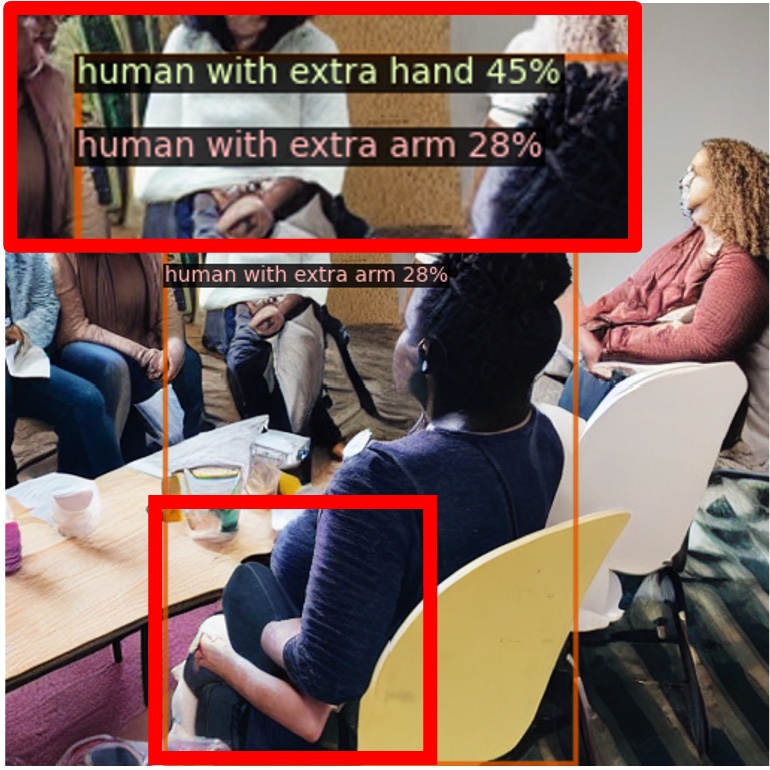}
    \caption{Extra hand \& arm in SD1.4}
    \label{fig:supp_ood_sd14_3}
  \end{subfigure}
  \hfill
  \begin{subfigure}[b]{0.22\linewidth}
    \centering
    \includegraphics[width=\linewidth]{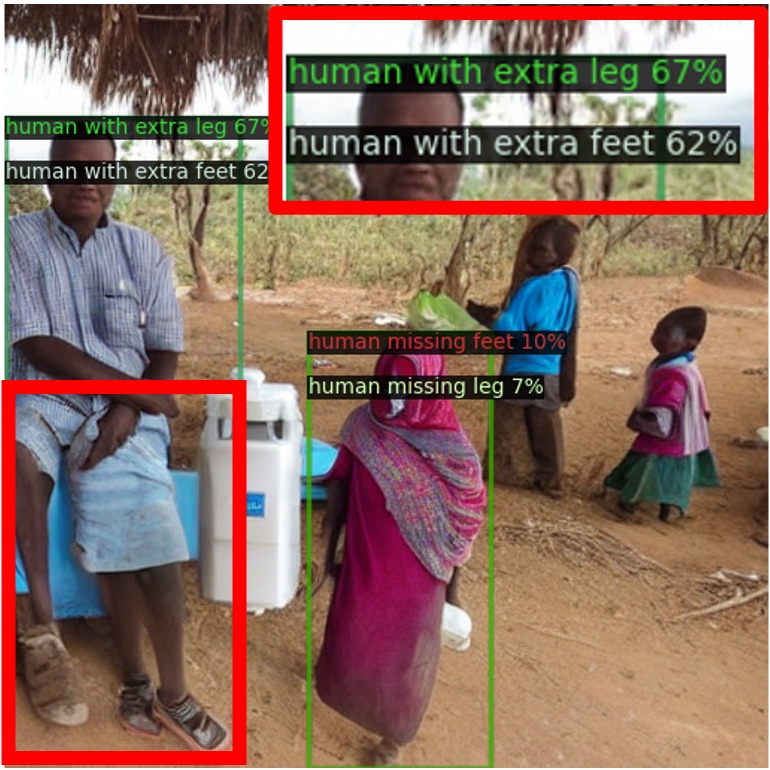}
    \caption{Extra feet \& leg in SD1.4}
    \label{fig:supp_ood_sd14_4}
  \end{subfigure}

\vspace{0.2cm}

  \begin{subfigure}[b]{0.22\linewidth}
    \centering
    \includegraphics[width=\linewidth]{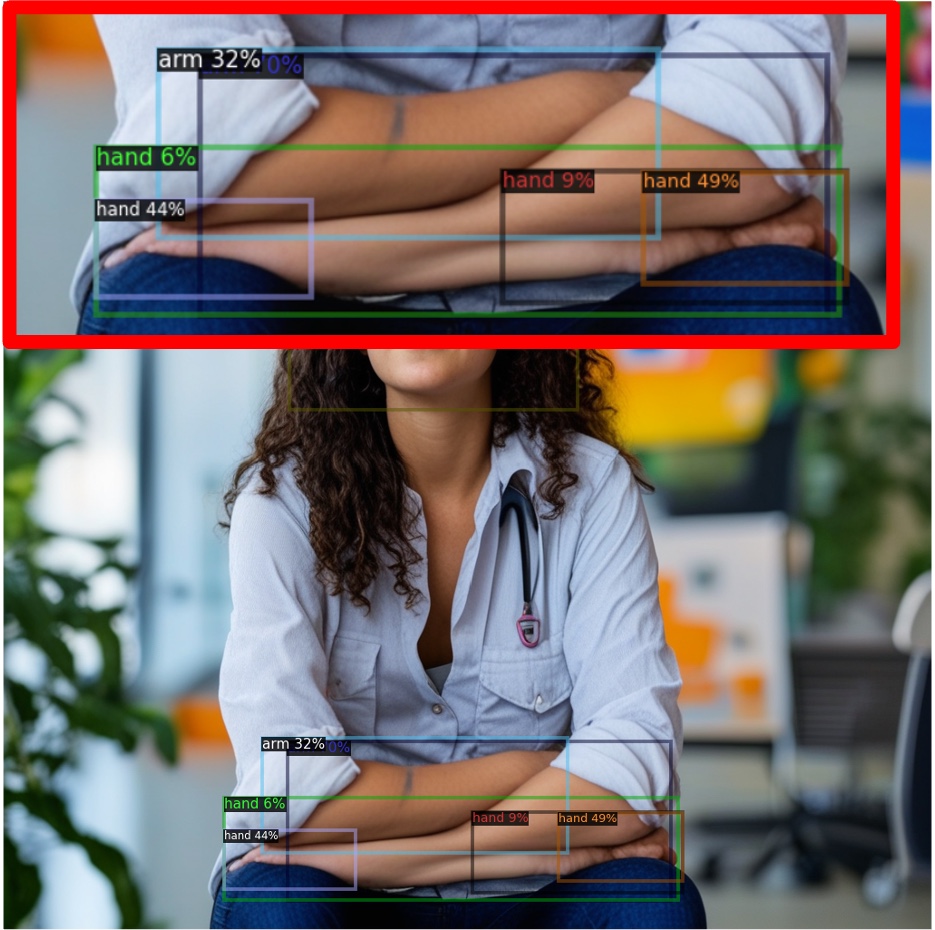}
    \caption{Weird arm in PixArt-\(\Sigma\)}
    \label{fig:supp_ood_pix_1}
  \end{subfigure}
  \hfill
  \begin{subfigure}[b]{0.22\linewidth}
    \centering
    \includegraphics[width=\linewidth]{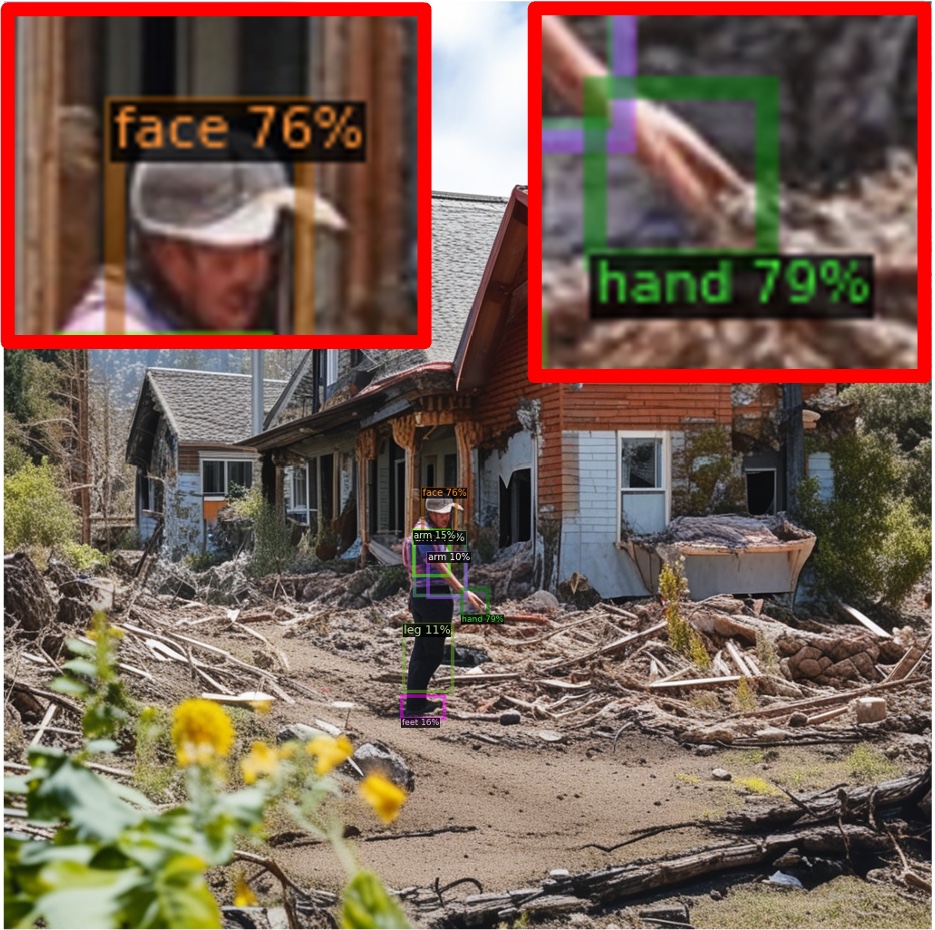}
    \caption{Weird face \& arm in PixArt-\(\Sigma\)}
    \label{fig:supp_ood_pix_2}
  \end{subfigure}
  \hfill
  \begin{subfigure}[b]{0.22\linewidth}
    \centering
    \includegraphics[width=\linewidth]{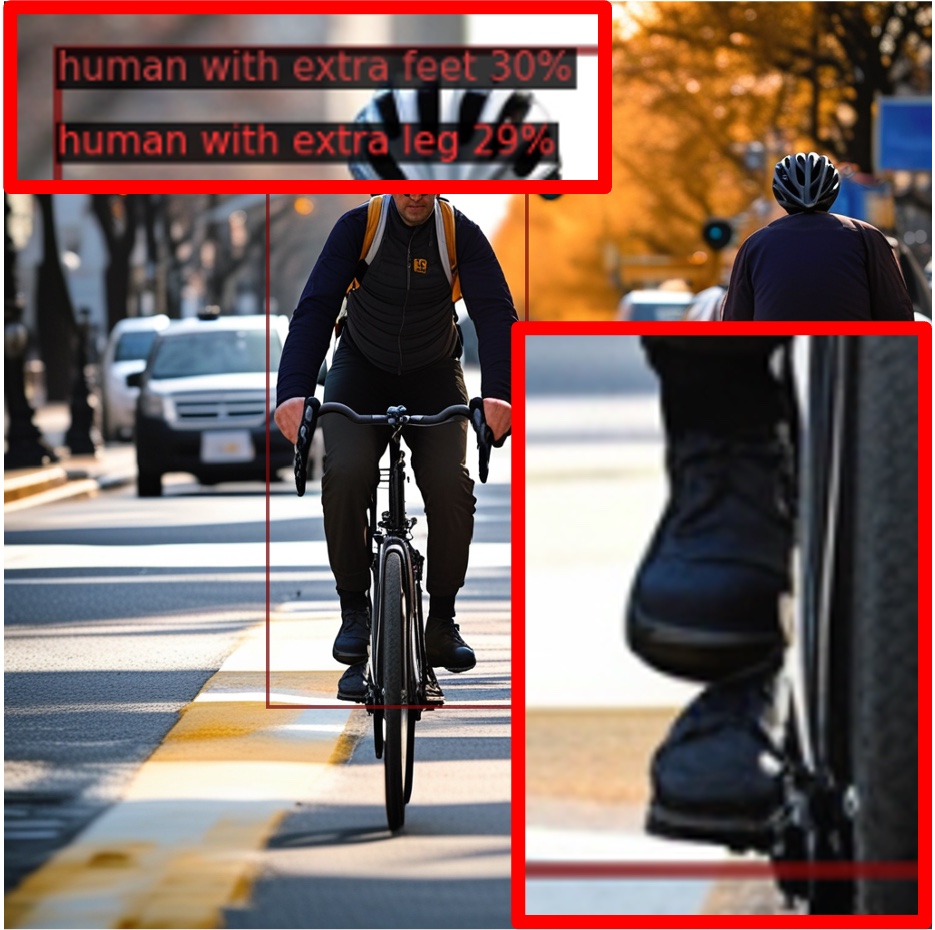}
    \caption{Extra feet \& leg in PixArt-\(\Sigma\)}
    \label{fig:supp_ood_pix_3}
  \end{subfigure}
  \hfill
  \begin{subfigure}[b]{0.22\linewidth}
    \centering
    \includegraphics[width=\linewidth]{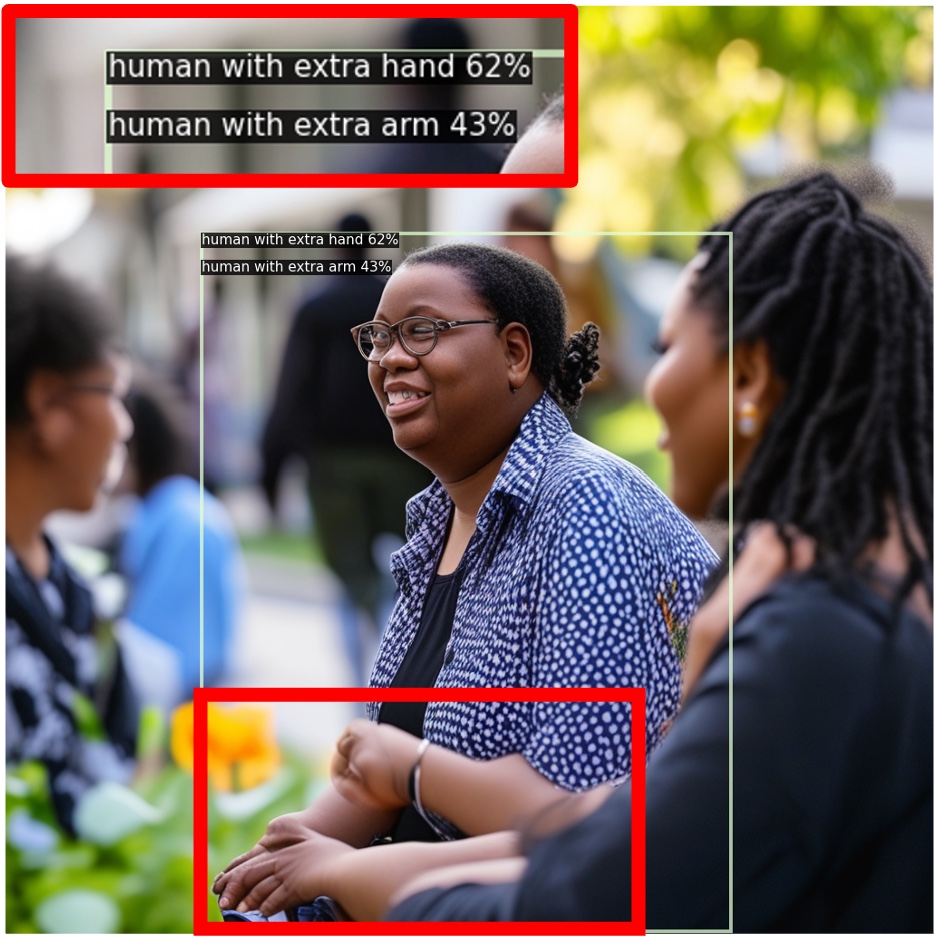}
    \caption{Extra hand \& arm in PixArt-\(\Sigma\)}
    \label{fig:supp_ood_pix_4}
  \end{subfigure}

\vspace{0.2cm}

  \begin{subfigure}[b]{0.22\linewidth}
    \centering
    \includegraphics[width=\linewidth]{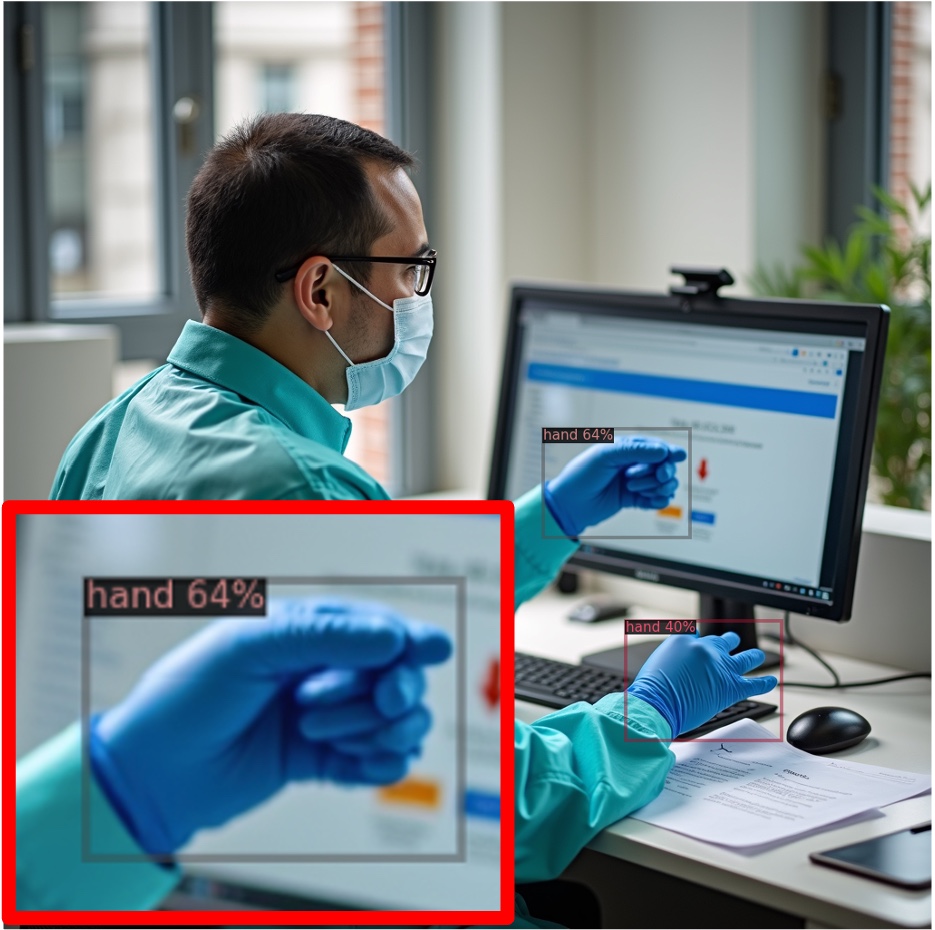}
    \caption{Weird hand in FLUX.1-dev}
    \label{fig:supp_ood_flux_1}
  \end{subfigure}
  \hfill
  \begin{subfigure}[b]{0.22\linewidth}
    \centering
    \includegraphics[width=\linewidth]{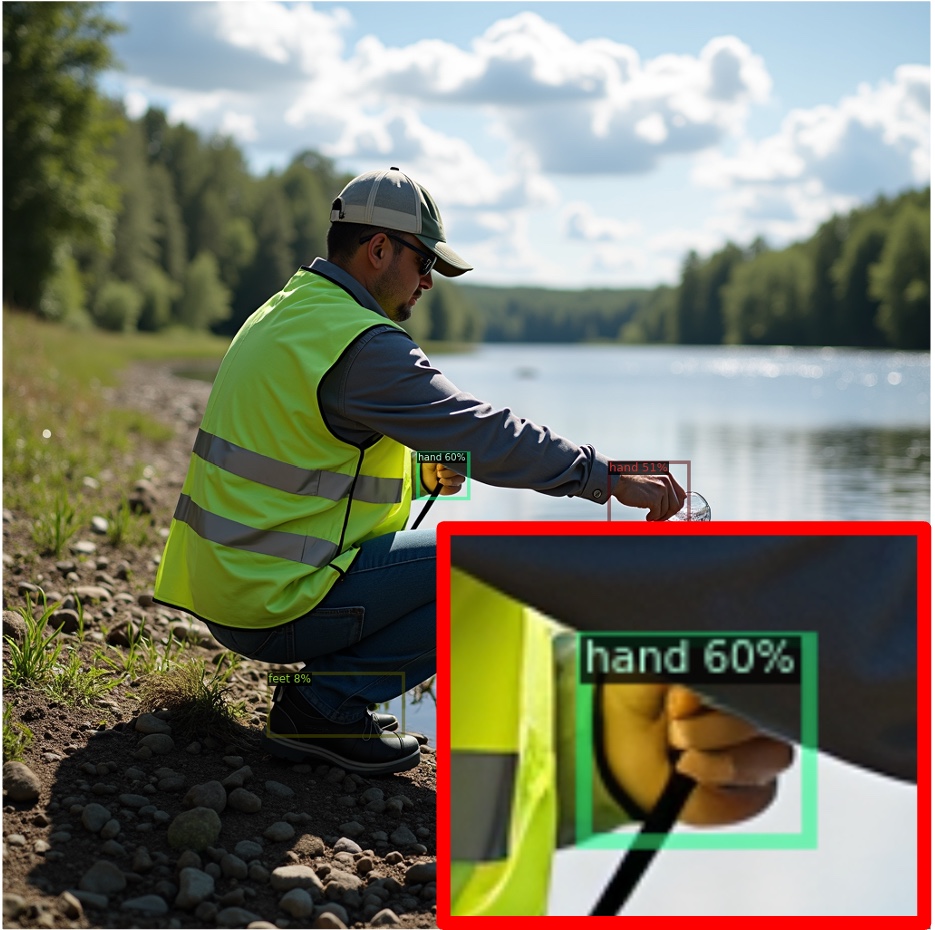}
    \caption{Weird hand in FLUX.1-dev}
    \label{fig:supp_ood_flux_2}
  \end{subfigure}
  \hfill
  \begin{subfigure}[b]{0.22\linewidth}
    \centering
    \includegraphics[width=\linewidth]{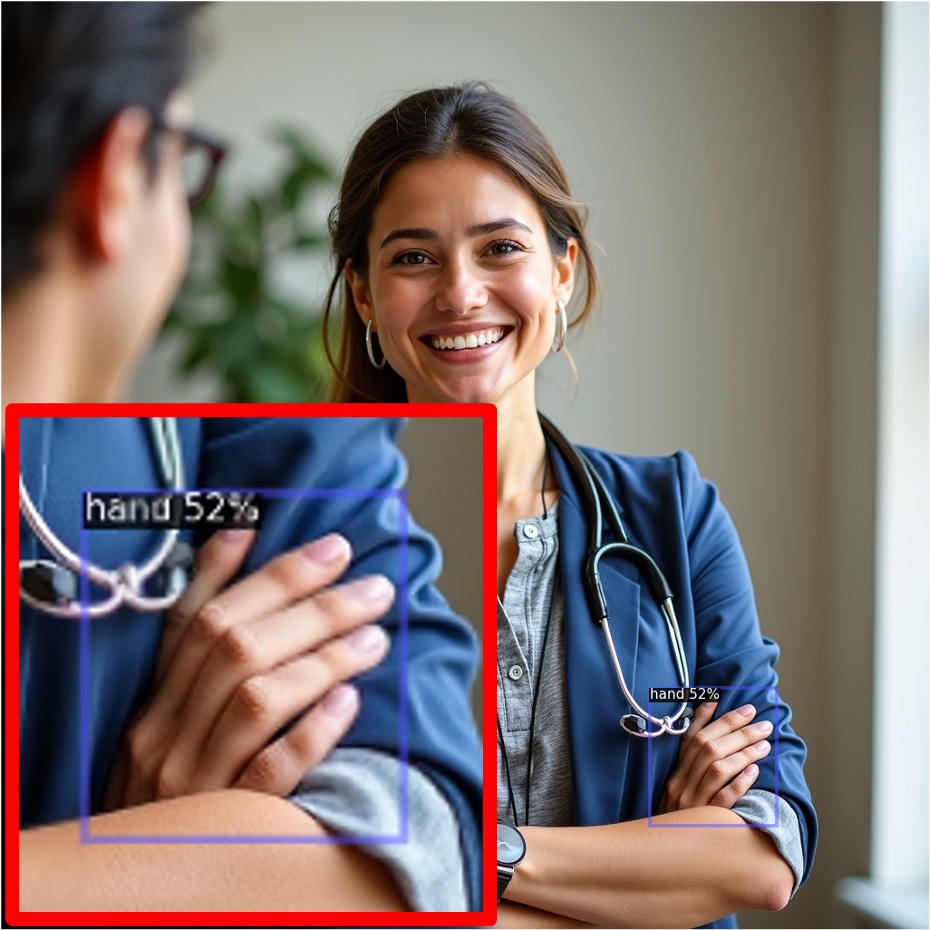}
    \caption{Weird hand in FLUX.1-dev}
    \label{fig:supp_ood_flux_3}
  \end{subfigure}
  \hfill
  \begin{subfigure}[b]{0.22\linewidth}
    \centering
    \includegraphics[width=\linewidth]{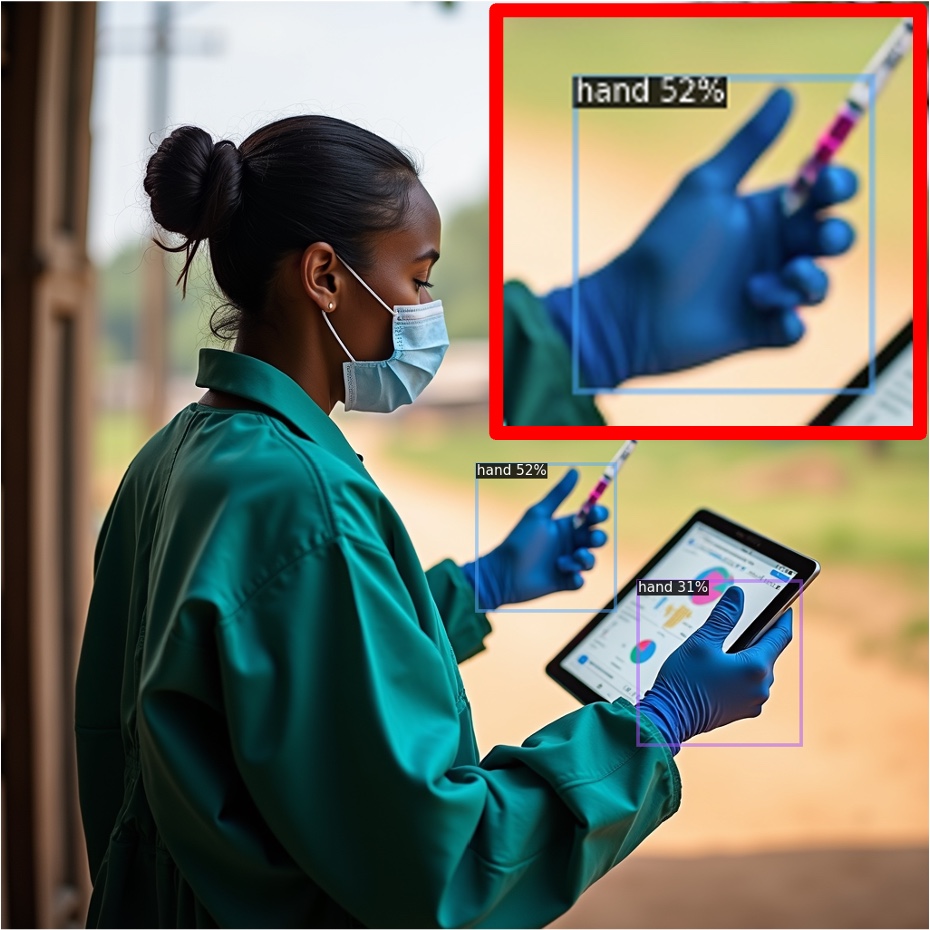}
    \caption{Weird hand in FLUX.1-dev}
    \label{fig:supp_ood_flux_4}
  \end{subfigure}

\vspace{0.2cm}

  \begin{subfigure}[b]{0.22\linewidth}
    \centering
    \includegraphics[width=\linewidth]{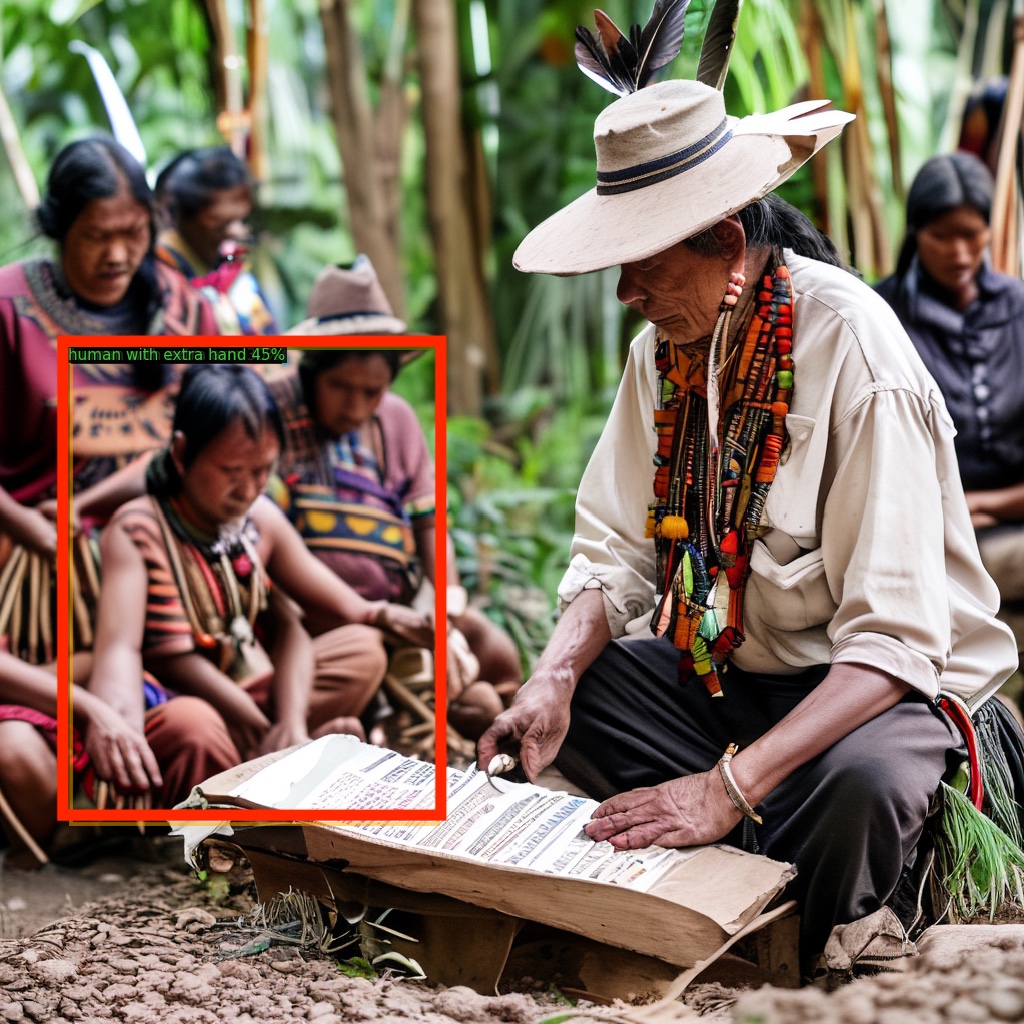}
    \caption{Extra hand \& arm in Sana}
    \label{fig:supp_ood_sana_1}
  \end{subfigure}
  \hfill
  \begin{subfigure}[b]{0.22\linewidth}
    \centering
    \includegraphics[width=\linewidth]{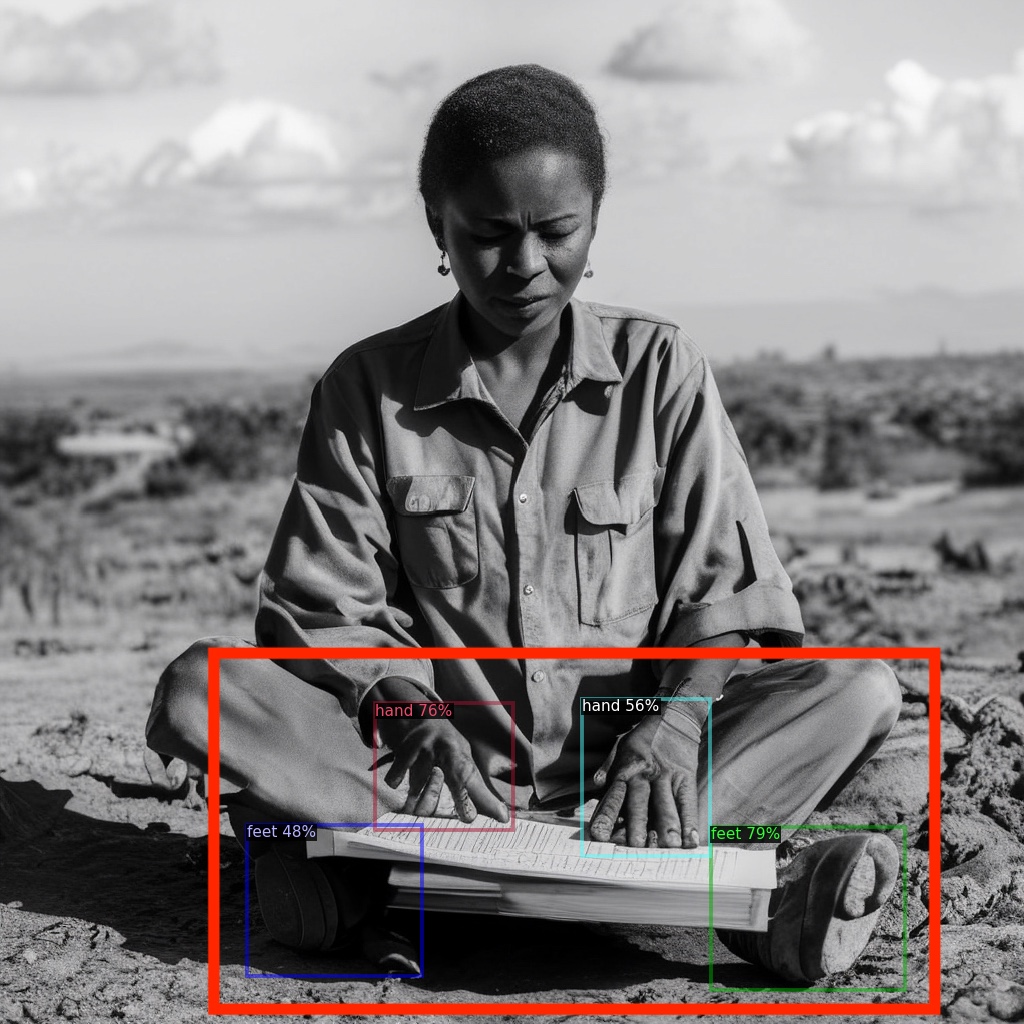}
    \caption{Weird hand \& feet in Sana}
    \label{fig:supp_ood_sana_2}
  \end{subfigure}
  \hfill
  \begin{subfigure}[b]{0.22\linewidth}
    \centering
    \includegraphics[width=\linewidth]{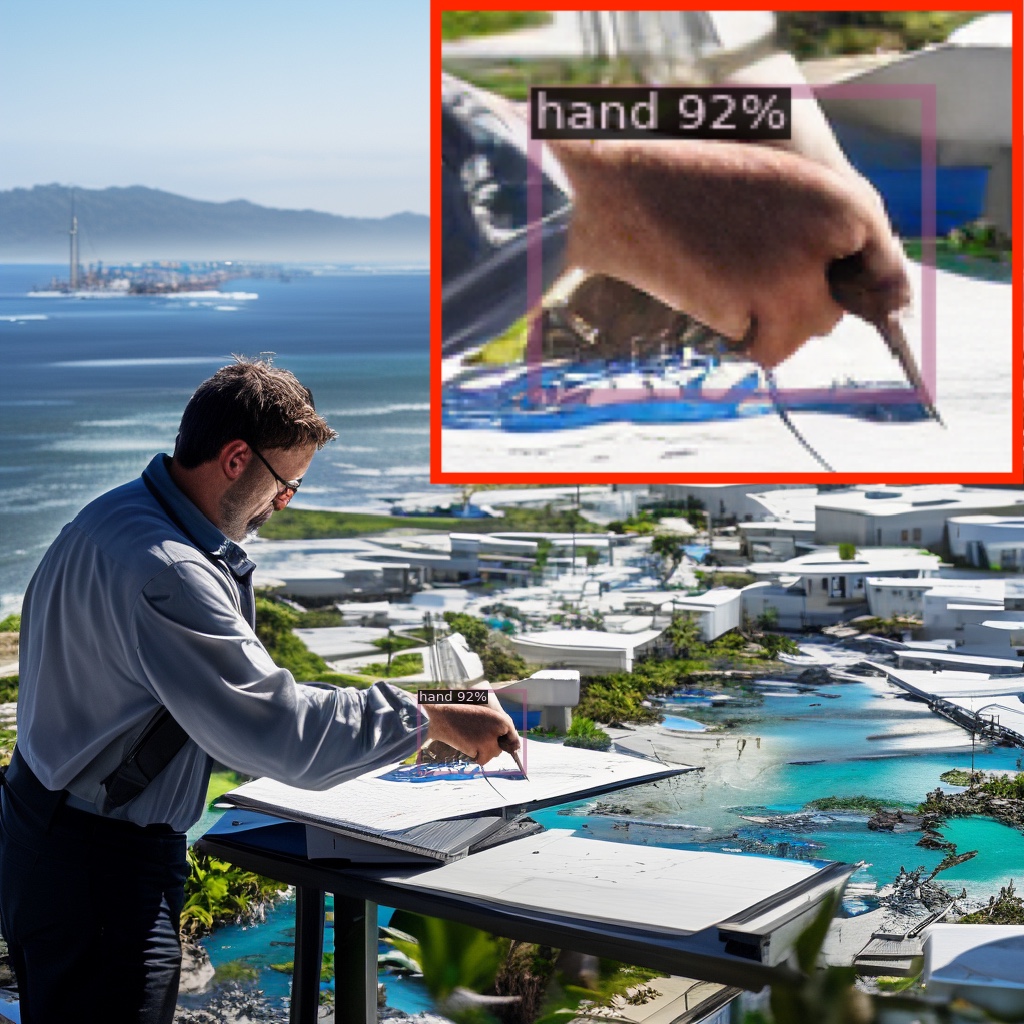}
    \caption{Weird hand in Sana}
    \label{fig:supp_ood_sana_3}
  \end{subfigure}
  \hfill
  \begin{subfigure}[b]{0.22\linewidth}
    \centering
    \includegraphics[width=\linewidth]{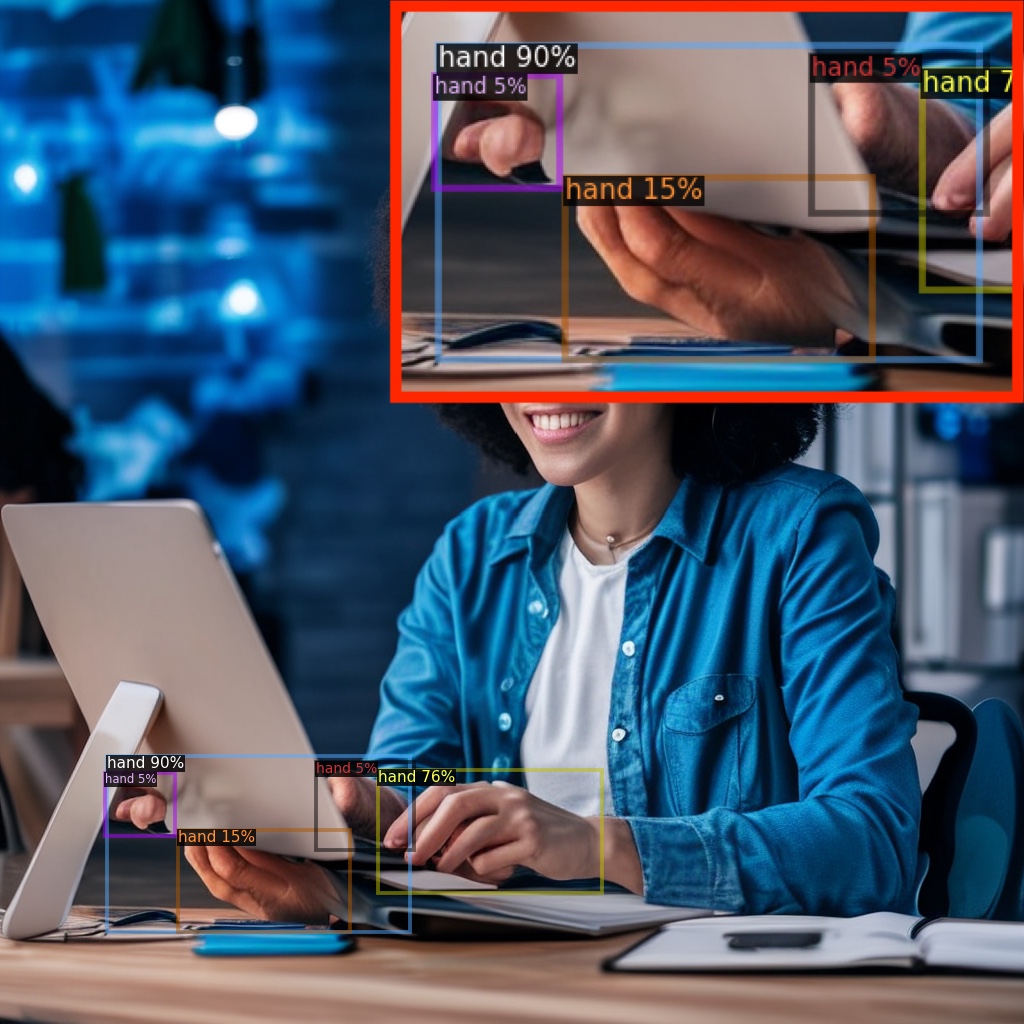}
    \caption{Weird hand in Sana}
    \label{fig:supp_ood_sana_4}
  \end{subfigure}

\vspace{0.2cm}

  \begin{subfigure}[b]{0.22\linewidth}
    \centering
    \includegraphics[width=\linewidth]{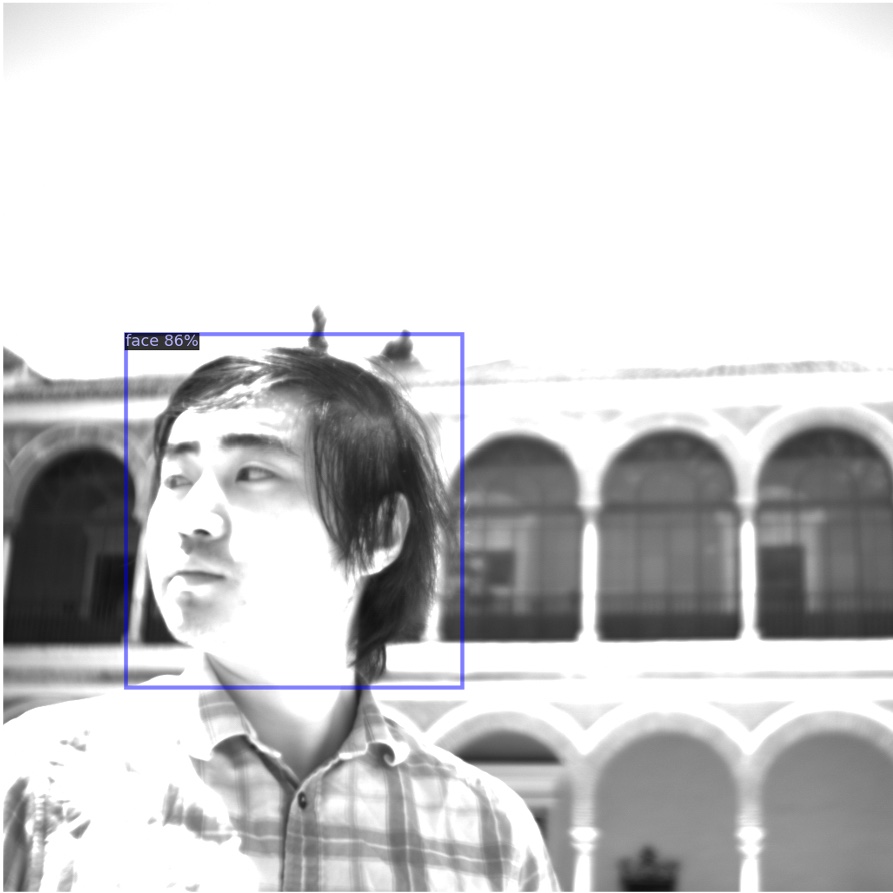}
    \caption{Weird face in 300W}
    \label{fig:supp_ood_300w_1}
  \end{subfigure}
  \hfill
  \begin{subfigure}[b]{0.22\linewidth}
    \centering
    \includegraphics[width=\linewidth]{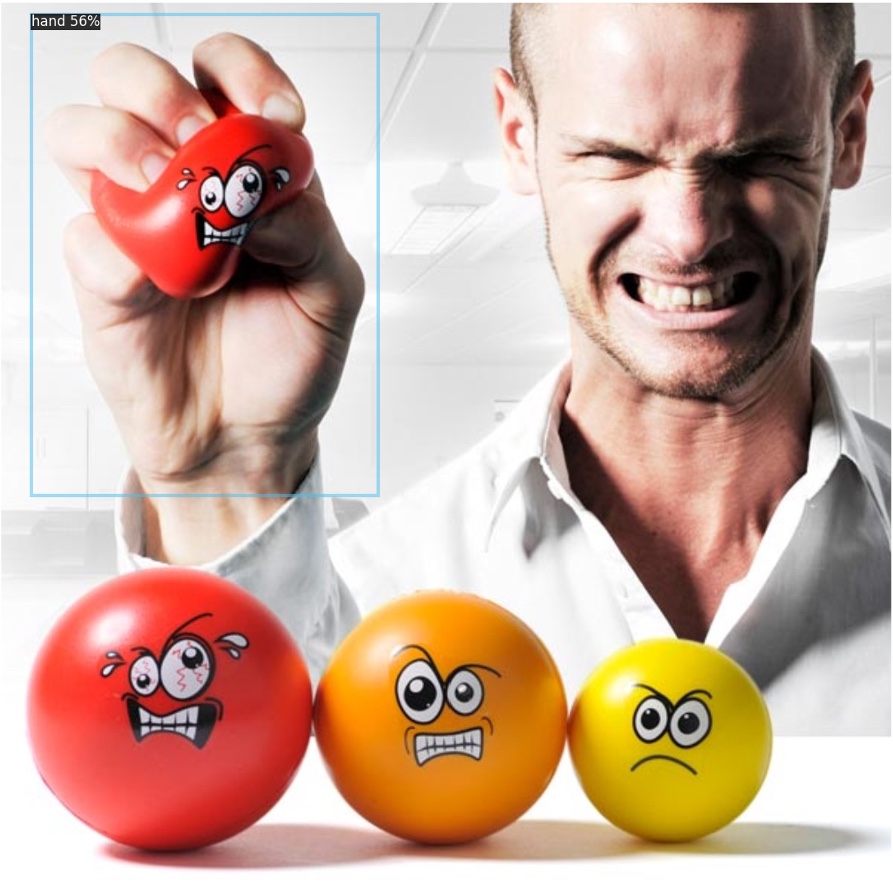}
    \caption{Weird hand in 300W}
    \label{fig:supp_ood_300w_2}
  \end{subfigure}
  \hfill
  \begin{subfigure}[b]{0.22\linewidth}
    \centering
    \includegraphics[width=\linewidth]{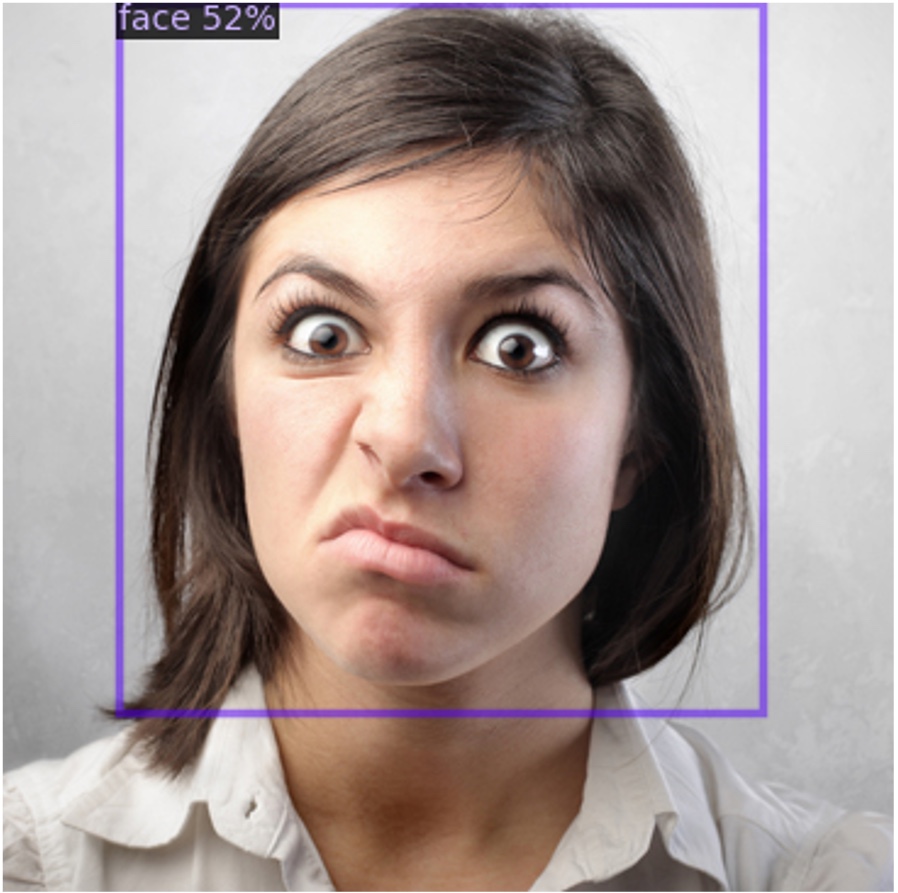}
    \caption{Weird face in 300W}
    \label{fig:supp_ood_300w_3}
  \end{subfigure}
  \hfill
  \begin{subfigure}[b]{0.22\linewidth}
    \centering
    \includegraphics[width=\linewidth]{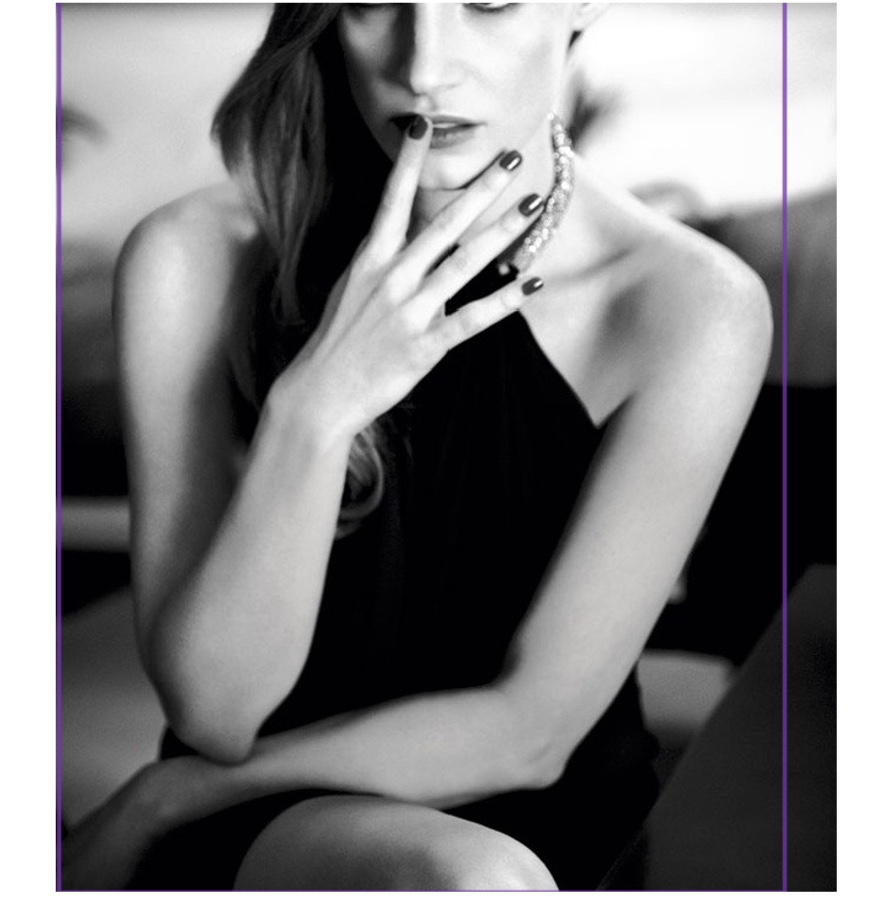}
    \caption{Extra hand and arm in 300W}
    \label{fig:supp_ood_300w_4}
  \end{subfigure}
\vspace{-0.2cm}
  \caption{Examples of predictions from our ~\abbrourmodel~ on unseen domains including SD1.4 (first row), PixArt-\(\Sigma\) (second row), FLUX.1-dev (third row), Sana (fourth row) and 300W (last row).
  }
  \label{fig:supp_ood}
\end{figure*}

\begin{figure*}[t] 
  \centering
  \begin{subfigure}[b]{0.32\linewidth}
    \centering
    \includegraphics[width=\linewidth]{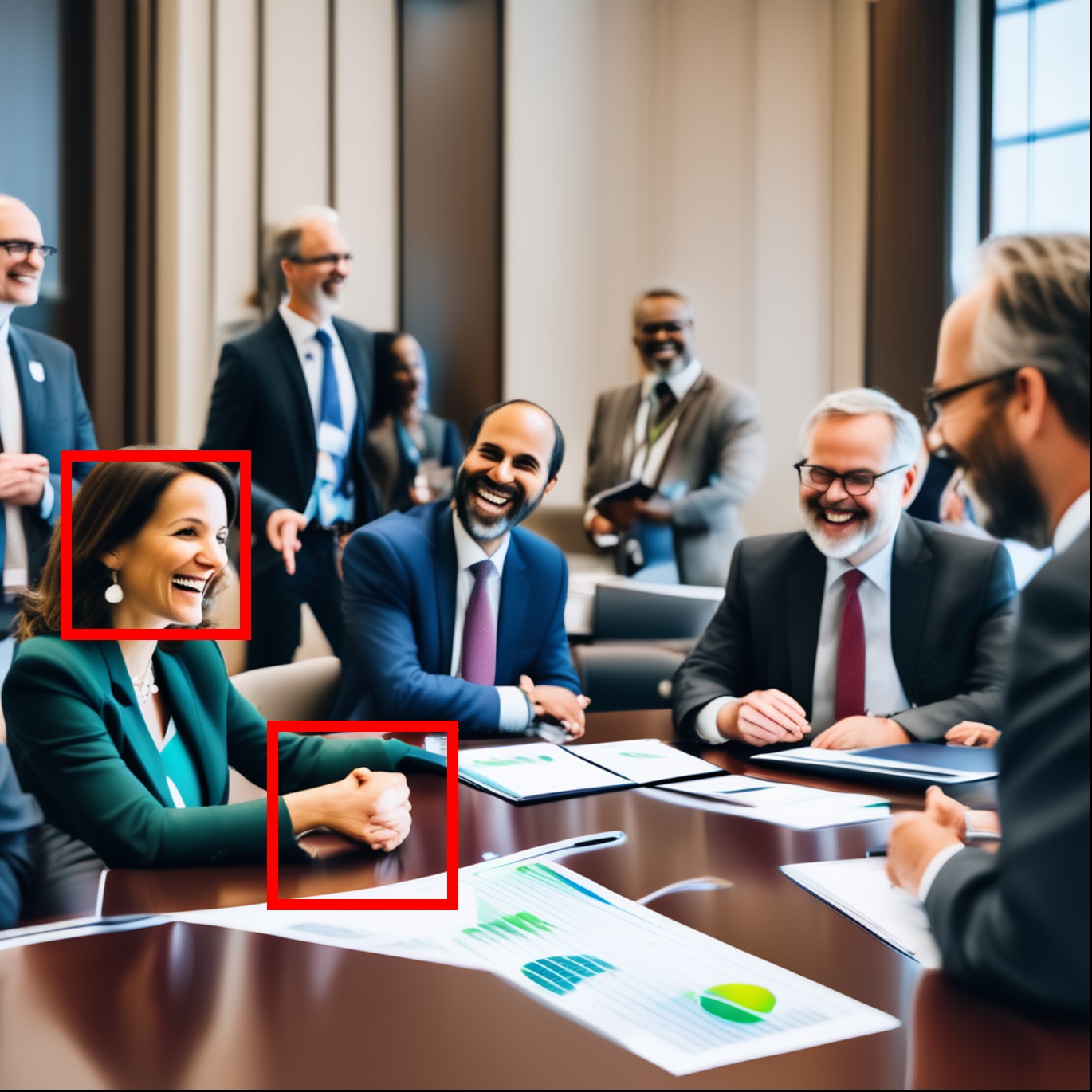}
    \label{fig:supp_ft_1_neg}
  \end{subfigure}
  \hfill
  \begin{subfigure}[b]{0.32\linewidth}
    \centering
    \includegraphics[width=\linewidth]{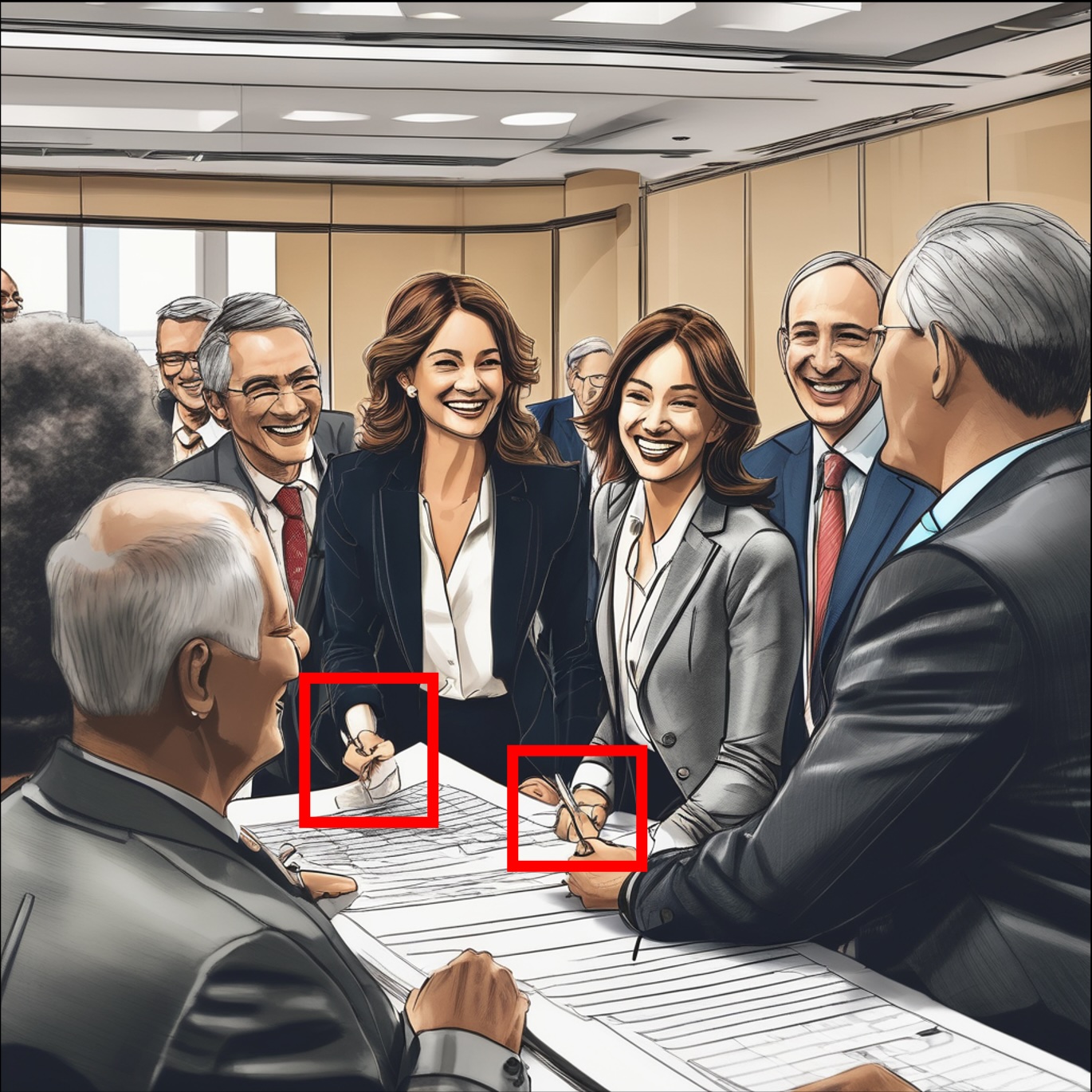}
    \label{fig:supp_ft_1_ori}
  \end{subfigure}
  \hfill
  \begin{subfigure}[b]{0.32\linewidth}
    \centering
    \includegraphics[width=\linewidth]{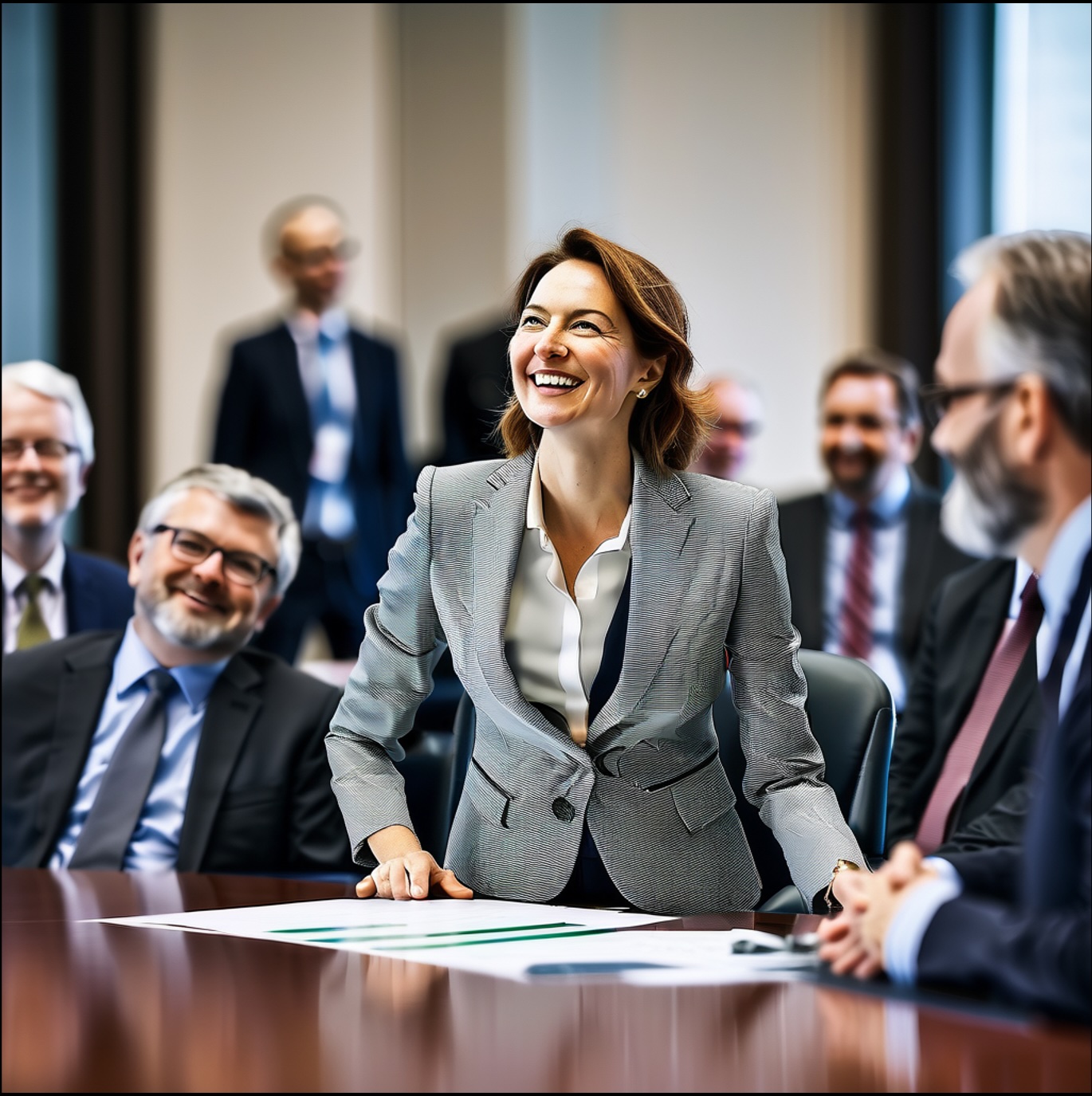}
    \label{fig:supp_ft_1_ours}
  \end{subfigure}

  \begin{subfigure}[b]{0.32\linewidth}
    \centering
    \includegraphics[width=\linewidth]{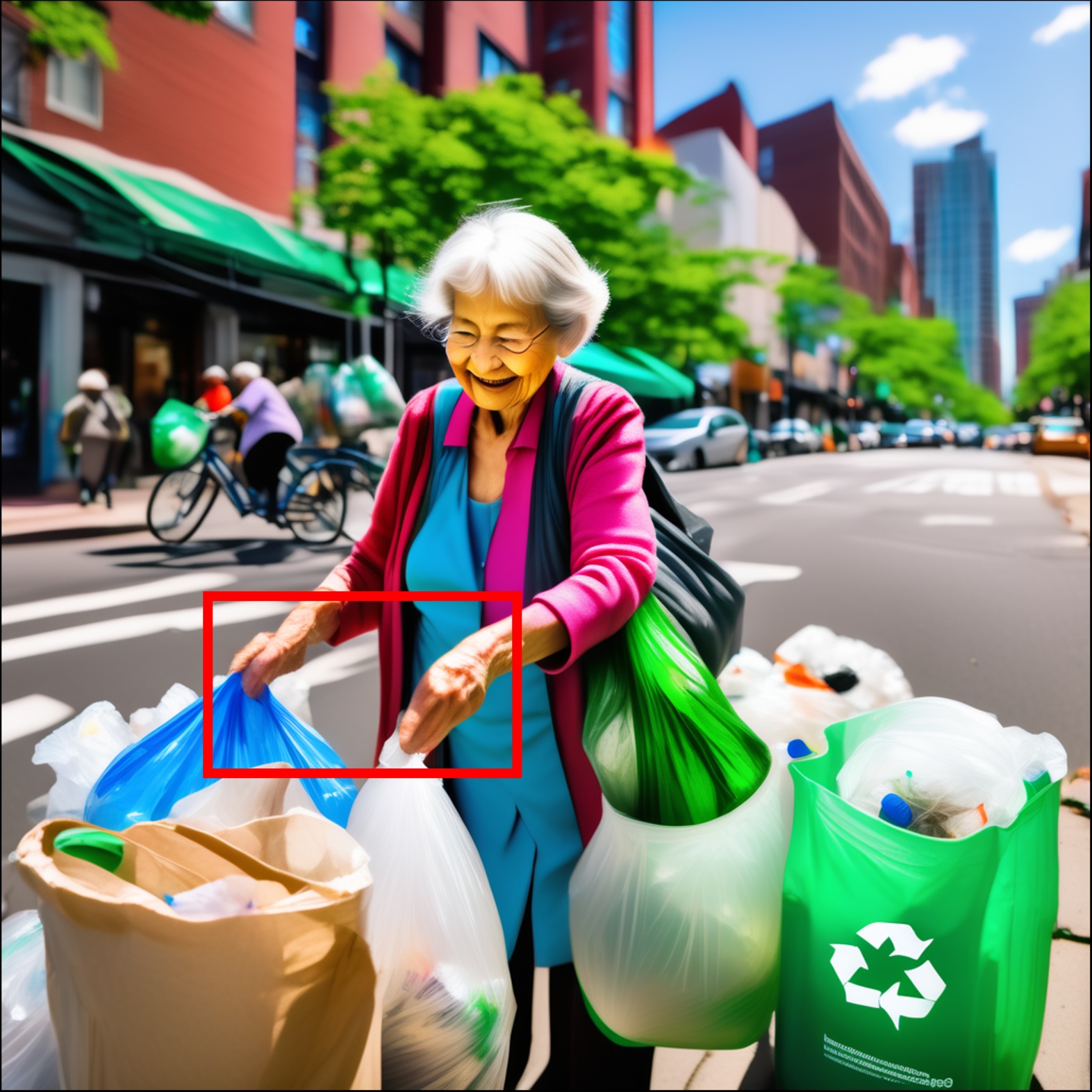}
    \label{fig:supp_ft_2_neg}
  \end{subfigure}
  \hfill
  \begin{subfigure}[b]{0.32\linewidth}
    \centering
    \includegraphics[width=\linewidth]{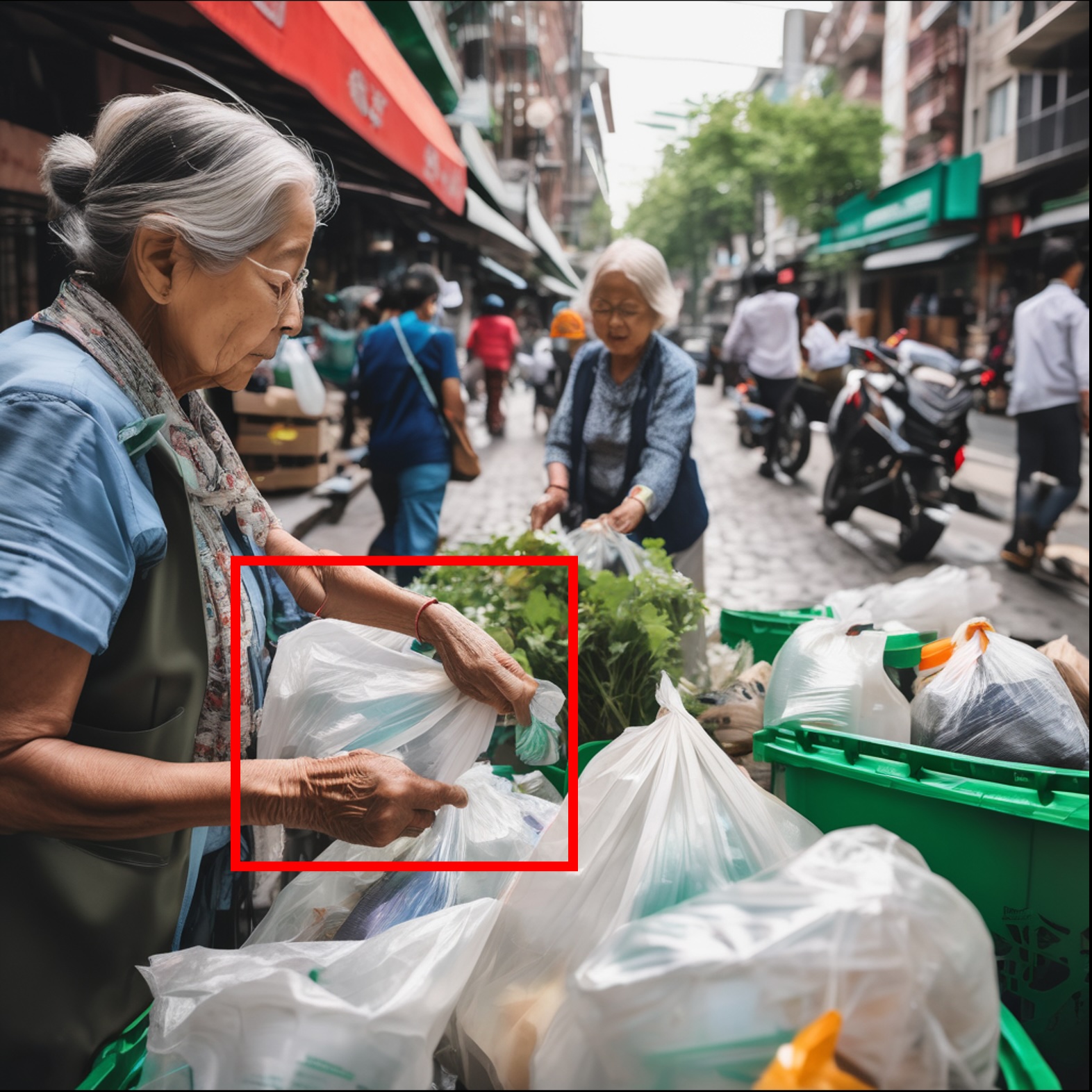}
    \label{fig:supp_ft_2_ori}
  \end{subfigure}
  \hfill
  \begin{subfigure}[b]{0.32\linewidth}
    \centering
    \includegraphics[width=\linewidth]{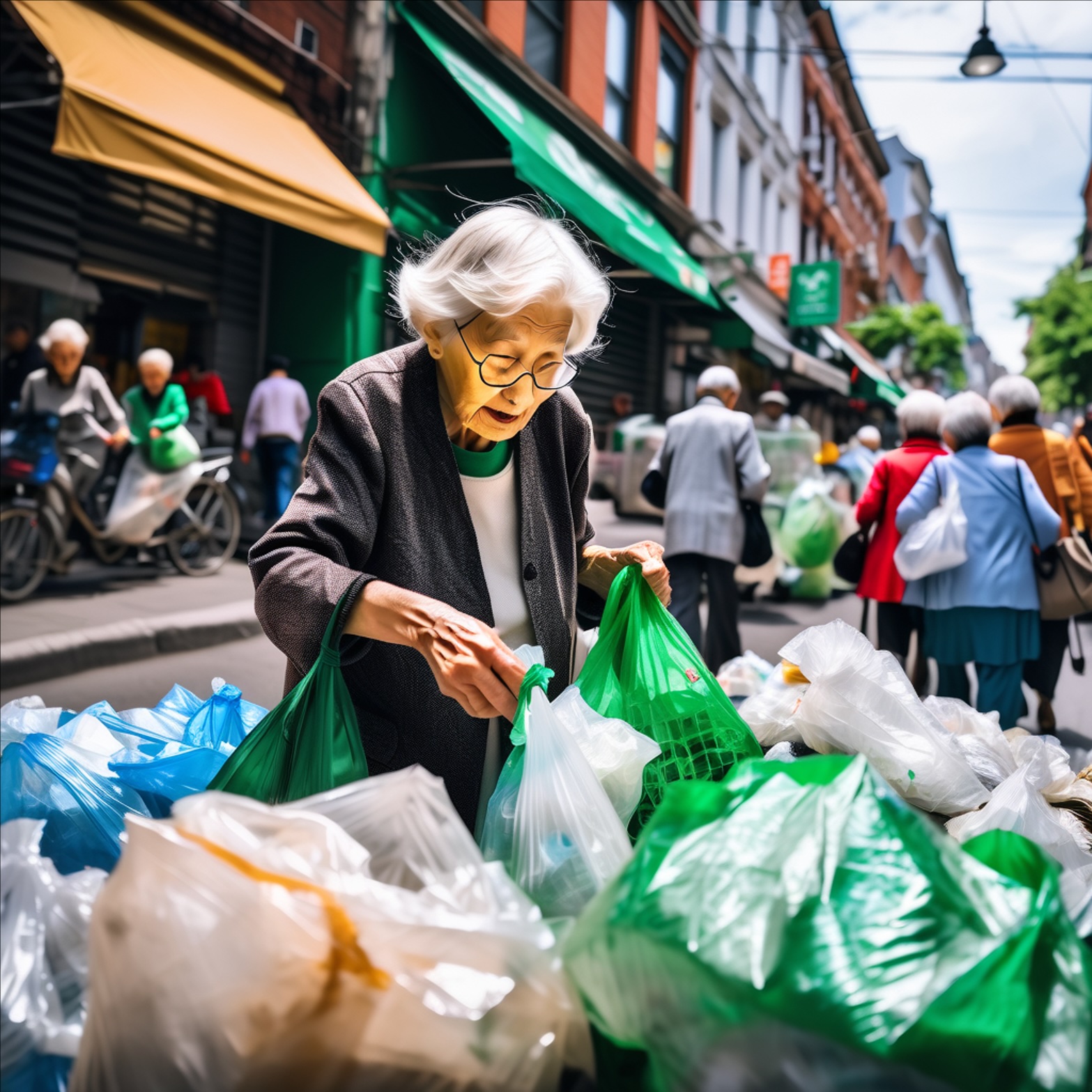}
    \label{fig:supp_ft_2_ours}
  \end{subfigure}

  \begin{subfigure}[b]{0.32\linewidth}
    \centering
    \includegraphics[width=\linewidth]{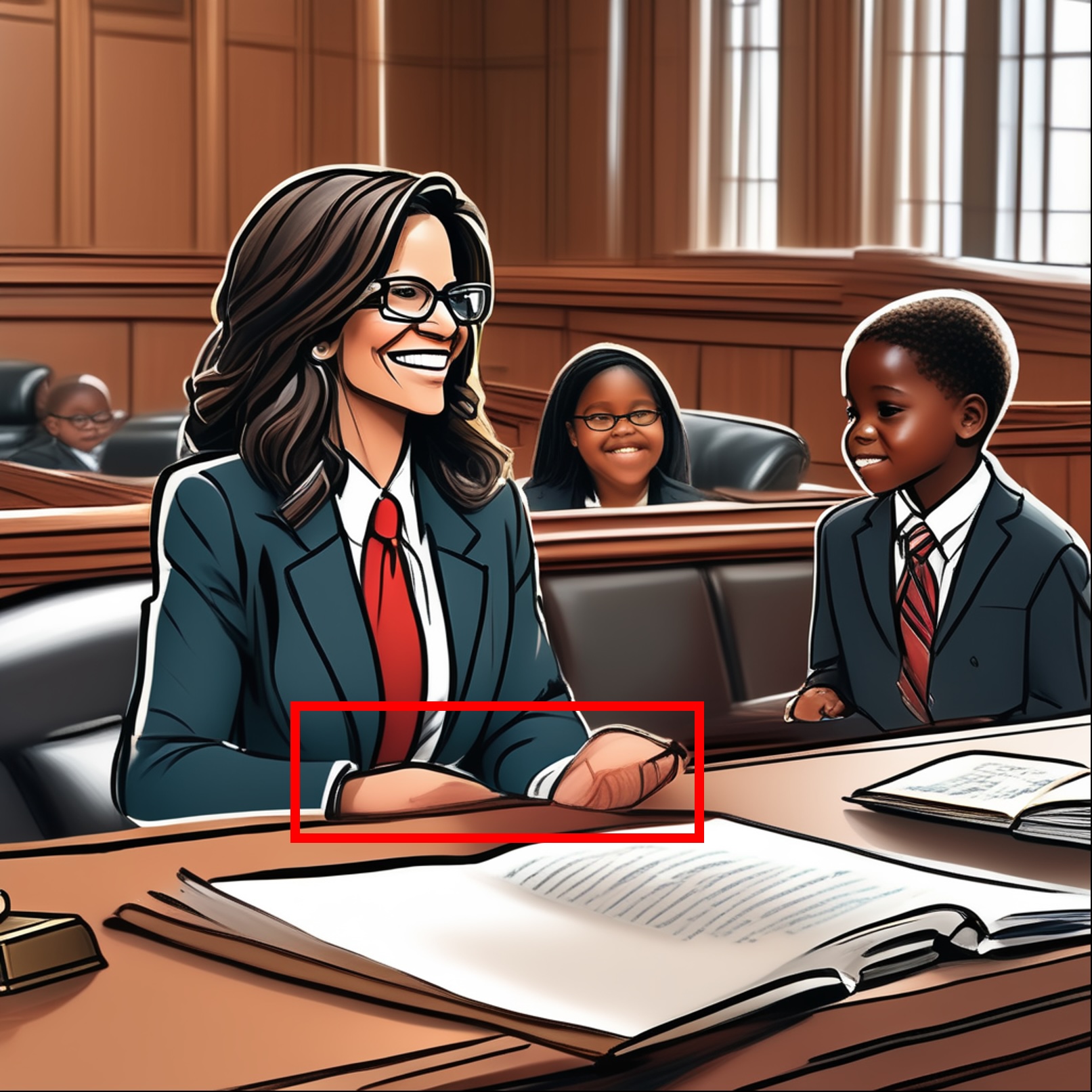}
    \caption*{Negative As Positive}
    \label{fig:supp_ft_3_neg}
  \end{subfigure}
  \hfill
  \begin{subfigure}[b]{0.32\linewidth}
    \centering
    \includegraphics[width=\linewidth]{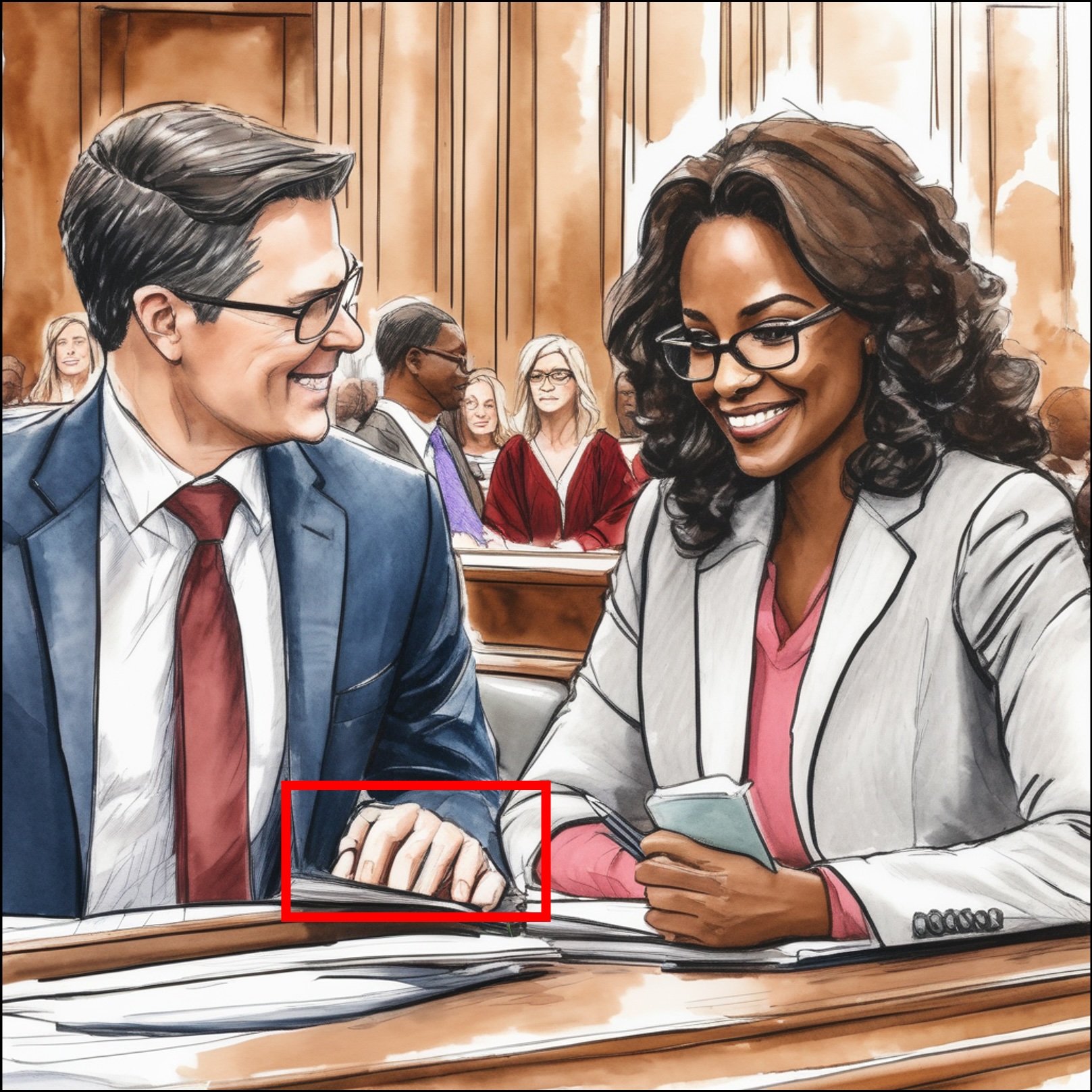}
    \caption*{Original SDXL}
    \label{fig:supp_ft_3_ori}
  \end{subfigure}
  \hfill
  \begin{subfigure}[b]{0.32\linewidth}
    \centering
    \includegraphics[width=\linewidth]{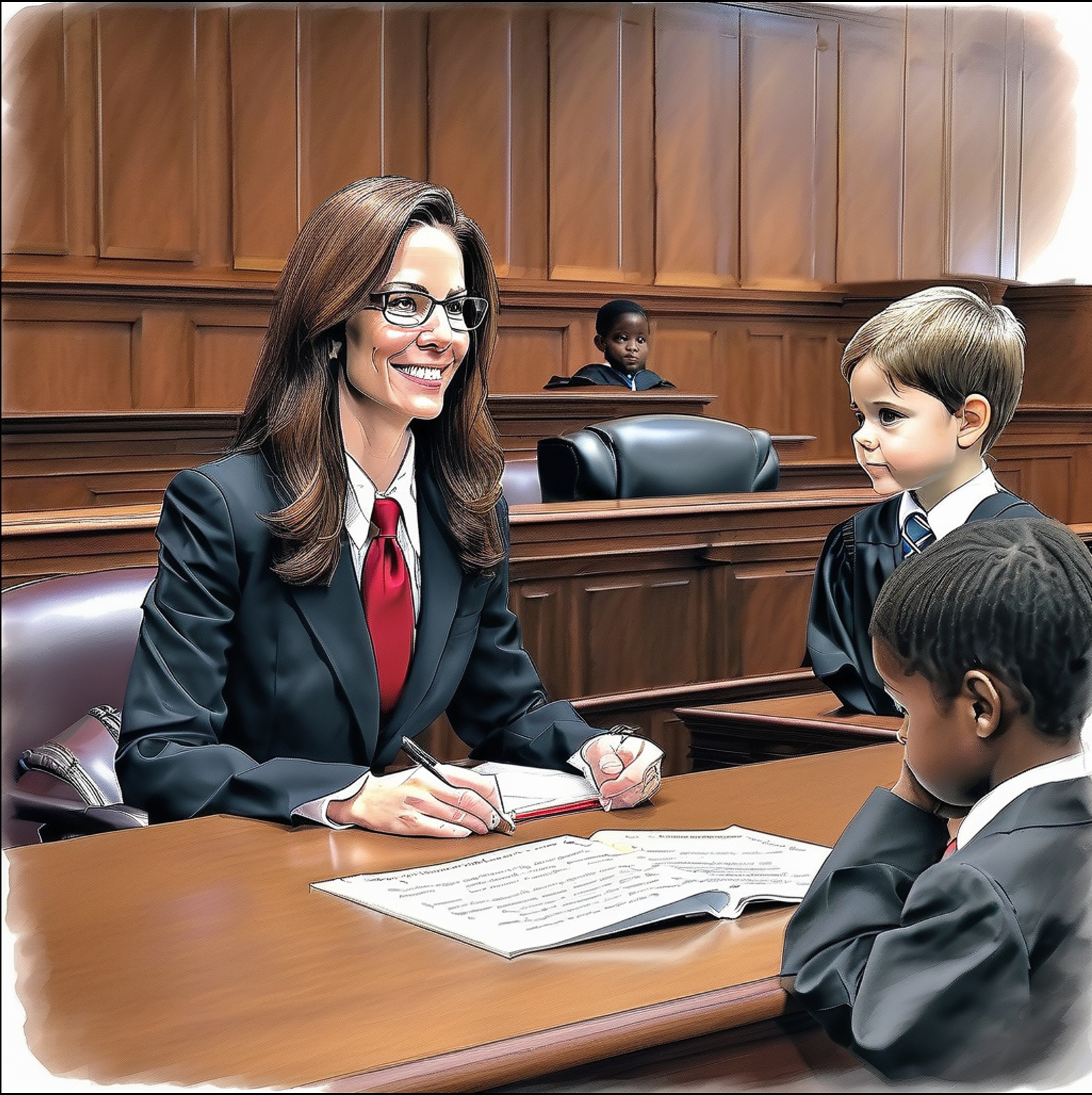}
    \caption*{Finetuned SDXL}
    \label{fig:supp_ft_3_ours}
  \end{subfigure}
\end{figure*}

\begin{figure*}[t] 
  \ContinuedFloat 
  \centering
  \begin{subfigure}[b]{0.32\linewidth}
    \centering
    \includegraphics[width=\linewidth]{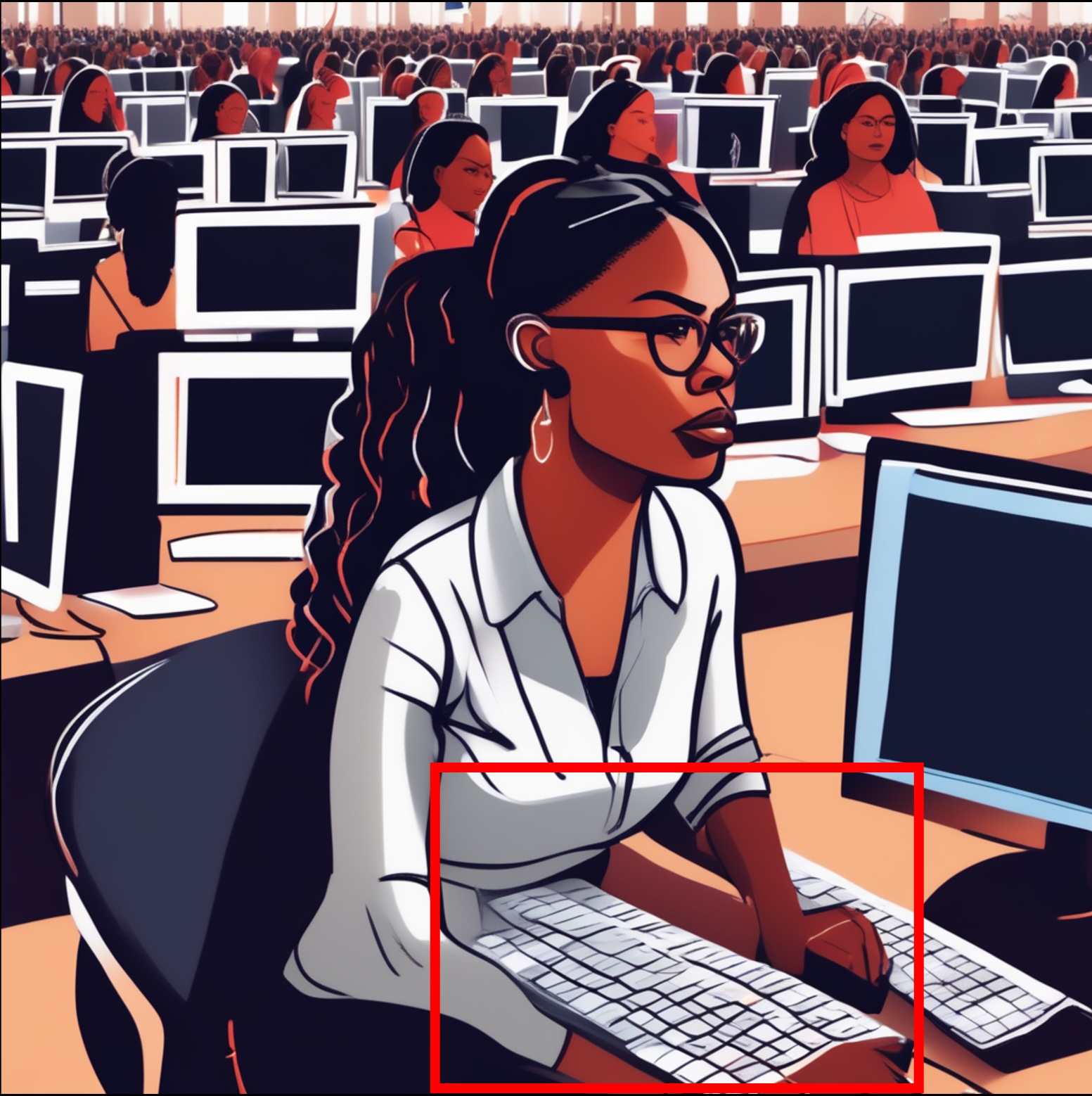}
    \label{fig:supp_ft_4_neg}
  \end{subfigure}
  \hfill
  \begin{subfigure}[b]{0.32\linewidth}
    \centering
    \includegraphics[width=\linewidth]{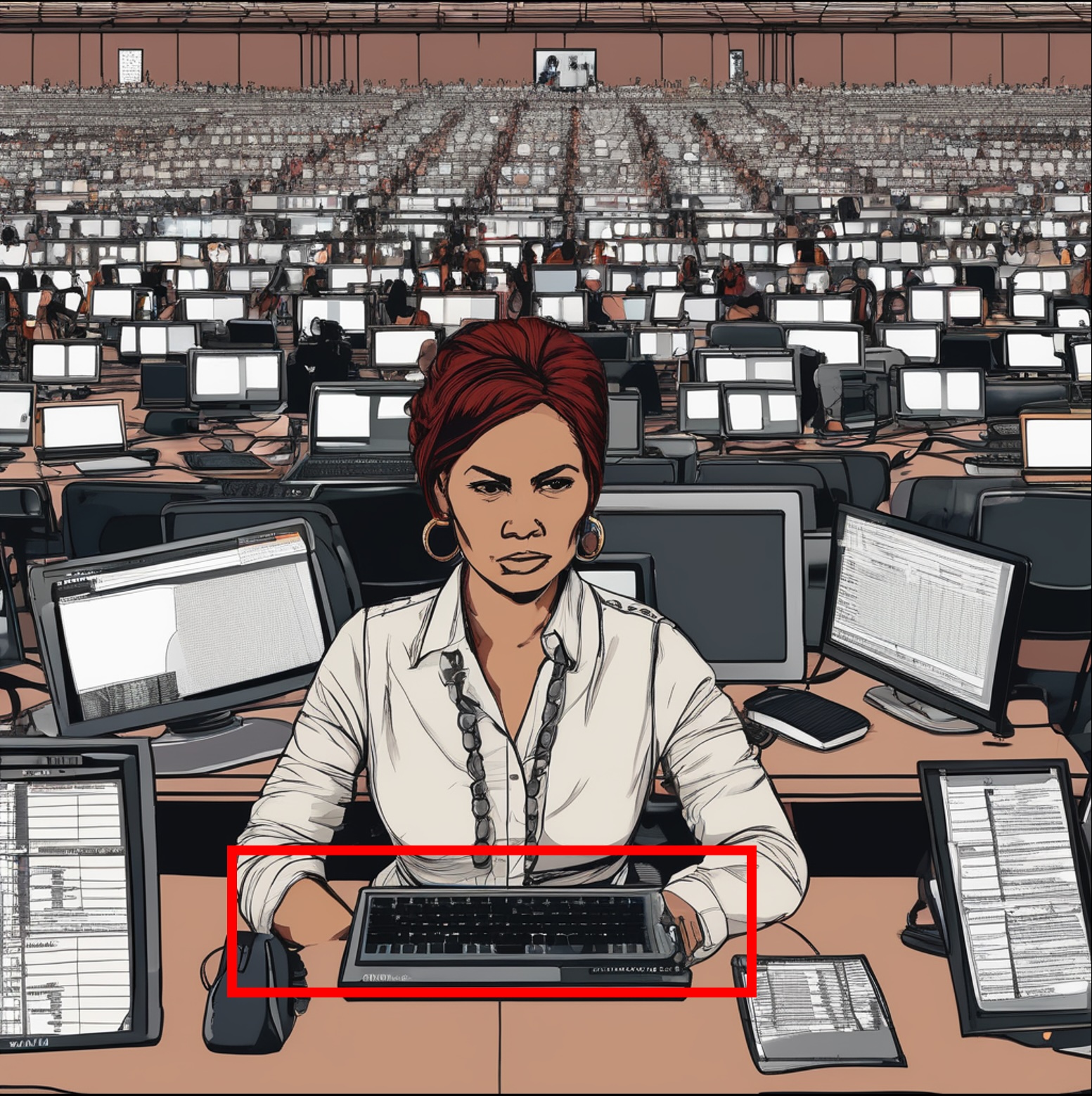}
    \label{fig:supp_ft_4_ori}
  \end{subfigure}
  \hfill
  \begin{subfigure}[b]{0.32\linewidth}
    \centering
    \includegraphics[width=\linewidth]{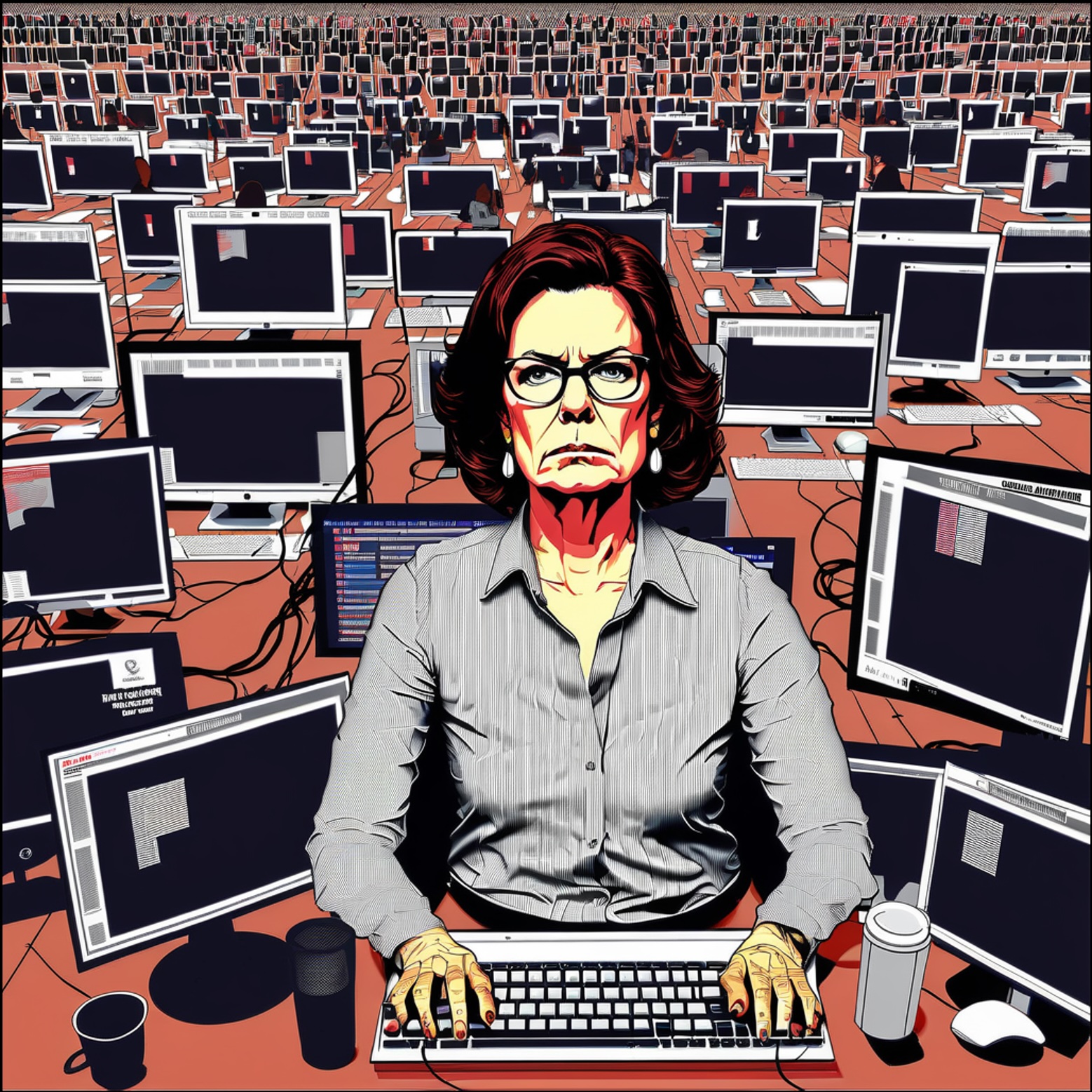}
    \label{fig:supp_ft_4_ours}
  \end{subfigure}

  \begin{subfigure}[b]{0.32\linewidth}
    \centering
    \includegraphics[width=\linewidth]{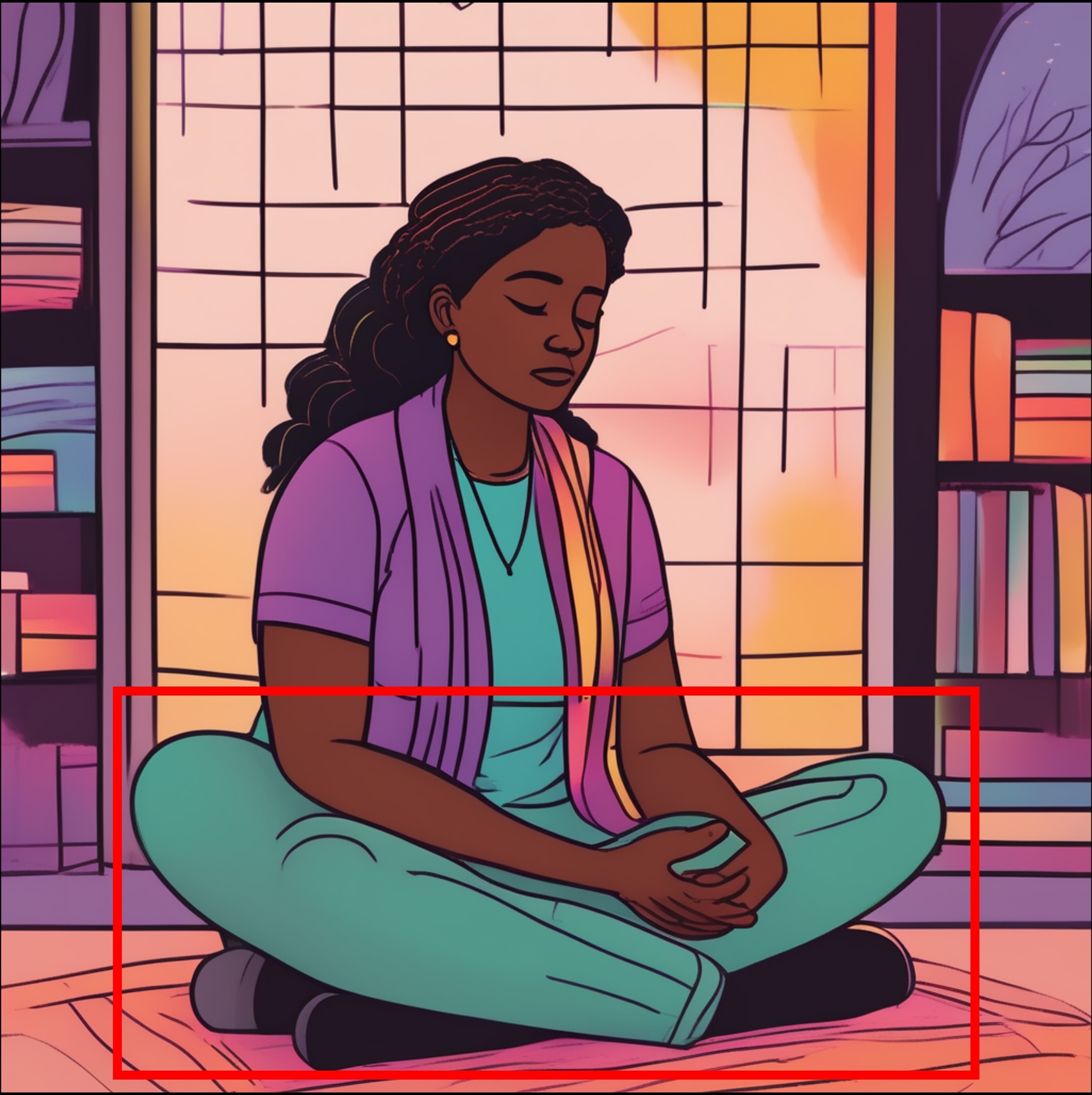}
    \label{fig:supp_ft_5_neg}
  \end{subfigure}
  \hfill
  \begin{subfigure}[b]{0.32\linewidth}
    \centering
    \includegraphics[width=\linewidth]{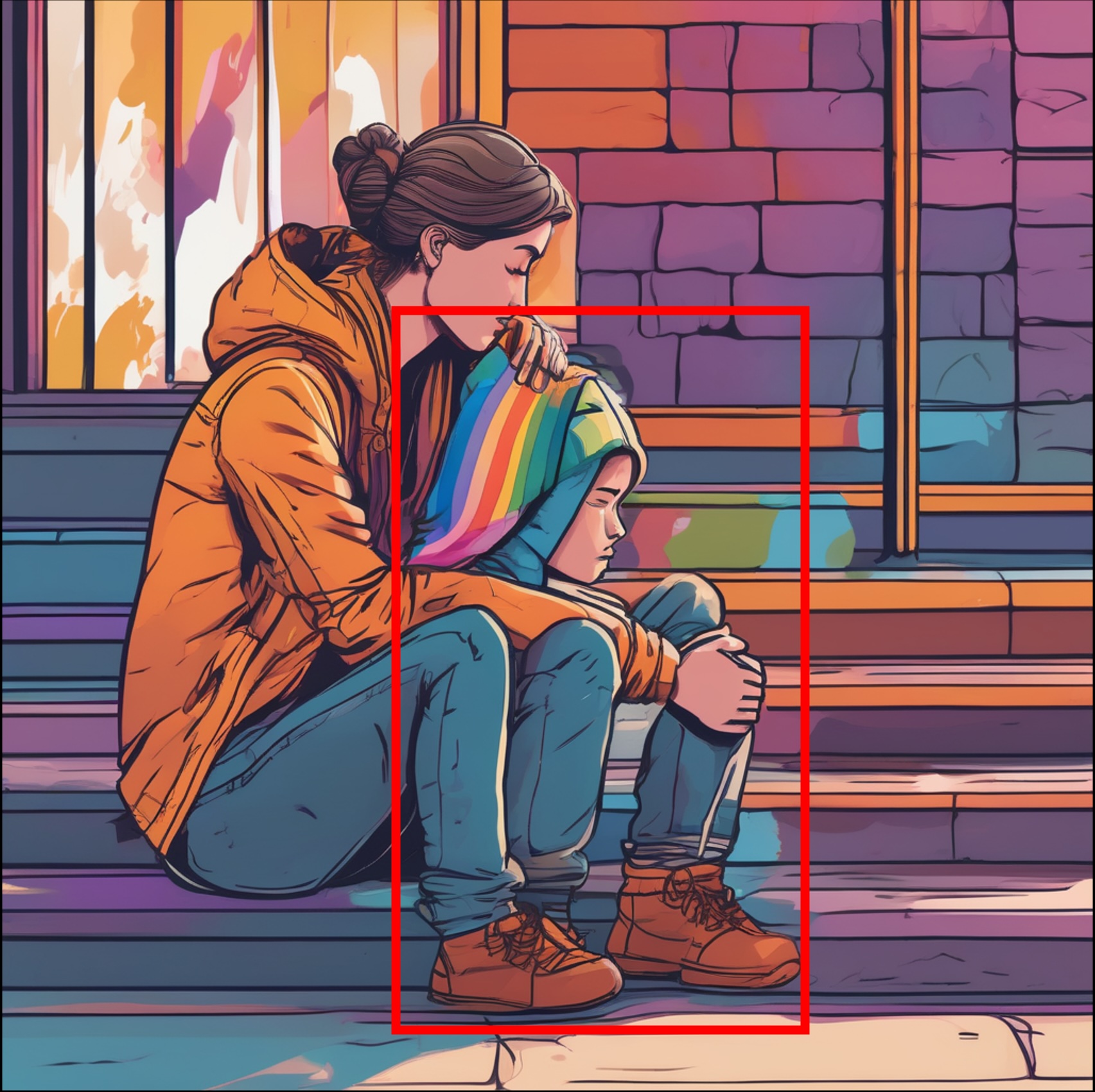}
    \label{fig:supp_ft_5_ori}
  \end{subfigure}
  \hfill
  \begin{subfigure}[b]{0.32\linewidth}
    \centering
    \includegraphics[width=\linewidth]{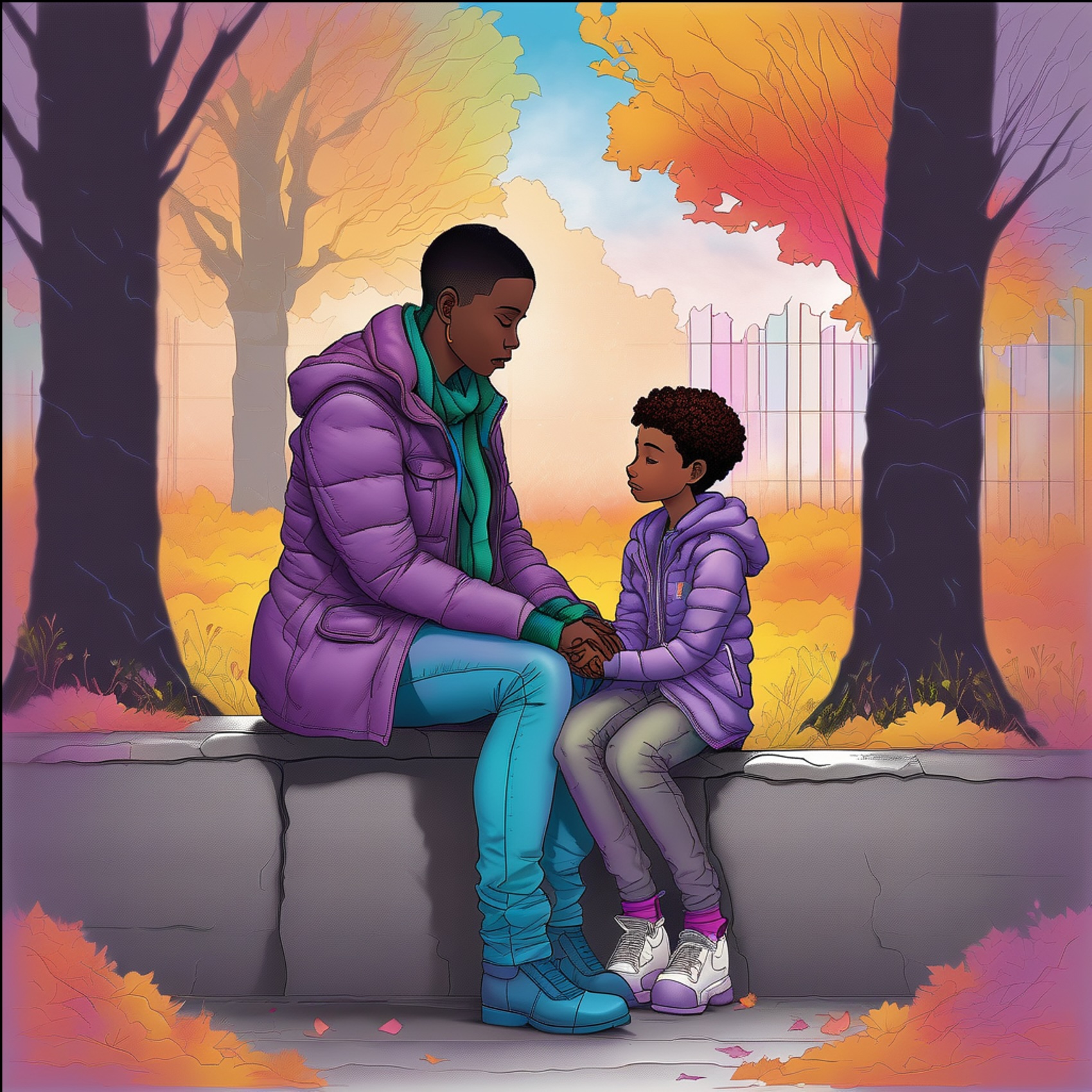}
    \label{fig:supp_ft_5_ours}
  \end{subfigure}

  \begin{subfigure}[b]{0.32\linewidth}
    \centering
    \includegraphics[width=\linewidth]{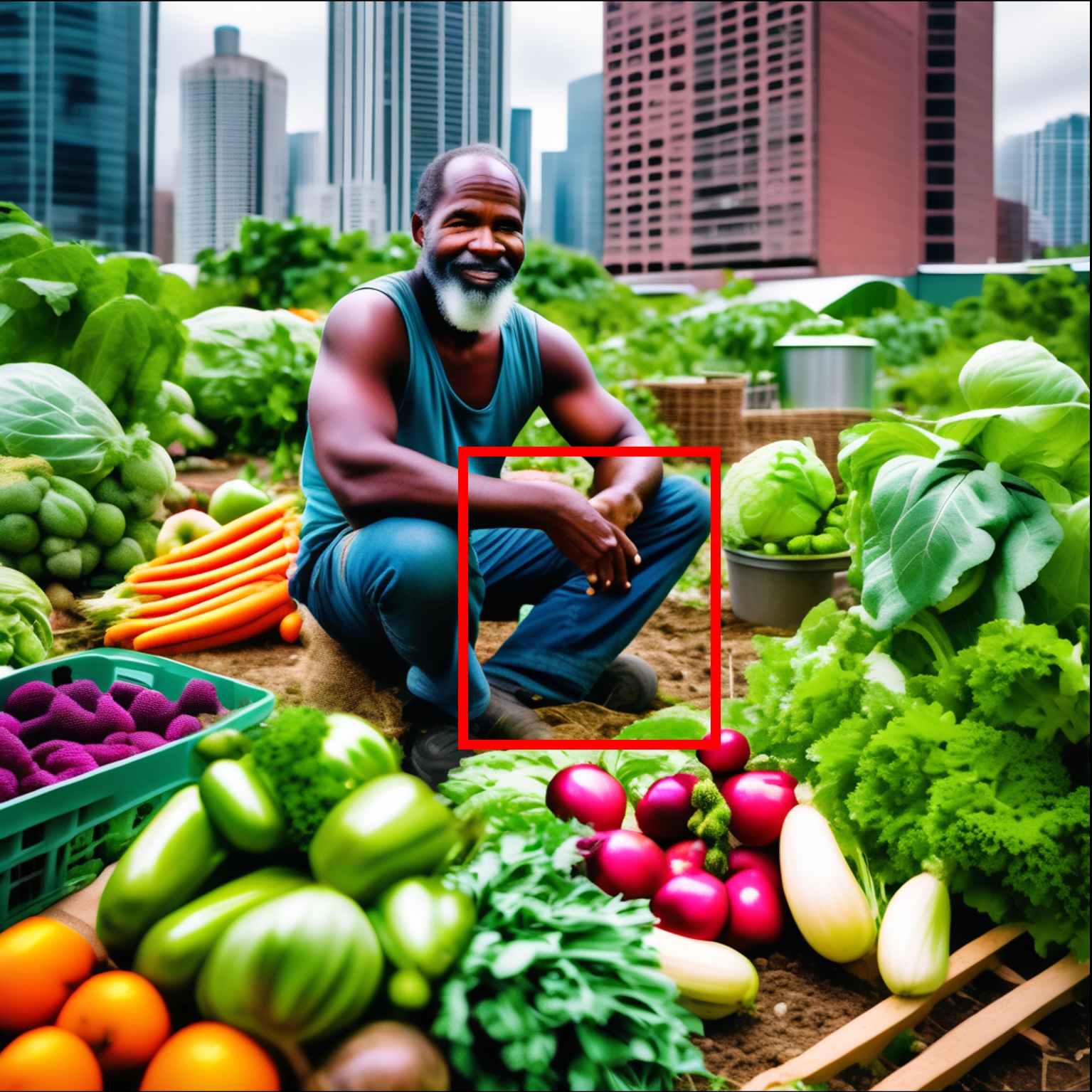}
    \caption*{Negative As Positive}
    \label{fig:supp_ft_6_neg}
  \end{subfigure}
  \hfill
  \begin{subfigure}[b]{0.32\linewidth}
    \centering
    \includegraphics[width=\linewidth]{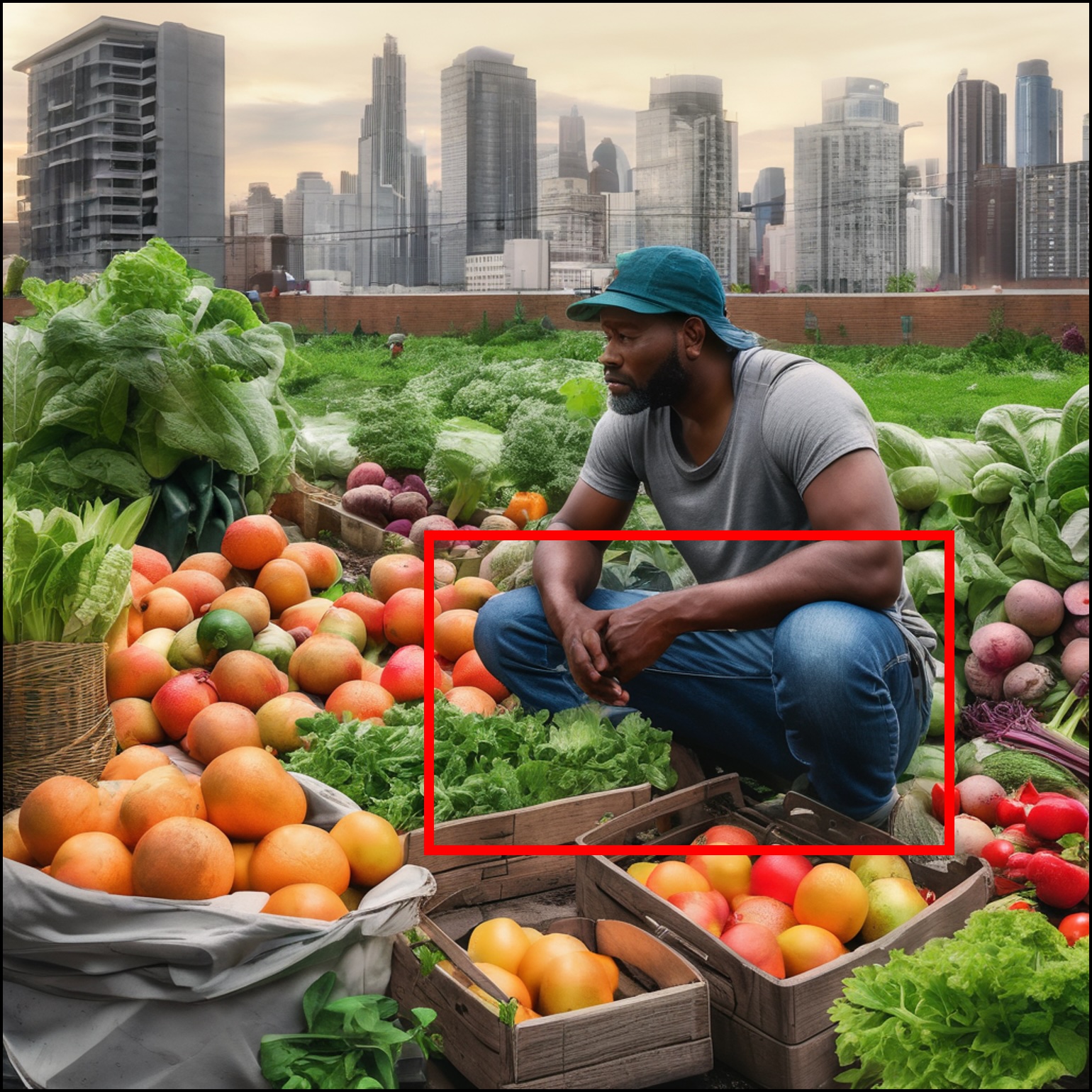}
    \caption*{Original SDXL}
    \label{fig:supp_ft_6_ori}
  \end{subfigure}
  \hfill
  \begin{subfigure}[b]{0.32\linewidth}
    \centering
    \includegraphics[width=\linewidth]{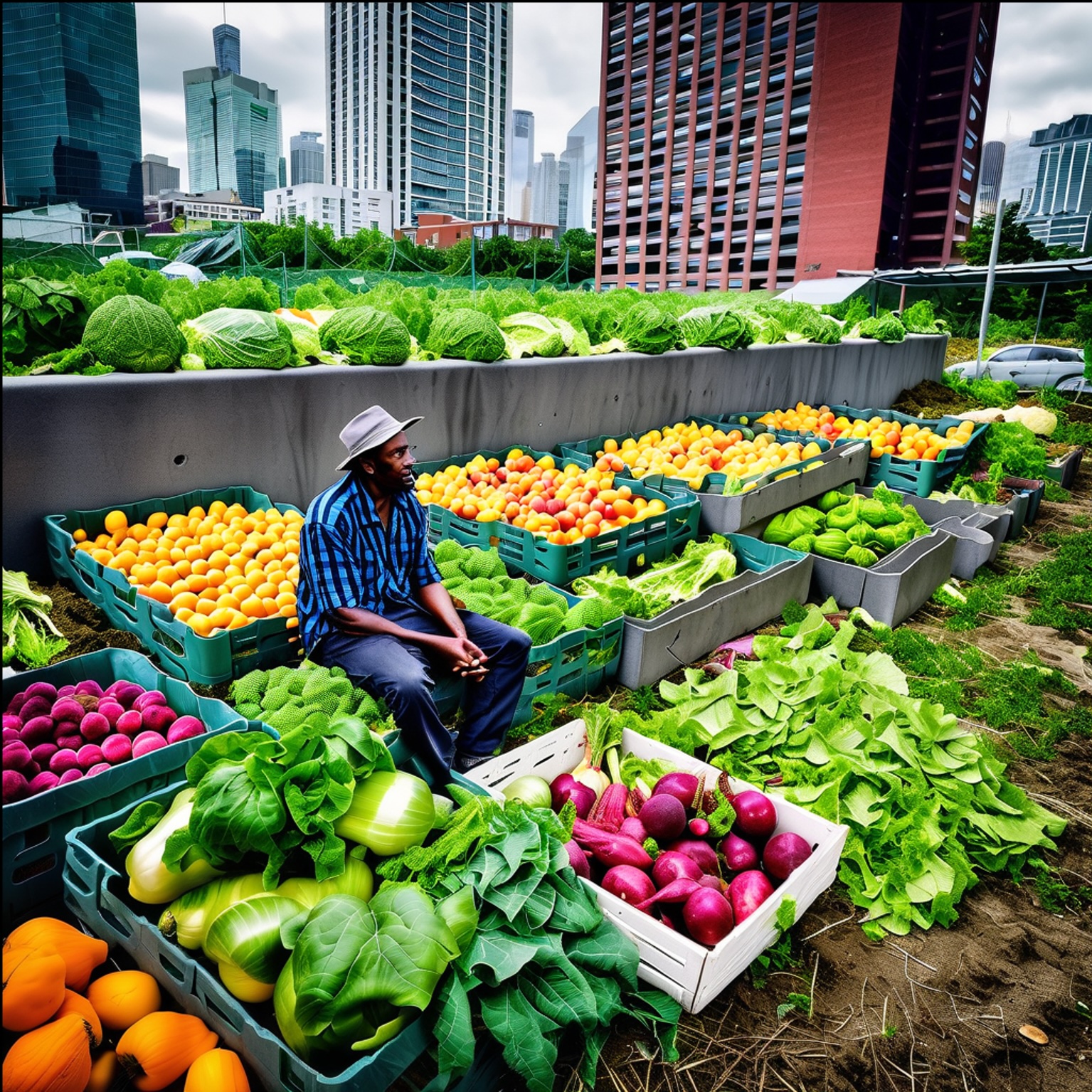}
    \caption*{Finetuned SDXL}
    \label{fig:supp_ft_6_ours}
  \end{subfigure}
  \caption{More results from the original SDXL model and our finetuned SDXL model. In the first column, images generated by the finetuned model use special identifiers as part of the prompt instead of as negative prompts. In the second and third columns, images from the original and the finetuned SDXL models use special identifiers as negative prompts. }
\label{fig:supp_ft}
\end{figure*}